\documentclass[runningheads]{llncs}

\usepackage{eccv}

\usepackage{eccvabbrv}

\usepackage{graphicx}
\usepackage{booktabs}
\usepackage{soul}
\usepackage[accsupp]{axessibility}  

\makeatletter
\def\thanks#1{\protected@xdef\@thanks{\@thanks\protect\footnotetext{#1}}}
\makeatother

\usepackage{orcidlink}
\usepackage{marvosym}
\usepackage{multirow}   
\usepackage{booktabs}
\usepackage{listings}
\usepackage{xcolor}
\usepackage[table]{xcolor}
\usepackage{minitoc}
\usepackage{enumitem}
\usepackage{pifont}
\usepackage{amsmath}
\usepackage{amssymb}
\usepackage{mathtools}
\usepackage{makecell}
\usepackage{tcolorbox}
\usepackage{wrapfig}
\usepackage{booktabs}
\usepackage{multirow}
\usepackage{caption}
\usepackage[normalem]{ulem}

\usepackage{hyperref}

\definecolor{darkgreen}{HTML}{04bf29}
\definecolor{darkred}{HTML}{D1191F}

\definecolor{grey}{HTML}{bfbfbf}
\definecolor{clipr}{HTML}{E9F9D1}

\definecolor{cliprGrey}{HTML}{786E6E}
\definecolor{cliprBlue}{HTML}{5675A2}
\definecolor{cliprLightBlue}{HTML}{E8EDF8}
\definecolor{cliprPurple}{HTML}{722E5F}
\definecolor{cliprLightPurple}{HTML}{FBF3FA}

\usepackage{etoolbox}
\usepackage{risys-title}

\RisysTitle{ReasonCLIP-58M: Visually Grounded Commonsense Reasoning Supervision for CLIP}
\RisysAuthor[1]{Sicheng Zhang}
\RisysAuthor[1,2,*]{Muzammal Naseer}
\RisysAuthor[3]{Binzhu Xie}
\RisysAuthor[1]{Naufal Suryanto}
\RisysAuthor[3]{Shi Qiu}
\RisysAuthor[1]{Jamal Bentahar}
\RisysAuthor[4]{Naveed Akhtar}
\RisysAuthor[5]{Mubarak Shah}
\RisysAffil[1]{Khalifa University}
\RisysAffil[2]{University of Western Australia}
\RisysAffil[3]{The Chinese University of Hong Kong}
\RisysAffil[4]{University of Melbourne}
\RisysAffil[5]{University of Central Florida}
\RisysContribution[*]{Corresponding author}
\RisysGithub{https://github.com/RISys-Lab/ReasonCLIP}
\RisysModel{https://hf.co/collections/RISys-Lab/reasonclip-models}
\RisysDataset{https://hf.co/collections/RISys-Lab/reasonclip-data}
\RisysBenchmark{https://hf.co/datasets/RISys-Lab/RCLIP-Bench}

\RisysAbstract{CLIP and its variants are widely adopted visual backbones in multimodal systems, but their pretraining remains dominated by descriptive image-text alignment. As downstream applications increasingly demand visually grounded commonsense inference and compositional reasoning, it remains unclear whether CLIP-style encoders can support such reasoning without architectural changes. To address this, we present ReasonCLIP-58M, a continual pretraining framework that integrates large-scale reasoning supervision into CLIP-style models through our two-stage strategy, which progressively integrates reasoning signals while preserving descriptive alignment, followed by category-structured reasoning supervision. To support this framework, we construct two complementary datasets and a benchmark: ReasonLite-42M, with open-form, visually verifiable reasoning captions; ReasonPro-16M, with category-specific reasoning supervision; and RCLIP-Bench for diagnostic evaluation of visually grounded reasoning. We train a family of ReasonCLIP that improves visually grounded commonsense and compositional reasoning while also enhancing zero-shot retrieval performance. As a drop-in visual encoder for multimodal large language models such as LLaVA-NeXT, ReasonCLIP delivers consistent gains without additional inference cost, demonstrating that structured reasoning supervision enhances the expressive capacity of CLIP-style visual representations.}

\begin{document}
\RisysMakeTitle


 



\section{Introduction}
\begin{figure}[htbp]
\centering
\includegraphics[width=\linewidth]{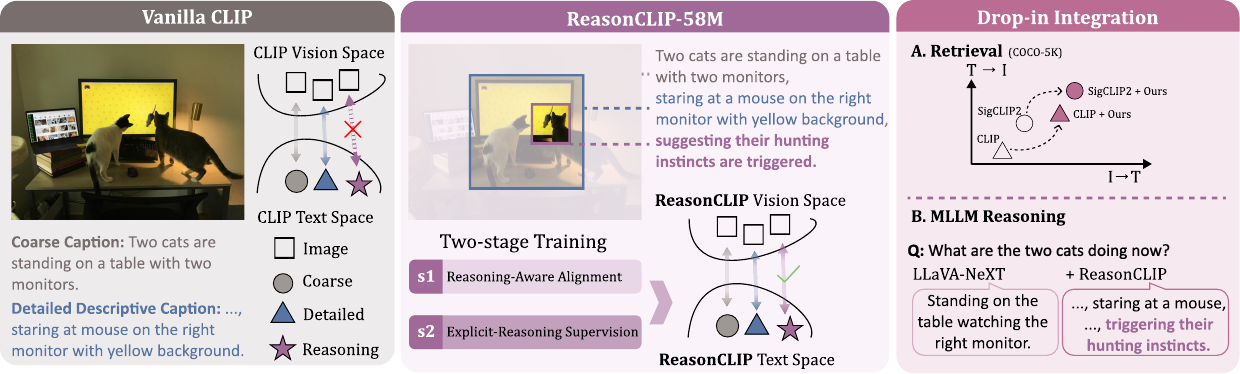}
\captionof{figure}{
\textbf{ReasonCLIP-58M: From Description to Reasoning.}
ReasonCLIP-58M enables visually grounded reasoning in CLIP through two-stage continual pretraining with reasoning-aware and category-level supervision, improving zero-shot retrieval, structured reasoning, and MLLM performance without modifying the backbone.
}
\label{fig:teaser}
\end{figure}
Contrastive Language-Image Pretraining (CLIP) \cite{radford2021learning} has emerged as a foundational vision-language model which aligns image and text representations through large-scale contrastive learning. Since its introduction, CLIP-style visual encoders \cite{zhai2023sigmoid,sun2024eva,tschannen2025siglip} have become the default backbone for multimodal systems, including Multimodal Large Language Models (MLLMs) \cite{li2023blip,alayrac2022flamingo,li2024llava,Qwen3-VL}. 
Despite their widespread adoption, most CLIP models are pretrained on large-scale \emph{descriptive} image-text datasets \cite{jia2021scaling,schuhmann2021laion}, where supervision primarily encourages semantic correspondence rather than structured reasoning over visual content. While highly effective for retrieval, such training is not explicitly designed to support visually grounded reasoning~\cite{jiang2025mme,zhang2024mmvp,xu2025visulogic,kamath2026scale,cui2024more,wan2022bridging}.

Advanced MLLMs increasingly require capabilities beyond descriptive alignment, including compositional reasoning~\cite{johnson2017clevr,hudson2019gqa}, commonsense inference~\cite{yue2024mmmu,hu2025video}, and multi-step understanding~\cite{wei2022chain,xu2024llava}, when processing complex and diverse real-world scenarios.
%
However, as summarized in Tab.~\ref{tab:main: model_compare}, recent CLIP advances largely focus on scaling descriptive data \cite{datacomp,xu2023demystifying,zhai2023sigmoid} or refining architectures \cite{zhang2024long,sun2024eva,bolyaperception,kwon2023efficient}. This emphasis creates a mismatch between pretraining objectives and reasoning-oriented downstream demands, questioning whether descriptive alignment alone is sufficient for visually grounded reasoning.

Motivated by this gap, we revisit CLIP and investigate whether its representation space can be extended through reasoning-oriented continual pretraining \emph{without modifying its architecture}. 
As illustrated in Fig.~\ref{fig:teaser}, we propose ReasonCLIP-58M, a two-stage continual pretraining framework to train a family of CLIP-style visual encoders with over 58.9M visually grounded commonsense reasoning samples. This design progressively enhances reasoning capacity while explicitly preserving the model’s original descriptive alignment.


In the first stage, we incrementally introduce reasoning signals to preserve the model’s original descriptive alignment while enhancing its sensitivity to visually grounded inference. To support this, we construct ReasonLite-42M, which augments descriptive captions with visually verifiable reasoning statements.
In the second stage, we strengthen supervision with structured category-level signals using ReasonPro-16M, a dataset spanning five visually grounded reasoning types: \textit{Spatial/Geometric, Attribute/State, Creature/Action, Temporal/Phase, and Intuitive Physics}. This stage encourages the model to explicitly distinguish and organize diverse reasoning patterns within its visual representation space.
%
%
Using this progressive training strategy, we train CLIP and SigLIP variants at multiple scales to validate cross-architecture generality, resulting in ReasonCLIP and ReasonSigLIP models, respectively. To systematically evaluate reasoning ability, we introduce RCLIP-Bench, a diagnostic benchmark that decomposes visually grounded reasoning into three hierarchical levels: visual grounding, evidence awareness, and structured reasoning.

\begin{table}[t]
\centering
\caption{\textbf{Comparison of CLIP-style methods.} Existing CLIP improvements typically follow two directions: descriptive data engineering (Rows. 1-6) and architectural refinement (Rows. 7-11). ReasonCLIP-58M training framework enables reasoning-oriented, non-intrusive continual pretraining while preserving backbone compatibility. Post. and Cont. denote post-training and continual pre-training, respectively. KG. denotes Knowledge Graph.}
\label{tab:main: model_compare}
\setlength{\tabcolsep}{0.5pt}
\resizebox{\linewidth}{!}{
\begin{tabular}{l>{\centering\arraybackslash}p{2.5cm}
                >{\centering\arraybackslash}p{5cm}
                >{\centering\arraybackslash}p{3.5cm}
                >{\centering\arraybackslash}p{1.8cm}
                >{\centering\arraybackslash}p{1.3cm}
                >{\centering\arraybackslash}p{1.3cm}
                >{\centering\arraybackslash}p{1.3cm}
                >{\centering\arraybackslash}p{1.3cm}
                >{\centering\arraybackslash}p{4.7cm}}
\toprule
\multirow{2}{*}{Model} 
& \multirow{2}{*}{Training} 
& \multicolumn{4}{c}{Dataset} 
& \multicolumn{3}{c}{Non-Intrusive}
& \multirow{2}{*}{Supervision} 
\\
\cmidrule(lr){3-6} \cmidrule(lr){7-9}
& & Name / Source & Construction & Scale & Reason & Text & Vision & Extra& \\

\midrule
CLIP \cite{radford2021learning} & Pretrain & WIT& Web Crawl &  0.4B & \textcolor{darkred}{\ding{55}}  & - & - & - & Description \\

DataComp \cite{datacomp} & Pretrain & CommonPool&  Web Crawl & 1.4B & \textcolor{darkred}{\ding{55}}  & \textcolor{darkgreen}{\ding{51}} & \textcolor{darkgreen}{\ding{51}} & \textcolor{darkgreen}{\ding{51}} & Description \\

MetaCLIP \cite{xu2023demystifying,chuang2025meta} & Pretrain &MetaCLIP & Web Crawl & 2.5B & \textcolor{darkred}{\ding{55}}  & \textcolor{darkgreen}{\ding{51}} & \textcolor{darkgreen}{\ding{51}} & \textcolor{darkgreen}{\ding{51}} & Description \\

Eva-CLIP \cite{sun2023eva,eva02} & Pretrain & LAION & Web Crawl & 2.0B &\textcolor{darkred}{\ding{55}}    & \textcolor{darkgreen}{\ding{51}}  & \textcolor{darkred}{\ding{55}} & \textcolor{darkred}{\ding{55}} & Description \\

SigLIP \cite{zhai2023sigmoid, tschannen2025siglip} & Pretrain &WebLI& Web Crawl& 10.0B& \textcolor{darkred}{\ding{55}}   & \textcolor{darkgreen}{\ding{51}}  & \textcolor{darkgreen}{\ding{51}} & \textcolor{darkgreen}{\ding{51}}  &Description \\

PE-Core \cite{bolyaperception} & Pretrain &MetaCLIP, PVD & MLLM Anno. (PVD) & 5.4B  & \textcolor{darkred}{\ding{55}}  & \textcolor{darkgreen}{\ding{51}} & \textcolor{darkred}{\ding{55}}& \textcolor{darkred}{\ding{55}} & Description\\

\midrule
LongCLIP \cite{zhang2024long} & Post. & ShareGPT4V  & N/A & 1.0M &\textcolor{darkred}{\ding{55}}& \textcolor{darkred}{\ding{55}}& \textcolor{darkgreen}{\ding{51}}& \textcolor{darkred}{\ding{55}}  & Long-Description\\

CLIP-MoE \cite{zhang2025clip} & Post. &Recap-DataComp, ShareGPT4V& N/A &  2.2M  & \textcolor{darkred}{\ding{55}}& \textcolor{darkred}{\ding{55}}& \textcolor{darkred}{\ding{55}}& \textcolor{darkred}{\ding{55}}  & Mixture-of-Experts\\

LLM2CLIP \cite{huang2024llm2clip} & Cont. & DreamLIP, LAION, YMCC & N/A & 60.0M & \textcolor{darkred}{\ding{55}} & \textcolor{darkred}{\ding{55}} & \textcolor{darkgreen}{\ding{51}}& \textcolor{darkred}{\ding{55}} & LLM-Augmented\\


READ-CLIP \cite{kwon2025enhancing} & Post. & COCO & N/A &  0.1M  & \textcolor{darkred}{\ding{55}} & \textcolor{darkgreen}{\ding{51}} & \textcolor{darkgreen}{\ding{51}} & \textcolor{darkred}{\ding{55}}& Compositional Reasoning \\

DANCE \cite{ye2023improving} & Cont. & COCO, VG, SBU, CC3M, CC12M & KG. Anno. & 14.1M  & \textcolor{darkred}{\ding{55}} & \textcolor{darkgreen}{\ding{51}} & \textcolor{darkgreen}{\ding{51}} & \textcolor{darkgreen}{\ding{51}}& Knowledge Injection \\

\midrule
\textbf{ReasonCLIP} & Cont. & CC12M & MLLM Anno. &  58.9M & \textcolor{darkgreen}{\ding{51}} & \textcolor{darkgreen}{\ding{51}}& \textcolor{darkgreen}{\ding{51}} & \textcolor{darkgreen}{\ding{51}} & \textbf{Commonsense Reasoning} \\
\bottomrule
\end{tabular}
}
\end{table}

Extensive experiments demonstrate that ReasonCLIP-58M consistently improves performance on visually grounded and compositional reasoning benchmarks across multiple backbones and model scales. Notably, reasoning-oriented continual pretraining also yields consistent gains in zero-shot retrieval tasks, suggesting that CLIP’s representation space remains highly extensible. Furthermore, when integrated as a drop-in backbone into MLLM, such as LLaVA-NeXT \cite{liu2024llavanext}, ReasonCLIP's encoder delivers stable performance gains without additional inference overhead.
Our contributions are summarized as follows:
\setlist{nolistsep}
\begin{itemize}[noitemsep,leftmargin=*]

\item \textbf{Explicit visually grounded commonsense reasoning supervision.} We introduce ReasonCLIP‑58M, a continual pretraining framework and datasets that injects visually grounded commonsense reasoning into CLIP-style encoders while preserving descriptive alignment and architectural compatibility.

\item \textbf{Two-stage reasoning-aware continual pretraining.} We design a structured two-stage training scheme that decouples reasoning enhancement from descriptive alignment preservation and enables organized reasoning patterns without degrading the original representation space.

\item \textbf{Reasoning-enhanced CLIP/SigLIP variants}. We train ReasonCLIP and ReasonSigLIP models at multiple scales that consistently improve visually grounded and compositional reasoning as well as zero-shot retrieval.

\item \textbf{Comprehensive evaluation and RCLIP-Bench.} We evaluate on diverse reasoning tasks and introduce RCLIP-Bench, which diagnoses visual grounding, evidence awareness, and structured visual reasoning.





\end{itemize}

\section{Related Work}


\noindent \textbf{Supervision for CLIP.}\quad 
Most efforts to improve CLIP~\cite{radford2021learning} strengthen \textit{descriptive language supervision} for image-text alignment. A dominant line scales and filters large image-text corpora, including OpenCLIP~\cite{ilharco_gabriel_2021_5143773}, ALIGN~\cite{jia2021scaling}, EVA-CLIP~\cite{sun2024eva}, SigLIP~\cite{zhai2023sigmoid,tschannen2025siglip}, and MetaCLIP~\cite{xu2023demystifying,chuang2025meta}. 
Beyond data scaling, supervision is enhanced from both modalities: vision-side improvements leverage higher-quality video data or architectural refinements (e.g., Perception Encoder~\cite{bolyaperception}, EVA-CLIP~\cite{sun2024eva,sun2023eva,eva02}, Vitamin \cite{chen2024vitamin}), while text-side methods reduce caption noise, enrich descriptions, or strengthen text encoders (e.g., \cite{chen2024sharegpt4v,li2024if,liu2024clips,zheng2024dreamlip,zhang2024long,xie2025fg,wang2025fix,wang2025vitrix,koukounas2024jina,koukounas2024jinaclipv2multilingualmultimodalembeddings,jiang2024e5,chen2023altclip,huang2024llm2clip,zhang2024rca}). 
Despite these advances, CLIP remains primarily optimized for descriptive alignment rather than reasoning-oriented representation learning. Recent works \cite{sahin2024enhancing, zhang2024contrasting,yuksekgonul2023when,zheng2024iterated,basu2024distilling,herzig2023incorporating,oh2024preserving,li2021supervision}, including CLIP-Event~\cite{li2022clip}, TripletCLIP~\cite{patel2024tripletclip}, and READ-CLIP~\cite{kwon2025enhancing}, introduce structured supervision or auxiliary objectives to improve compositional reasoning, typically via fine-tuning, while methods such as DANCE~\cite{ye2023improving} incorporate external knowledge to enhance commonsense capabilities. However, large-scale visually grounded reasoning signals during pretraining remain underexplored.
\noindent \textbf{Benchmarks for CLIP.}\quad
The evaluation of CLIP and its variants spans multiple levels. Fundamentally, zero-shot performance is measured on standard classification benchmarks \cite{deng2009imagenet,recht2019imagenet,hendrycks2021natural,wang2019learning,barbu2019objectnet,xiao2010sun,fei2004learning,krizhevsky2009learning}, as well as zero-shot image-text retrieval benchmarks \cite{lin2014microsoft,young-etal-2014-image}. Recent evaluations further extend to long-form \cite{chen2024sharegpt4v,zhang2024long}, detailed \cite{onoe2024docci}, and multilingual \cite{thapliyal2022crossmodal} retrieval settings. 
For dense prediction and grounding tasks, CLIP features are commonly evaluated on semantic segmentation \cite{everingham2015pascal,zhou2017scene}, depth estimation \cite{silberman2012indoor,jampani2023navi}, and referring expression comprehension \cite{yu2016modeling,mao2016generation}. To assess higher-level reasoning capabilities, vision-and-language association is evaluated using WinoGAViL \cite{bitton2022winogavil}, while compositional reasoning is evaluated using benchmarks that test attribute, object, and relational substitution \cite{hsieh2023sugarcrepe,ma2023crepe,kamath2023whatsup,yuksekgonul2022and,thrush2022winoground,dumpala2024sugarcrepe++}.
However, most existing CLIP benchmarks assess descriptive alignment or compositional correctness without disentangling distinct failure modes in visually grounded reasoning, motivating the introduction of RCLIP-Bench for fine-grained diagnostic evaluation.\\
More detailed related works are discussed in Appendix \ref{appendix: Literature Review}. 

\section{ReasonCLIP-58M and RCLIP-Bench}
\begin{figure}[t]
        \centering
        \includegraphics[width=\linewidth]{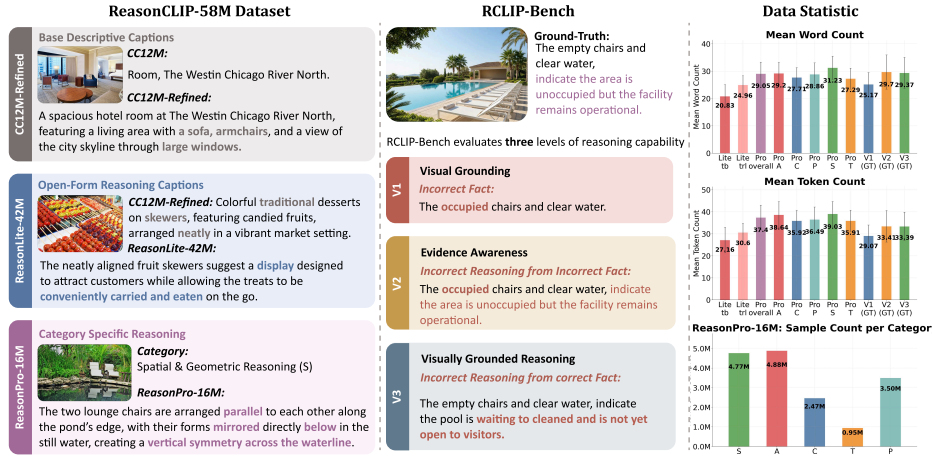}    
        \caption{
    \textbf{Overview of ReasonCLIP-58M \& RCLIP-Bench.}
    \textit{ReasonCLIP-58M} provides visually grounded reasoning supervision for CLIP-style models. Built from a refined version of \textit{CC12M}~\cite{changpinyo2021conceptual}, it comprises \textit{ReasonLite-42M} for open-form reasoning and \textit{ReasonPro-16M} for category-specific reasoning. We also introduce \textit{RCLIP-Bench}, a dedicated benchmark for three-level evaluation of visually grounded reasoning. We report the token and word distributions for each dataset.
    } 
        \label{fig:main_dataset_example}
    \vspace{-10pt}
\end{figure}

Existing reasoning datasets are primarily designed for 
MLLMs and involve complex reasoning, making them unsuitable for direct training of CLIP-style encoders. To bridge this gap, we introduce a novel two-stage training framework that fosters progressive, visually grounded reasoning. To facilitate this, we 
constructed a specialized suite of datasets (ReasonLite-42M and ReasonPro-16M)
that supports the distinct learning objectives of each stage and a new benchmark (RCLIP-Bench) that explicitly evaluates three hierarchical reasoning levels.
The overall design and representative examples are illustrated in Fig.~\ref{fig:main_dataset_example} with additional implementation details in Appendix~\ref{appendix: Dataset Cards}.

\label{Sec: Dataset}
\subsection{ReasonLite-42M Dataset: Open-Form Reasoning}

\textbf{ReasonLite} is an \emph{open-form reasoning} dataset containing diverse and verifiable reasoning statements grounded in visual evidence. These statements extend beyond factual descriptions and complement descriptive supervision with inference patterns that encourage reasoning-aware visual representations. As shown in Fig.~\ref{fig:main_dataset_example}, our reasoning statements leverage visually observable evidence (\textcolor{cliprGrey}{\textit{``neatly aligned''}}) in the image to infer higher-level design intentions and functional advantages (\textcolor{cliprBlue}{\textit{``conviently carried and eaten''}}). To ensure data quality, we exclude samples that exhibit false causality, over-extension, hallucination, or complex multi-step reasoning beyond what is visually supported. More details are shown in Appendix~\ref{appendix: Reasoning Level}.

\noindent \textbf{Dataset Construction.} \quad 
We use CC12M \cite{changpinyo2021conceptual} as the image source and retain 10.4M valid images. Due to the relatively low quality of the original CC12M captions, we employ Qwen2.5-VL-72B \cite{bai2025qwen2} to regenerate three high-quality descriptive texts per image, ensuring factual consistency and descriptive completeness. We denote these regenerated captions as \textbf{\textit{\textcolor{cliprGrey}{Text Base}}} ($T_b$) and refer to the resulting dataset as \textit{CC12M-Refined}. This step yields 31.2M relabeled image-text pairs. From CC12M-Refined, 4.7M images are used to construct ReasonLite, while the remaining 5.7M are reserved for ReasonPro.

We then input each image and its corresponding $T_b$ into the annotation model and instruct it to extract key visual evidence grounded in $T_b$, producing an evidence set $E$. Within the same generation request, the model partitions $E$ into three semantically coherent clusters $\{E_1, E_2, E_3\}$. For each cluster $E_i$, the model performs commonsense reasoning based on these cues and generates a corresponding caption \textbf{\textit{\textcolor{cliprBlue}{Text Reason Lite}}} ($T_{rl}$). The $T_{rl}$ captions express visually verifiable commonsense inferences derived from the grouped evidence, and the prompts explicitly constrain all reasoning statements to remain grounded in the image and consistent with $T_b$.
In total, this process yields 42M image-caption pairs in \textbf{ReasonLite-42M}. Additional construction details, including statistics, prompt design, and computational cost, are provided in Appendices~\ref{appendix: CC12M-Enhanced} and~\ref{appendix: ReasonLite-42M}.

\subsection{ReasonPro-16M Dataset: Category-Specific Reasoning}

\textbf{ReasonPro} introduces category-specific commonsense reasoning captions to provide more structured and controllable supervision than the open-form generation in ReasonLite. While ReasonLite encourages diverse reasoning patterns, ReasonPro explicitly targets predefined visual reasoning types, enabling clearer supervision signals and fine-grained per-category analysis aligned for CLIP.

\begin{table}[t]
\centering
\caption{Five categories of perception-grounded reasoning types in ReasonPro-16M.}
\setlength{\tabcolsep}{6pt}
\resizebox{\linewidth}{!}{
\begin{tabular}{clp{15cm}}
\toprule
\textbf{ID} & \textbf{Category} & \textbf{Definition} \\
\midrule
S & \textbf{Spatial / Geometric}   & Reasoning about spatial relations (position, direction, distance, occlusion, containment) and how object placement or geometry affects visibility, reachability, or motion. \\
\midrule
A & \textbf{Attribute / State}     & Reasoning about intrinsic properties or visible conditions of objects, including appearance, texture, deformation, openness, or on/off states, based on visual cues. \\
\midrule
C & \textbf{Creature / Action }       & Reasoning about human or animal posture, gesture, and interaction with objects to infer current or imminent behavior. \\
\midrule
T & \textbf{Temporal / Phase}      & Reasoning about the temporal stage of an event (past, ongoing, or upcoming) based on motion continuity, trajectory, or dynamic context. \\
\midrule
P & \textbf{Intuitive Physics}    & Reasoning about intuitive physical relationships (stability, support, contact, forces) inferred directly from visual cues, without relying on semantic common sense. \\
\bottomrule
\end{tabular}}
\label{tab:reasoning_categories}
\vspace{-10pt}
\end{table}

\noindent \textbf{Reasoning Categories.}\quad
In Table~\ref{tab:reasoning_categories}, we define five visually grounded reasoning patterns compatible with CLIP-style encoders: \textit{Spatial/Geometric (S), Attribute/State (A), Creature/Action (C), Temporal/Phase (T), and Intuitive Physics (P)}. For each category, captions are constructed from category-specific descriptive cues such that the inferred commonsense reasoning remains explicitly grounded in the corresponding visual attribute, relation, action, temporal stage, or physical condition. Consistent with ReasonLite, all reasoning in ReasonPro is grounded in factual descriptions and observable visual evidence. Detailed definitions and examples for each category are provided in Appendix~\ref{Appendix:Reasoning Categories Definition}.

\noindent \textbf{Dataset Construction.}\quad
All annotations in this stage are conducted with Qwen3-VL-32B~\cite{qwen3technicalreport}, which provides stronger category discrimination and more structured outputs than the model used for ReasonLite. Starting from the 5.7M images reserved from CC12M-Refined, we first perform a multi-label category annotation stage. Given the predefined reasoning categories and their definitions, the annotation model assigns a label set $Y \subseteq \{S, A, C, T, P\}$ to each image, indicating the visually supported reasoning types. To ensure that each image supports multiple reasoning patterns and yields richer supervision, we retain only images with $|Y| \geq 3$, yielding 5.5M images for ReasonPro construction.

For each retained image, we randomly select three reasoning categories $Y' \subseteq Y$. We then feed the image, its corresponding $T_b$, the selected categories $Y'$, and the category definitions into the annotation model within a single generation request. The model generates one category-specific reasoning caption for each selected category, producing three \textbf{\textit{\textcolor{cliprPurple}{Text Reason Pro}}} ($T_{rp}$) samples per image.
In total, ReasonPro contains 5.5M images and 16.6M image-caption pairs spanning up to five reasoning categories. Further details are provided in Appendix~\ref{appendix: ReasonPro-16M}.

\subsection{RCLIP-Bench: Visually Grounded Reasoning Benchmark}

\textbf{RCLIP-Bench} is a benchmark for fine-grained diagnostic evaluation of visually grounded commonsense reasoning. Although existing benchmarks effectively measure image-text alignment using descriptive captions \cite{lin2014microsoft,young-etal-2014-image,subramanian2022reclip}, they are insufficient for assessing visually grounded reasoning. 
RCLIP-Bench evaluates three levels of reasoning capability: \textit{Visual Grounding} (V1), \textit{Evidence Awareness} (V2), and \textit{Visually Grounded Reasoning} (V3). This hierarchical design enables precise diagnosis of where a model’s reasoning fails.

\noindent\textbf{Benchmark Construction.} \quad 
RCLIP-Bench is constructed from high-quality image–caption pairs in the \textbf{DOCCI} dataset~\cite{onoe2024docci}, which provides detailed human-authored descriptions serving as reliable factual references. For each image, the original DOCCI caption is treated as the positive instance. We then generate three types of hard negatives corresponding to V1-V3.
The benchmark follows the same five reasoning categories defined in ReasonPro-16M (S/A/C/T/P). As illustrated in Fig.~\ref{fig:main_dataset_example}, RCLIP-Bench adopts a three-tier hierarchy of hard negatives, each targeting a distinct failure mode:

\setlist{nolistsep}
\begin{itemize}[noitemsep,leftmargin=*]
    \item \textbf{Incorrect Facts (V1).} These captions contain factually inconsistent descriptions and directly test visual grounding ability. Within V1, errors are divided into five sub-types: \textit{subject/object form}, \textit{noun substitution}, \textit{relational misuse}, \textit{attribute confusion}, and \textit{sentence structure alteration}. Failure at this level indicates insufficient perceptual accuracy.

    \item \textbf{Incorrect Reasoning from Incorrect Facts (V2).} These captions begin with an incorrect factual premise and derive flawed reasoning from it. It evaluates evidence awareness: whether the model can detect incorrect premises rather than being misled by a coherent but visually unsupported narrative.

    \item \textbf{Incorrect Reasoning from Correct Facts (V3).} These captions describe the scene factually correctly but draw an incorrect reasoning. This level tests whether a model can distinguish valid commonsense reasoning from invalid conclusions based on the same visual facts.

\end{itemize}
Evaluation is conducted in a contrastive retrieval setting, where the model must rank the correct caption above its corresponding hard negatives.

\subsection{Quality Control}

We adopt a two-stage quality-control process to ensure the reliability of the dataset. Prior work suggests that large-scale training is relatively robust to moderate noise \cite{jia2021scaling}. Therefore, our focus is on preventing systematic errors and generation bias rather than eliminating all minor imperfections.

\noindent\textbf{Manual Validation.} \quad Before large-scale generation, we conduct a pilot validation round with 500 samples per stage. Each sample is independently reviewed by five graduate students. Prompts are iteratively refined, and large-scale generation proceeds only after the pass rate exceeds 99.5\%.

\noindent\textbf{Automatic Filtering.} \quad For ReasonLite, we apply rule-based filtering to remove overly long, short, malformed, or degenerate generations, with a removal rate below 0.01\textperthousand. In ReasonPro, 3.97\% (0.20M) of samples are filtered during category annotation due to inconsistent or low-confidence labels, and additional structural and grounding constraints are enforced during caption generation. After large-scale generation, we randomly inspect 500 samples again, achieving a pass rate above 99.0\% across all stages.
For RCLIP-Bench, quality control ensures clear separation and difficulty stratification across the three negative tiers (V1/V2/V3). Negative captions are generated with GPT-5 \cite{singh2025openai} under tier-specific prompts, followed by filtering to remove degenerate outputs, near-duplicates, and tier violations, resulting in a removal rate of approximately 13\%. Additional quality control details are provided in Appendix~\ref{appendix: Quality Control}.

\section{ReasonCLIP Training Framework}
\label{Sec: Methodology}

\begin{figure}[t]
    \centering
    \includegraphics[width=\linewidth]{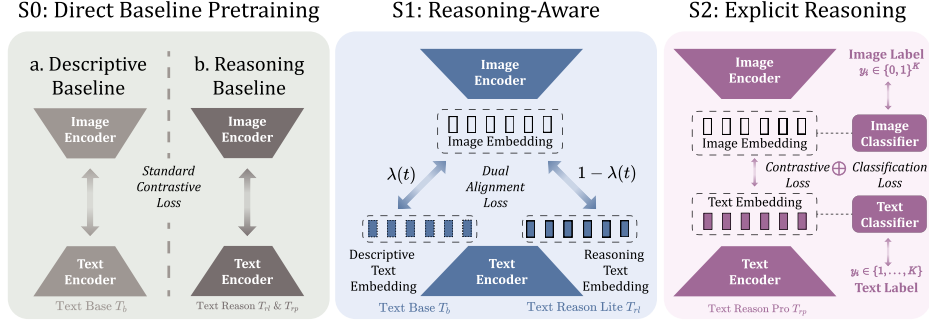}
    \caption{\textbf{Overview of the ReasonCLIP training framework.} ReasonCLIP uses a stage-wise continual pretraining strategy that progressively integrates reasoning capacity into the visual encoder while preserving its foundational semantic alignment, enabling seamless integration as a drop-in backbone in MLLMs.}
    \label{fig:training_framework}
\end{figure}

We propose a two-stage continual pretraining framework that integrates visually grounded reasoning into CLIP-style visual encoders while preserving descriptive alignment. As shown in Fig.~\ref{fig:training_framework}, training consists of three stages: baseline pretraining (Stage 0), reasoning-aware alignment (Stage 1), and explicit category-level reasoning supervision (Stage 2). We further demonstrate drop-in integration of the resulting ReasonCLIP encoder into downstream multimodal systems (Stage 3), enabling stable adaptation without additional inference cost.

\subsection{Stage 0: Baseline Continual Pretraining}

\textbf{Stage 0} establishes two baseline pretraining setups for comparison: purely descriptive alignment (\textit{S0-Des.}) and naive reasoning supervision (\textit{S0-Rea.}). 

In the \textit{S0-Des.} model, only descriptive captions $T_b$ from CC12M-Refined are used for training. In the \textit{S0-Rea.} model, all reasoning captions (including $T_{rl}$ from ReasonLite and $T_{rp}$ from ReasonPro) are directly mixed as training targets, without distinguishing reasoning types or stages.
Both settings optimize the same standard image-text alignment objective. Formally, given an image-text pair $(x, t)$, the model is optimized with a backbone-specific alignment loss:
\vspace{-4pt}
\begin{equation}
\mathcal{L}^{(0)} = \mathcal{L}_{\text{align}}(x, t),
\vspace{-4pt}
\end{equation}
where $\mathcal{L}_{\text{align}}$ denotes the backbone-specific image–text alignment objective. For example, CLIP-style models adopt the symmetric InfoNCE loss, while SigLIP models use a sigmoid-based binary cross-entropy loss.

\subsection{Stage 1: Reasoning-Aware Alignment}

\textbf{Stage 1} introduces visually grounded commonsense reasoning awareness into the CLIP-style model while preserving its original descriptive alignment.

We continue pretraining on the ReasonLite-42M dataset using both descriptive captions $T_b$ and reasoning captions $T_{rl}$ as supervision signals. This dual-supervision objective maintains the original alignment structure while gradually incorporating reasoning signals, encouraging the model to become sensitive to reasoning cues without disrupting its learned visual semantics. To prevent excessive parameter drift from the original backbone, we introduce an $\ell_2$ regularization term toward the initial weights.

Formally, for a given image $x$ with descriptive caption $T_b$ and reasoning caption $T_{rl}$, the Stage 1 objective is defined as
\vspace{-5pt}
\begin{equation}
\mathcal{L}^{(1)}
=
\lambda(t)\,\mathcal{L}_{\text{align}}(x, T_b)
+
\bigl(1-\lambda(t)\bigr)\,\mathcal{L}_{\text{align}}(x, T_{rl})
+
\beta \|\theta - \theta_0\|_2^2,
\vspace{-5pt}
\end{equation}
where $\mathcal{L}_{\text{align}}$ denotes the backbone-specific image–text alignment loss, $\lambda(t)$ is a time-dependent weight at training step $t$, $\theta$ and $\theta_0$ represent the current and original model parameters, respectively, and $\beta$ controls the  regularization.

To gradually shift emphasis toward reasoning supervision, $\lambda(t)$ is defined as
\vspace{-5pt}
\begin{equation}
\lambda(t) = 
\lambda_{\min} 
+ 
(\lambda_{\max} - \lambda_{\min})
\cdot 
\mathrm{clip}\!\left(
\frac{t - t_1}{t_2 - t_1},\; 0,\; 1
\right).
\vspace{-5pt}
\end{equation}
In practice, $\lambda(t)$ decreases over time, progressively increasing the influence of reasoning captions while retaining descriptive alignment as the dominant signal in early training. This scheduling enables the smooth integration of reasoning signals within a unified representation space 

\subsection{Stage 2: Explicit Reasoning Supervision}

\textbf{Stage 2} builds upon the reasoning awareness established in Stage 1 and shifts the objective toward explicit category-level reasoning supervision.

To this end, Stage 2 training is conducted on the ReasonPro-16M dataset, where descriptive captions $T_b$ are no longer used as supervision. The primary objective is to align images with category-specific reasoning captions $T_{rp}$. Since alignment alone does not explicitly encourage discriminability across different reasoning categories, we introduce a category-discrimination branch that imposes category-level supervision, thereby enhancing the separability of distinct reasoning patterns while maintaining a unified feature space.

Formally, for a given image $x$ with corresponding reasoning caption $T_{rp}$, let $Y \subseteq \{S, A, C, T, P\}$ denote the multi-label reasoning categories associated with image $x$, and let $c$ denote the single reasoning category of caption $T_{rp}$.
The overall optimization objective in Stage 2 is
\vspace{-5pt}
\begin{equation}
\mathcal{L}^{(2)}
=
\mathcal{L}_{\text{align}}(x, T_{rp})
+
\gamma\bigl(\mathcal{L}_{\text{img-cls}}(x, Y)
+
\mathcal{L}_{\text{txt-cls}}(T_{rp}, c)\bigr),
\vspace{-5pt}
\end{equation}
where $\mathcal{L}_{\text{img-cls}}$ and $\mathcal{L}_{\text{txt-cls}}$ denote the image-side multi-label and text-side single-label classification losses, respectively; $\gamma$ balances alignment and category-level supervision. These losses are defined as
\vspace{-5pt}
\begin{equation}
\mathcal{L}_{\text{img-cls}}
=
\mathrm{BCE}\bigl(g_v(x), \mathbf{y}\bigr),
\quad
\mathcal{L}_{\text{txt-cls}}
=
\mathrm{CE}\bigl(g_t(T_{rp}), c\bigr),
\vspace{-5pt}
\end{equation}
with $\mathbf{y} \in \{0,1\}^5$ the multi-hot reasoning label vector of image $x$, and $c \in \{1,\dots,5\}$ the reasoning category of caption $T_{rp}$. The functions $g_v(\cdot)$ and $g_t(\cdot)$ are lightweight MLP-based classification heads attached to the image and text encoders, respectively, and are used only during training (discarded at inference).

\subsection{Stage 3: Drop-in Integration}

\textbf{Stage 3} demonstrates the integration of ReasonCLIP into downstream multimodal systems. We apply the proposed continual pretraining framework to a CLIP-style backbone and replace the original visual encoder with the resulting ReasonCLIP encoder. The target system then proceeds with its standard alignment or fine-tuning procedure.
Formally, let a multimodal system be denoted as $\mathcal{M}$ with visual encoder $\phi_{\text{Baseline}}$. The integration can be expressed as
\vspace{-5pt}
\begin{equation}
\mathcal{M}(x, \cdot)
=
\mathcal{F}\bigl(\phi_{\text{Baseline}}(x), \cdot\bigr)
\longrightarrow
\mathcal{F}\bigl(\phi_{\text{ReasonCLIP}}(x), \cdot\bigr),
\vspace{-5pt}
\end{equation}
where $\mathcal{F}$ denotes the remaining components of the multimodal system. This replacement enables downstream systems to leverage reasoning-aware visual representations without altering their existing pipelines.

\section{Experiments}
\label{Sec: Experiments}

\subsection{Implementation Details}
\textbf{Training Setting.}  \quad Data generation and training are conducted on NVIDIA A100 64G GPUs, requiring around 3.8k and 3.5k GPU hours, respectively.
Following Sec.~\ref{Sec: Methodology}, each stage is trained for one epoch on its corresponding dataset with an effective batch size of 24{,}576 or 32{,}768, depending on the model scale.
We train six CLIP~\cite{radford2021learning} and SigLIP~\cite{zhai2023sigmoid,tschannen2025siglip} variants at different scales.
More detailed training configurations for each stage are provided in Appendix~\ref{appendix: Training Details and More Results}.


We integrate both CLIP-L/14-336 and ReasonCLIP-L/14-336 into the LLaVA NeXT~\cite{liu2024llavanext} framework with Qwen3-1.7B~\cite{qwen3technicalreport} as the language backbone, keeping the vision tower frozen during training.
Following the standard training pipeline, we use 558K samples for pretraining and 779K samples for fine-tuning.

\noindent \textbf{Baselines.}  \quad
We adopt a tiered comparison strategy.
First, we compare against the original backbones (CLIP and SigLIP) to measure performance gains from reasoning-oriented continual pretraining.
Second, we include data-centric methods such as MetaCLIP~\cite{xu2023demystifying} and DataComp~\cite{datacomp}.
We further compare with approaches involving limited structural modifications (e.g., Long-CLIP~\cite{zhang2024long} and Eva-CLIP-02 \cite{eva02}), as well as compositional reasoning methods such as READ-CLIP~\cite{kwon2025enhancing}.
As a non-intrusive framework, to ensure fair comparison, we exclude large-scale architectural or LLM-augmented approaches such as EVA-CLIP-18B \cite{sun2024eva}, PE-Core~\cite{bolyaperception}, and LLM2CLIP~\cite{huang2024llm2clip}.

\noindent \textbf{Evaluation.}  \quad
We re-evaluate all models under a unified protocol based on CLIP-bench~\cite{cherti_clip_bench}.
Please refer to each result section and Appendix~\ref{Appendix: Detailed Evaluation Settings} for details.

\begin{table*}[t]
\centering
\caption{\textbf{Zero-Shot Text-Image Retrieval Results.} $^{*}$All 31K images are used.} 
\label{tab:main: Text-Image Retrieval}
\setlength{\tabcolsep}{2pt}
\resizebox{\linewidth}{!}{
\begin{tabular}{lcc
|cc|cc     
|ccc|ccc
|cc|cc
|ccc|ccc}    
\toprule
& & &
\multicolumn{4}{c|}{COCO-5K \cite{lin2014microsoft}} &
\multicolumn{6}{c|}{Flickr-30K \cite{young-etal-2014-image} } &
\multicolumn{4}{c|}{Urban-1K \cite{zhang2024long}} &
\multicolumn{6}{c}{RCLIP-V3-5K \textbf{(Ours)}} \\
Model & Res. & Data & 
\multicolumn{2}{c}{I $\rightarrow$ T} & \multicolumn{2}{c|}{T $\rightarrow$ I} &
\multicolumn{3}{c}{I $\rightarrow$ T} & \multicolumn{3}{c|}{T $\rightarrow$ I} &
\multicolumn{2}{c}{I $\rightarrow$ T} & \multicolumn{2}{c|}{T $\rightarrow$ I} &
\multicolumn{3}{c}{I $\rightarrow$ T} & \multicolumn{3}{c}{T $\rightarrow$ I} \\
& & &
R@1 & R@5  & R@1 & R@5 &
R@1 & R@5 & R@10 & R@1 & R@5 & R@10 &
R@1 & R@5 & R@1 & R@5 &
R@1 & R@5 & R@10 & R@1 & R@5 & R@10 \\
\bottomrule
\rowcolor{gray!15}
\multicolumn{23}{l}{\textit{Scale - ViT Base (86M)}} \\
 OpenCLIP \cite{ilharco_gabriel_2021_5143773} & 224{\scriptsize/32} & 0.4B
& 52.3 & 76.3   & 34.2  &  60.0       
& 39.7  & 63.8  &  72.8 &  24.0 &  44.0 & 53.3
&  55.8 &  78.1  & 55.4 & 77.9
& 54.8 & 77.8 & 84.7 & 27.8 & 47.2 & 54.5
\\

MetaCLIP \cite{xu2023demystifying} & 224{\scriptsize/32} & 0.4B
& 51.8  & 76.4     & 35.9  & 61.8   
& 39.3  &63.2   & 72.4  &  25.7 & 46.7  & 56.2
& 57.2  &79.7  & 52.6   &  73.7
& \underline{56.5} & \underline{80.6} & 87.6 & 30.1 & 50.2 & 57.8
\\

DataComp \cite{datacomp} & 224{\scriptsize/32} & 1.4B
&  \underline{53.4} & \underline{77.5}   & \underline{37.2} & 62.3   
& 39.0  &62.7 & 71.9 & 24.6 &  44.9 & 54.1
& \underline{64.4}  & \underline{85.9}  & 59.9   & \underline{88.3}
& \textbf{60.8} & \textbf{82.6} & \textbf{88.8} & \underline{32.2} & \underline{52.0} & 59.3
\\

CLIP \cite{radford2021learning} & 224{\scriptsize/32} & 0.4B
 & 50.0 & 75.0  & 30.4   & 56.0 
 & 40.7 & 64.7  & 73.7  & 21.8  & 41.6  &51.1
&  61.0 & 84.8 & 46.8 & 72.8
& 51.0 & 75.3 & 84.1 & 26.7 & 47.2 & 55.5
\\

\rowcolor{gray!7}
\textbf{\textcolor{cliprBlue}{+ Stage 1}} & \textcolor{cliprBlue}{224{\scriptsize/32}} & \textbf{\textcolor{cliprBlue}{+42M}}
 & \textbf{56.2} &  \textbf{78.8}   & \textbf{37.9}  & \textbf{64.1} 
 & \textbf{42.5} & \textbf{68.0}  & \textbf{77.2}  & \textbf{29.1}  & \textbf{51.1}  & \textbf{60.7}
 & \textbf{70.4} & \textbf{91.6} & \textbf{68.6} & \textbf{90.2}
 &  56.0 & 79.3 & 87.0 & 30.4 & 51.5 & \underline{59.4}
 
\\

\rowcolor{gray!7}
\textbf{\textcolor{cliprPurple}{+ Stage 2}} & \textcolor{cliprPurple}{224{\scriptsize/32}} & \textbf{\textcolor{cliprPurple}{+16M}}
&  52.3 & 77.2   & 37.0  & \underline{62.6}  
& \underline{41.9} & \underline{66.4} & \underline{75.7} & \underline{27.5} & \underline{49.0} & \underline{58.4} 
& 59.2 & 83.0 & \underline{60.4} & 83.9
& 55.3 & 80.5 & \underline{88.0} & \textbf{33.8} & \textbf{56.0} & \textbf{64.0}
\\
\bottomrule

\rowcolor{gray!15}
\multicolumn{23}{l}{\textit{Scale - ViT Large (307M)}} \\
 OpenCLIP \cite{ilharco_gabriel_2021_5143773} & 224{\scriptsize/14} & 2.0B
& 63.4  & 84.0  & 46.5 &  71.1 
& 55.0  & 79.0  & 86.4  & 39.4 & 62.2 & 70.9
&  72.0 &  90.7   &  68.0  &   89.0
& 67.1 & 87.3 & 92.8 & 37.3 & 56.8 & 63.0
\\

 EVA-CLIP-02 \cite{sun2023eva}  & 224{\scriptsize/14} & 2.0B
& 63.7  & 84.3  & \textbf{47.5} & 71.2
& \underline{58.2}  & \underline{81.0}  & \underline{87.7}  & \textbf{42.2}  & \textbf{64.7}  & \textbf{73.0}
& 73.4 & 91.0 & 70.0  & 87.0
& \underline{67.6} & 87.6 & 92.8 & 38.3 & 58.0 & 64.1
\\

MetaCLIP \cite{xu2023demystifying} & 224{\scriptsize/14}  & 2.5B
& \underline{64.4}  &  \underline{85.0}  & \underline{47.1}  & 71.4 
& 57.1  & 80.4  &  87.6 &  40.9 &  63.8 & 72.3
& 74.4  &  89.5   &  69.9  &   85.9
&  \textbf{69.1} & \underline{88.9} & \underline{93.6} & \underline{38.9} & 58.3 & 64.4
\\

Long-CLIP~\cite{zhang2024long} & 224{\scriptsize/14} & 0.4B
&  63.3 & 84.7 & \underline{47.1}  & \textbf{71.9}  
& 54.5  & 78.8  & 86.4  & \underline{41.5}  & \underline{64.4}  & \underline{72.9}
&  \textbf{82.8} &  \textbf{96.6} &  \textbf{86.1}  & \textbf{96.4}
&  59.7 & 83.8 & 90.6 & 37.7 & 58.0 & 64.7
\\

CLIP \cite{radford2021learning}   & 224{\scriptsize/14}  & 0.4B
&  56.3 & 79.5  & 36.6  & 61.2  
&  48.8 &  73.0 & 81.1  & 28.2  &  49.6 & 59.0
& 68.5 & 88.9 & 55.9 & 80.4
& 54.0 & 77.9 & 86.1 & 31.7 & 52.3 & 59.5
\\

\rowcolor{gray!7}
  \textbf{\textcolor{cliprBlue}{+ Stage 1}} & \textcolor{cliprBlue}{224{\scriptsize/14}}& \textbf{\textcolor{cliprBlue}{+42M}}
& \textbf{64.5}  & \textbf{85.4}  & 46.7  & \underline{71.6}  
&  \textbf{59.9} & \textbf{82.4}  &  \textbf{88.8} & 40.9  & 63.7  & 72.4
&  \underline{80.5} & \underline{96.2} & \underline{79.7} & \underline{94.2} 
& 66.3 & 86.2 & 92.1 & 38.0 & \underline{58.4} & \underline{65.1}
\\
\rowcolor{gray!7}
 \textbf{\textcolor{cliprPurple}{+ Stage 2}} & \textcolor{cliprPurple}{224{\scriptsize/14}} & \textbf{\textcolor{cliprPurple}{+16M}}
& 59.9  & 82.5   & 46.2 & 70.8 
& 55.6 & 78.8 & 85.8 & 39.0 & 61.8 & 70.5
& 74.3 & 92.9 & 75.5 & 91.5
& 67.1 & \textbf{89.1} & \textbf{94.1} & \textbf{41.9} & \textbf{63.1} & \textbf{69.3}
\\
\midrule

ViTamin \cite{chen2024vitamin} & 336{\scriptsize/16} & 1.0B
& 64.3 & 85.0 & 47.3 & 71.0
& 56.1 & 79.6 & 86.6 & 39.9 & 62.3 & 70.8 
& \underline{80.6}& \underline{93.7}& 76.0& 93.2
& \textbf{69.7} & \underline{88.3} & 92.8 & 39.8 & 58.5 & 64.1\\

EVA-CLIP-02  \cite{sun2023eva}   & 336{\scriptsize/14} & 2.0B
& 64.2  & 85.2   & \underline{47.9}  & \underline{71.7}  
& 59.7  & 82.1  & 88.7  & \underline{43.4}  &   \underline{66.0}& \underline{74.1}
& 77.1 & 91.9 & 70.7 & 87.8
& 67.9 & 88.0 & \underline{93.0} & 39.1 & 58.7 & 64.6
\\


SigLIP2 \cite{zhai2023sigmoid}& 384{\scriptsize/16} & 10.0B
& \textbf{70.9} & \textbf{89.6}  & \textbf{55.5} & \textbf{78.1} 
& \textbf{68.6} & \textbf{88.3} & \textbf{93.1} & \textbf{51.2} & \textbf{73.1} & \textbf{80.2}
&66.9 & 86.4 & 61.9 & 80.3
&66.5 & 82.3 & 86.9 & \underline{41.8} & \underline{59.5} & \underline{65.4}
\\

CLIP \cite{radford2021learning} & 336{\scriptsize/14} & 0.4B
& 58.0 & 80.9    & 37.0  & 61.6  
&  51.4 & 75.5  & 83.3  & 31.6  & 53.5  & 62.7
& 73.0 & 90.3 & 57.0 & 79.5
& 55.9 & 79.5 & 86.6 & 33.2 & 53.6 & 60.7
\\

\rowcolor{gray!7}
  \textbf{\textcolor{cliprBlue}{+ Stage 1}} & \textcolor{cliprBlue}{336{\scriptsize/14}} & \textbf{\textcolor{cliprBlue}{+42M}}
& \underline{65.1} & \underline{85.7}   &  46.9 & \underline{71.7}  
&  \underline{61.0} &  \underline{83.0} & \underline{89.5}  &  41.9 &  64.8 & 73.2
& \textbf{83.0} & \textbf{96.3} & \textbf{81.9} & \textbf{94.5}
&  67.2 & 86.7 & 92.7 & 38.5 & 58.7 & 65.3
\\
\rowcolor{gray!7}
 \textbf{\textcolor{cliprPurple}{+ Stage 2}} & \textcolor{cliprPurple}{336{\scriptsize/14}} & \textbf{\textcolor{cliprPurple}{+16M}}
& 61.3  & 83.0   &  47.1 & 71.3  
& 57.2 & 79.8 & 86.8 & 40.8 & 63.4 & 72.0
& 75.5 & 91.8 & \underline{78.2} & \underline{93.5}
& \underline{68.0} & \textbf{89.9} & \textbf{95.0} & \textbf{42.8} & \textbf{63.4} & \textbf{69.8}
\\

\bottomrule

\rowcolor{gray!15}
\multicolumn{23}{l}{\textit{Scale - ViT So400m (428M)}} \\
SigLIP \cite{zhai2023sigmoid} & 384{\scriptsize/14} & 10.0B
& 72.6 & 90.1 & 54.3 & 76.8 
& 69.9 & 89.3 & 93.9 & 51.3 & 73.0 & 80.2
& 74.5 & 91.9 & 73.4 & 88.3
& \underline{70.5} & \underline{86.3} & \textbf{90.2} & 43.4 & 61.6 & 66.8
\\
\rowcolor{gray!7}
 \textbf{\textcolor{cliprBlue}{+ Stage 1}} & 384{\scriptsize/14} & \textbf{\textcolor{cliprBlue}{+42M}}
& \underline{73.6} & \textbf{91.2} & 56.5 & 79.2 
& 69.3 & 88.8 & 93.8 & 52.3 & 73.8 & 80.9
& \underline{77.5} & \textbf{93.1} & \textbf{82.3} & \textbf{94.5}
& 68.3 & 84.4 & 87.7 & \underline{45.5} & \underline{63.7} & \underline{69.3}
\\
\rowcolor{gray!7}
 \textbf{\textcolor{cliprPurple}{+ Stage 2}} & 384{\scriptsize/14} & \textbf{\textcolor{cliprPurple}{+16M}}
& \textbf{73.7} & \underline{90.8}  & 57.3 & 79.9
& \underline{73.3} & \underline{90.0} & \underline{94.3} & \underline{53.9} & \underline{75.3} & \underline{82.3}
& \textbf{78.6} & 92.3 & \underline{80.6} & \underline{93.8}
& \textbf{72.4} & \textbf{86.4} & \underline{89.5} & \textbf{49.5} & \textbf{67.8} & \textbf{73.0}
\\

SigLIP2 \cite{tschannen2025siglip} & \textcolor{cliprPurple}{384{\scriptsize/14}} & 10.0B
& 71.2 & 89.7& 55.7 & 78.5 
& 69.9 & 89.3 & 93.9 & 51.3 & 73.0 & 80.2
& 64.1 & 84.7 & 65.0 & 84.1
& 64.9 & 78.2 & 80.7 & 38.5 & 54.8 & 60.0
\\
\rowcolor{gray!7}
 \textbf{\textcolor{cliprBlue}{+ Stage 1}} & 384{\scriptsize/14} & \textbf{\textcolor{cliprBlue}{+42M}}
& 72.8 & 89.8  & \underline{58.2} & \underline{80.4} 
& 69.6 & 89.0 & 93.6 & 53.6 & 74.8 & 81.7
& 75.0 & \underline{92.8} & 75.5 & 91.0
&  64.5 & 77.7 & 81.0 & 40.5 & 56.7 & 61.8
\\
\rowcolor{gray!7}
 \textbf{\textcolor{cliprPurple}{+ Stage 2}} &\textcolor{cliprPurple}{384{\scriptsize/14}} & \textbf{\textcolor{cliprPurple}{+16M}}
& 73.3 & 89.9  & \textbf{59.0} & \textbf{81.1} 
& \textbf{73.8} & \textbf{91.2} & \textbf{94.9} & \textbf{54.8} & \textbf{76.2} & \textbf{83.1}
& 73.9 & 91.7 & 75.5 & 92.3
& 68.4 & 80.5 & 82.7 & 43.8 & 60.6 & 65.2 
\\

\bottomrule
\rowcolor{gray!15}
\multicolumn{23}{l}{\textit{Scale - ViT Giant Opt (1B)}} \\
SigLIP2 \cite{tschannen2025siglip} & 384{\scriptsize/16} & 10.0B
& \underline{72.4} & \textbf{90.6}& 56.5 & 78.6 
& 69.0 & 89.0 & 93.7 & 53.0 & 74.4 & 81.3
& 58.6 & 78.3 & 58.2 & 78.2
& \underline{65.2} & \underline{79.5} & \textbf{83.1} & 40.0 & 57.0 & 62.5
\\
\rowcolor{gray!7}
 \textbf{\textcolor{cliprBlue}{+ Stage 1}} & \textcolor{cliprBlue}{384{\scriptsize/16}} & \textbf{\textcolor{cliprBlue}{+42M}}
& 72.0 & \underline{89.5}  & \underline{59.3} & \underline{81.3}
& \underline{71.4} & \underline{90.1} & \underline{94.5} & \underline{55.5} & \underline{76.6} & \underline{83.3}
& \textbf{72.2} & \textbf{92.9} & \textbf{77.9} & \textbf{92.3}
&  65.1 & 78.4 & 81.1 & \underline{43.6} & \underline{61.1} & \underline{66.6}
\\
\rowcolor{gray!7}
 \textbf{\textcolor{cliprPurple}{+ Stage 2}} & \textcolor{cliprPurple}{384{\scriptsize/16}} & \textbf{\textcolor{cliprPurple}{+16M}}
& \textbf{73.7} & 89.3 & \textbf{60.1} & \textbf{81.7}
& \textbf{75.3} & \textbf{91.6} & \textbf{95.3} & \textbf{56.7} & \textbf{77.5} & \textbf{84.0}
& \underline{71.0} & \underline{90.4} & \underline{75.5} & \underline{90.0}
& \textbf{67.4} & \textbf{79.7} & \underline{81.9} & \textbf{47.2} & \textbf{64.8} & \textbf{70.2}
\\

\bottomrule
\end{tabular}}
\end{table*}


\subsection{Results}

\noindent \textbf{Observation $\textbf{1.}$}  \quad \textit{ReasonCLIP consistently improves model performance on both descriptive retrieval and reasoning retrieval tasks across backbones and model scales.} In Table~\ref{tab:main: Text-Image Retrieval}, we report image-text retrieval results on COCO~\cite{lin2014microsoft}, Flickr~\cite{young-etal-2014-image}, Urban1K~\cite{zhang2024long}, and the proposed RCLIP-Bench, covering standard retrieval, long-caption retrieval, and commonsense reasoning caption retrieval settings. ReasonCLIP and ReasonSigLIP both outperform their respective baselines in descriptive and reasoning retrieval, even including strong large-scale variants. Notably, despite being trained on a comparatively smaller reasoning corpus, our models surpass approaches that rely on substantially larger pretraining datasets or architectural modifications (e.g., MetaCLIP, Eva-CLIP-02, Long-CLIP). These results suggest that reasoning-oriented continual pretraining is effective for enhancing visually grounded representation quality.


\begin{table}[t]
\centering
\caption{\textbf{Commonsense reasoning results.} $^\dagger$Human evaluation is conducted on 1,000 samples. We report Stage 1 results for our models. Full results shown in Table~\ref{tab:app: commonsense_reasoning}. }
\label{tab:main: commonsense_reasoning}
\setlength{\tabcolsep}{2pt}
\resizebox{\linewidth}{!}{
\begin{tabular}{lcc|ccc|cccccc|cccccc|cccccc}
\toprule
\multirow{3}{*}{Model}  
& \multirow{3}{*}{Res.} 
& \multirow{3}{*}{Data}  

& \multicolumn{3}{c|}{WinoGAViL \cite{bitton2022winogavil}} 
& \multicolumn{18}{c}{RCLIP-Bench \textbf{(Ours)}} \\

&  &  
& \multicolumn{3}{c|}{Candidates} 
& \multicolumn{6}{c|}{V1: Visual Grounding} 
& \multicolumn{6}{c|}{V2: Evidence Awareness} 
& \multicolumn{6}{c}{V3: Visual Reasoning} \\

&  &  
& 5-6 & 10-12 & Avg. 
& Att. & Nou. & Rel. & Sen. & Sub. & Avg.
& S. & A. & C. & T. & P. & Avg.
& S. & A. & C. & T. & P. & Avg. \\
\bottomrule
\rowcolor{gray!15}
\multicolumn{24}{l}{\textit{Scale - ViT Base (86M)}} \\
OpenCLIP \cite{ilharco_gabriel_2021_5143773}  & 224{\scriptsize/32} & 0.4B
&53.2 & 42.6 & 50.6
&21.7 & 26.6 & \textbf{16.7} & \textbf{19.2} & 29.1 & 22.7
& \underline{17.6} & \underline{18.5} & 18.3 & 28.2 & \underline{20.8} & 20.7
& 18.5 & \underline{26.3} & 22.0 & 18.8 & \underline{23.7} & 21.8
\\
MetaCLIP \cite{xu2023demystifying}  &224{\scriptsize/32} & 0.4B 
&\textbf{58.3} & \textbf{51.0} & \textbf{56.5}
& 20.0 & 27.5 & 16.1 & 18.5 & 29.9 & 22.4
& 12.6 & 15.3 & 19.8 & 26.9 & 17.3 & 18.4
& 16.7 & 21.7 & \textbf{26.9} & \textbf{28.9} & 19.5 & \underline{22.7}
\\
DataComp \cite{datacomp} & 224{\scriptsize/32} & 1.4B
& 55.6 & 45.8 & 53.3
&21.2 & 28.3 & \underline{16.3} & \underline{18.9} & \underline{30.8} & \underline{23.1}
& 14.2 & 16.1 & \underline{24.3} & \textbf{34.9} & 16.9 & \underline{21.3}
& \underline{19.4} & 22.9 & \underline{22.3} & \underline{23.9} & 22.8 & 22.3
\\
CLIP \cite{radford2021learning}     &224{\scriptsize/32} & 0.4B
& 53.5 & 41.5 & 50.6 
& \underline{22.4} & \underline{28.9} & 16.2 & 17.0 & 27.2 & 22.3
& 12.9 & 15.4 & 11.3 & 26.5 & 15.2 & 16.2
& \textbf{19.6} & 23.7 & 21.6 & 22.5 & \textbf{25.8} & 22.6
\\
\rowcolor{gray!10}\textbf{ReasonCLIP} & 224{\scriptsize/32} & \textbf{+42M} 
& \underline{57.8} & \underline{49.6} & \underline{55.8}
& \textbf{22.5} & \textbf{31.4} & 14.2 & 18.5 & \textbf{31.4} & \textbf{23.6}
&  \textbf{18.2} & \textbf{19.3} & \textbf{25.0} & \underline{31.5} & \textbf{21.3} & \textbf{23.1}
& 18.5 & \textbf{31.6} & 21.5 & 23.5 & 23.6 & \textbf{23.8} 
 \\

\bottomrule
\rowcolor{gray!15}
\multicolumn{24}{l}{\textit{Scale - ViT Large (307M)}} \\
OpenCLIP \cite{ilharco_gabriel_2021_5143773}   & 224{\scriptsize/14} & 2.0B
& 53.8 & 44.0 & 51.5
&22.4 & 32.3 & \underline{15.8} & 19.3 & 34.4 & 24.8
& \underline{14.2} & 15.0 & \textbf{25.5} & 29.3 & 17.5 & \underline{20.3}
& 19.9 & 18.6 & 19.9 & 27.7 & 19.0 & 21.0
\\
 Eva-CLIP-02 \cite{eva02}   & 224{\scriptsize/14} & 2.0B 
& 56.5  & 50.5 &55.0
& 22.4 & 33.7 & 13.3 & 18.8 & \underline{35.4} & 24.7
& 11.3 & 16.9 & \underline{19.6} & 30.8 & 14.7 & 18.7
& 21.8 & 22.5 & 20.0 & 25.4 & 19.0 & 21.7
\\
MetaCLIP \cite{xu2023demystifying}   &  224{\scriptsize/14}  & 2.5B 
& 59.6 & 53.6 & 58.2 
& 23.5 & 33.3 & 15.6 & \underline{19.7} & 35.1 & 25.4
& 13.3 & 17.6 & 18.7 & 24.0 & 16.3 & 18.0
&19.3 & 22.5 & \textbf{25.5} & \underline{30.1} & 23.8 & 24.3
\\
Long-CLIP~\cite{zhang2024long}  & 224{\scriptsize/14} & 0.4B &
\textbf{63.7} & \underline{56.5} & \textbf{62.0}
& \underline{24.0} & \underline{34.3} & \textbf{16.4} & \textbf{20.4} & 34.2 & \underline{25.9}
& \textbf{21.0} & \textbf{28.0} & \underline{19.6} & \textbf{35.6} & \textbf{22.2} & \textbf{25.3}
& \textbf{25.4} & \textbf{36.5} & 18.9 & \textbf{31.8} & 25.8 & \textbf{27.7}
\\
CLIP \cite{radford2021learning}    &  224{\scriptsize/14}  & 0.4B 
&   53.3 & 41.1 &50.4
&20.4 & 31.1 & 13.9 & 17.5 & 29.8 & 22.5
& 8.2 & 13.0 & 12.5 & 32.1 & \underline{20.8} & 17.3
& \underline{23.4} & 25.8 & \underline{24.1} & 21.2 & \textbf{29.3} & \underline{24.8}
\\
\rowcolor{gray!10}\textbf{ReasonCLIP}  &  224{\scriptsize/14} & \textbf{+42M}
&  \underline{62.5} & \textbf{56.6} & \underline{61.1}
& \textbf{25.8} & \textbf{36.7} & 15.0 & 19.3 & \textbf{36.0} & \textbf{26.6}
& 11.7 & \underline{19.2} & 15.7 & \underline{34.3} & 15.2 & 19.2
& 19.6 & \underline{28.3} & 18.0 & 26.7 & \underline{28.8} & 24.3
\\

\midrule
ViTamin \cite{chen2024vitamin}  &  336{\scriptsize/16}  & 1.0B  
& 54.4&46.8 & 53.7
& 21.3 & 33.3 & \textbf{15.4} & \textbf{20.1}& 34.7& \underline{25.1}
& \underline{14.6} & 15.5 & \underline{23.1} & 31.5 & 16.2 & \underline{20.2}
&20.2 & 20.2 & 16.1 & 24.3 & 21.3 & 20.4
\\
Eva-CLIP-02 \cite{sun2023eva}   &  336{\scriptsize/14}  & 2.0B  
& 56.8& 50.5 & 55.3
& \underline{22.3} & \underline{34.1} & 13.5 & 18.8 & \underline{36.0} & 24.9
& 11.1 & 17.7 & 18.6 & 29.7 & 14.5 & 18.3
&\underline{20.6} & 22.4 & 20.8 & \underline{25.5} & 19.2 & 21.7 
 \\
SigLIP2\cite{tschannen2025siglip}    & 384{\scriptsize/16}  & 10.0B  
& \textbf{63.2}  & \textbf{58.4} & \textbf{62.1} 
& 16.3 & 22.6 & 14.2 & 12.6 & 20.2 & 17.2
& \textbf{15.9} & \underline{18.7} & \textbf{24.8} & \underline{33.1} & \underline{20.4} & \textbf{22.6}
&19.4 & 21.5 & \underline{21.2} & 21.9 & 16.9 & 20.2
\\
CLIP \cite{radford2021learning}   & 336{\scriptsize/14}  & 0.4B 
& 53.0 & 41.4 & 50.2
& 21.3 & 31.7 & 13.7 & 17.6 & 30.3 & 22.9
& 8.2 & 13.3 & 13.5 & 32.1 & \textbf{21.3} & 17.7
&\textbf{24.2} & \underline{26.0} & \textbf{25.4} & 21.2 & \underline{28.3} & \underline{25.0} 
\\
\rowcolor{gray!10}\textbf{ReasonCLIP}  & 336{\scriptsize/14} & \textbf{+42M} 
&   \underline{62.3} & \underline{54.6} & \underline{60.2}
& \textbf{26.4} & \textbf{37.5} & \textbf{15.4} & \underline{19.7} & \textbf{36.8} & \textbf{27.2}
& 11.7 & \textbf{19.8} & 15.6 & \textbf{34.3} & 14.9 & 19.3
& 20.3 & \textbf{29.7} & 19.5 & \textbf{28.8} & \textbf{28.6} & \textbf{25.4}
\\

\bottomrule
\rowcolor{gray!15}
\multicolumn{24}{l}{\textit{Scale - ViT So400m (428M)}} \\
SigLIP \cite{zhai2023sigmoid}  & 384{\scriptsize/14}  & 10.0B 
& \underline{63.0} & \underline{57.7} & \underline{61.7} 
& \underline{24.0} & \underline{33.3} & 14.3 & \underline{21.2} & \underline{29.3} & \underline{24.4}
& 13.0 & 17.3 & 28.0 & \underline{30.3} & 22.2 & 22.2
&20.6 & \underline{25.6} & 17.7 & \textbf{30.5} & 14.5 & 21.8
\\
\rowcolor{gray!10}\textbf{ReasonSigLIP}& 384{\scriptsize/14} & \textbf{+42M} 
&58.7 & 50.9 & 56.8
& \textbf{32.1} & \textbf{38.5} & \textbf{17.2} & \textbf{25.1} & \textbf{41.1} & \textbf{30.8}
& 16.1 & \underline{29.1} & 28.8 & 24.9 & \textbf{30.1} & \underline{25.8}
&\textbf{26.2} & \textbf{34.3} & \underline{21.0} & 23.9 & \underline{24.5} & \textbf{26.0}
\\

SigLIP2 \cite{tschannen2025siglip} & 384{\scriptsize/14}  & 10.0B 
&62.1 & 56.8 & 60.8
& 15.7 & 19.9 & 14.3 & 12.1 & 17.6 & 15.9
& \underline{17.6} & 19.1 & \textbf{30.7} & \textbf{35.7} & 18.3 & 24.3
& 20.5 & 23.6 & \textbf{23.1} & \underline{25.0} & 16.2 & 21.7
\\

\rowcolor{gray!10}\textbf{ReasonSigLIP2}& 384{\scriptsize/14} & \textbf{+42M} 
&\textbf{65.2} & \textbf{60.1} & \textbf{64.0}
& 18.1 & 20.9 & \underline{15.0} & 12.5 & 17.8 & 16.9
& \textbf{17.2} & \textbf{27.5} & \underline{30.5} & 27.9 & \underline{28.0} & \textbf{26.2}
& 20.4 & \textbf{34.3} & 18.3 & 23.1 & \textbf{26.1} & \underline{24.4}
\\

\bottomrule
\rowcolor{gray!15}
\multicolumn{24}{l}{\textit{Scale - ViT Giant Opt (1B)}} \\
SigLIP2 \cite{tschannen2025siglip}   &384{\scriptsize/16}  & 10.0B  
&\textbf{64.4} & \textbf{61.5} & \textbf{63.7}
&15.0 & 20.4 & \textbf{15.2} & 11.5 & 15.6 & 15.6
& \textbf{16.1} & 19.3 & \textbf{32.1} & \textbf{28.7} & 21.2 & 23.5
& \textbf{20.2} & 23.5 & \textbf{20.5} & \textbf{29.8} & 16.9 & 22.2
\\
\rowcolor{gray!10}\textbf{ReasonSigLIP2}& 384{\scriptsize/16} & \textbf{+42M} 
&62.1 & 60.1 & 61.6
& \textbf{18.1} & \textbf{20.9} & 15.0 & \textbf{12.5} & \textbf{17.8} & \textbf{16.9}
& \textbf{16.1} & \textbf{28.4} & 27.8 & 24.8 & \textbf{30.0} & \textbf{25.4}
&20.1 & \textbf{30.4} & 15.9 & 25.6 & \textbf{25.8} & \textbf{23.6} 
 \\
\bottomrule

\rowcolor{gray!15}
\multicolumn{24}{l}{\textit{Unbounded Scale}} \\
GPT-5.2 & - & - & - &- & - & 92.7 & 83.6 & 78.2 & 85.5 & 94.5 & 86.5 & 92.7 & 83.6 & 89.1 & 87.3 & 80.0 & 86.5 & 89.1 & 80.0 & 90.9 & 87.3 & 94.5 & 88.4 \\
Human$^\dagger$ & - & - & - &- & - & \textbf{96.5} & \textbf{95.1} & \textbf{95.3} & \textbf{97.1} & \textbf{94.7} & \textbf{95.7} & \textbf{93.0} & \textbf{89.7} & \textbf{93.9} & \textbf{88.7} & \textbf{90.0} & \textbf{91.1} & \textbf{92.2} & \textbf{85.8} & \textbf{94.0} & \textbf{90.1} & \textbf{96.2}& \textbf{91.7}\\
\bottomrule
\end{tabular}
}
\end{table}

\noindent \textbf{Observation $\textbf{2.}$}  \quad \textit{RCLIP-Bench provides a hierarchical evaluation of visually grounded reasoning.} As shown in Table~\ref{tab:main: commonsense_reasoning}, while WinoGAViL is informative for vision-language associations, high scores on this benchmark do not reliably reflect truly visual reasoning ability. RCLIP-Bench addresses this by decomposing evaluation into three hierarchical levels, revealing that no single model consistently dominates across all levels, and that strong retrieval performance does not guarantee strong reasoning. Notably, our ReasonCLIP consistently improves over its base CLIP backbone across all three levels, demonstrating that reasoning-oriented continual pretraining effectively enhances visually grounded reasoning ability beyond what WinoGAViL alone can capture.

\begin{table}[t]
\centering
\caption{\textbf{Compositional reasoning results.}  We report Stage 2 results for our models. Full results shown in Table~\ref{tab:app: compositional_reasoning}. }
\label{tab:main: compositional_reasoning}
\setlength{\tabcolsep}{2.5pt}
\resizebox{\linewidth}{!}{
\begin{tabular}{p{2.5cm}lccccccccc}
\toprule
\multirow{2}{*}{Model} & \multirow{2}{*}{ViT}  & \multirow{2}{*}{Res.} &    \multirow{2}{*}{Data} 
& \multirow{2}{*}{WhatsUp \cite{kamath2023whatsup}}
& \multirow{2}{*}{VALSE \cite{parcalabescu2022valse}}
& \multirow{2}{*}{CREPE \cite{ma2023crepe}}
& \multirow{2}{*}{SugarCREPE \cite{hsieh2023sugarcrepe}}
& \multicolumn{2}{c}{SugarCrepe++ \cite{dumpala2024sugarcrepe++}}
& \multirow{2}{*}{Avg.}\\
\cmidrule(lr){9-10}
 & & & & & & & & ITT & TOT \\
\midrule
MetaCLIP  \cite{xu2023demystifying} & B{\scriptsize/32} & 224 & 0.4B & 46.5 & 69.1 & 18.2 & 76.9 & 58.1 & 50.6 & 53.2\\
CLIP \cite{radford2021learning} &B{\scriptsize/32} &  224 & 0.4B & 41.2 & 67.5 & 23.9 & 73.1 & 60.0 & 46.7 & 52.1 \\
\rowcolor{gray!10}\textbf{ReasonCLIP} & B{\scriptsize/32} & 224 & \textbf{+58M}
& 51.3 {\scriptsize \textcolor{red}{\textbf{$\uparrow$10.1}}}
& 72.5 {\scriptsize \textcolor{red}{\textbf{$\uparrow$5.0}}}
& 24.0 {\scriptsize \textcolor{red}{\textbf{$\uparrow$0.1}}}
& 75.6 {\scriptsize \textcolor{red}{\textbf{$\uparrow$2.5}}}
& 61.5 {\scriptsize \textcolor{red}{\textbf{$\uparrow$1.5}}}
& 61.5 {\scriptsize \textcolor{red}{\textbf{$\uparrow$14.8}}}
& 57.7 {\scriptsize \textcolor{red}{\textbf{$\uparrow$5.6}}}\\
\midrule
\multicolumn{10}{l}{-- \textit{Compositional Reasoning Methods: Fine-tuned on MS-COCO, 100K Samples}} \\
Triplet-CLIP \cite{patel2024tripletclip} &B{\scriptsize/32} &  224 & -
& 41.6 & 64.2 & 15.0 & 82.7 & 61.7 & 57.4 & 53.8 \\
GNM-CLIP \cite{sahin2024enhancing} & B{\scriptsize/32} & 224 & -
& 41.6 & 70.7 & 17.4 & 77.9 & 60.2 & 60.0 & 54.6 \\
CE-CLIP \cite{zhang2024contrasting} & B{\scriptsize/32} & 224 & -
& 40.7 & 76.0 & 34.8 & 86.0 & 55.7 & 57.0 & 58.4 \\
NegCLIP \cite{yuksekgonul2023when} & B{\scriptsize/32} & 224 & -
& 42.4 & 73.7 & 30.5 & 83.6 & 65.0 & 62.5 & 59.6 \\
READ-CLIP \cite{kwon2025enhancing} & B{\scriptsize/32} & 224 & -
& 43.9 & 76.2 & 41.5 & 87.0 & 69.8 & 66.2 & 64.1 \\

\rowcolor{gray!10}\makecell[l]{\textbf{ReasonCLIP}\\+ READ\cite{kwon2025enhancing}} &B{\scriptsize/32} & 224 & -
& 45.2 {\scriptsize \textcolor{red}{\textbf{$\uparrow$1.3}}}
& 79.2 {\scriptsize \textcolor{red}{\textbf{$\uparrow$3.0}}}
& 44.9 {\scriptsize \textcolor{red}{\textbf{$\uparrow$3.4}}}
& 88.5 {\scriptsize \textcolor{red}{\textbf{$\uparrow$1.5}}}
& 69.2 {\scriptsize \textcolor{blue}{\textbf{$\downarrow$0.6}}}
& 67.0 {\scriptsize \textcolor{red}{\textbf{$\uparrow$0.8}}}
& 65.7 {\scriptsize \textcolor{red}{\textbf{$\uparrow$1.6}}}\\

\midrule

Eva-CLIP-02 \cite{eva02} & L{\scriptsize/14} & 224 & 2.0B & 46.3 & 73.5 & 21.4 & 81.6 & 63.3 & 52.7 & 56.5 \\
MetaCLIP \cite{xu2023demystifying} & L{\scriptsize/14} & 224 & 2.5B & 48.6 & 71.2 & 19.7 & 80.9 & 64.4 & 54.0 & 56.5 \\
Long-CLIP \cite{zhang2024long} & L{\scriptsize/14} & 224 & 0.4B & 45.6 & 76.1 & 35.1 & 82.9 & 59.8 & 48.4 & 58.0 \\
CLIP \cite{radford2021learning} &L{\scriptsize/14} & 224 & 0.4B& 40.7 & 68.8 & 20.5 & 73.4 & 60.6 & 44.3 & 51.4  \\
\rowcolor{gray!10}\textbf{ReasonCLIP} & L{\scriptsize/14} & 224 & \textbf{+58M}
& 46.0 {\scriptsize \textcolor{red}{\textbf{$\uparrow$5.3}}}
& 75.9 {\scriptsize \textcolor{red}{\textbf{$\uparrow$7.1}}}
& 24.8 {\scriptsize \textcolor{red}{\textbf{$\uparrow$4.3}}}
& 78.2 {\scriptsize \textcolor{red}{\textbf{$\uparrow$4.8}}}
& 63.1 {\scriptsize \textcolor{red}{\textbf{$\uparrow$2.5}}}
& 59.5 {\scriptsize \textcolor{red}{\textbf{$\uparrow$15.2}}}
& 57.9 {\scriptsize \textcolor{red}{\textbf{$\uparrow$6.5}}}\\

SigLIP2 \cite{tschannen2025siglip} & L{\scriptsize/16} & 384 & 10.0B & 45.5 & 70.6 & 15.9 & 81.3 & 65.7 & 51.7 & 55.1\\
CLIP  \cite{radford2021learning}& L{\scriptsize/14} & 336 & 0.4B & 41.8 & 68.8 & 20.9 & 74.8 & 60.6 & 44.1 & 51.8  \\
\rowcolor{gray!10} \textbf{ReasonCLIP} & L{\scriptsize/14} & 336 & \textbf{+58M} 
& 46.3 {\scriptsize \textcolor{red}{\textbf{$\uparrow$5.5}}}
& 76.3 {\scriptsize \textcolor{red}{\textbf{$\uparrow$7.5}}}
& 24.8 {\scriptsize \textcolor{red}{\textbf{$\uparrow$3.9}}}
& 78.1 {\scriptsize \textcolor{red}{\textbf{$\uparrow$3.3}}}
& 63.8 {\scriptsize \textcolor{red}{\textbf{$\uparrow$3.2}}}
& 59.4 {\scriptsize \textcolor{red}{\textbf{$\uparrow$15.3}}} 
& 58.1 {\scriptsize \textcolor{red}{\textbf{$\uparrow$7.7}}}\\

\midrule
SigLIP \cite{zhai2023sigmoid} & So{\scriptsize/14} & 384 & 10.0B& 47.6 & 72.0 & 18.0 & 83.0 & 67.9 & 51.2 & 56.6\\
\rowcolor{gray!10}\textbf{ReasonSigLIP} & So{\scriptsize/14} &384 & \textbf{+58M}
& 49.7 {\scriptsize \textcolor{red}{\textbf{$\uparrow$2.1}}}
& 76.3 {\scriptsize \textcolor{red}{\textbf{$\uparrow$4.3}}}
& 20.1 {\scriptsize \textcolor{red}{\textbf{$\uparrow$2.1}}}
& 84.1 {\scriptsize \textcolor{red}{\textbf{$\uparrow$1.1}}}
& 73.1 {\scriptsize \textcolor{red}{\textbf{$\uparrow$5.2}}}
& 72.9 {\scriptsize \textcolor{red}{\textbf{$\uparrow$21.7}}} 
& 62.7 {\scriptsize \textcolor{red}{\textbf{$\uparrow$6.1}}}\\

\midrule
SigLIP2 \cite{tschannen2025siglip}  & G{\scriptsize/16} &384 & 10.0B & 46.9 & 71.4 & 16.5 & 83.0 & 67.4 & 48.8 & 55.7 \\
\rowcolor{gray!10}\textbf{ReasonSigLIP2} & G{\scriptsize/16} &384 & \textbf{+58M}
& 45.3 {\scriptsize \textcolor{blue}{\textbf{$\downarrow$1.6}}}
& 77.9 {\scriptsize \textcolor{red}{\textbf{$\uparrow$6.5}}}
& 20.5 {\scriptsize \textcolor{red}{\textbf{$\uparrow$4.0}}}
& 76.8 {\scriptsize \textcolor{blue}{\textbf{$\downarrow$6.2}}}
& 67.0 {\scriptsize \textcolor{blue}{\textbf{$\downarrow$0.4}}}
& 66.9 {\scriptsize \textcolor{red}{\textbf{$\uparrow$18.1}}}
& 60.2 {\scriptsize \textcolor{red}{\textbf{$\uparrow$4.5}}}\\
\bottomrule
\end{tabular}
}
\end{table}

\noindent \textbf{Observation $\textbf{3.}$}  \quad \textit{ReasonCLIP improves compositional reasoning across backbones without task-specific fine-tuning.} Table~\ref{tab:main: compositional_reasoning} compares compositional reasoning results across multiple benchmarks. Across different backbone architectures and model scales, ReasonCLIP and ReasonSigLIP consistently outperform their base models on nearly all benchmarks, with average gains of approximately +5 to +7 points, demonstrating that large-scale reasoning supervision stably enhances structured compositional reasoning. Compared to data-centric approaches that rely on larger or higher-quality pretraining corpora, ReasonCLIP still achieves superior performance under the same backbone. Moreover, ReasonCLIP achieves competitive performance against specialized compositional methods without task-specific fine-tuning. Furthermore, applying READ~\cite{kwon2025enhancing} on top of ReasonCLIP-B/32 yields additional gains, validating its effectiveness as a stronger backbone for specialized compositional methods.

\begin{table}[t]
\vspace{-1pt}
\centering
\setlength{\tabcolsep}{2pt}
\caption{MLLM results on LLaVA-NeXT.}
\resizebox{\linewidth}{!}{
\begin{tabular}{lcccccccccccc}
\toprule
Model 
& {\scriptsize AI2D \cite{kembhavi2016diagram}} 
& {\scriptsize ChartQA \cite{masry2022chartqa}}  
& {\scriptsize SciQA \cite{saikh2022scienceqa}}   
& {\scriptsize RealWordQA \cite{xai2024grok15v}}  
& {\scriptsize VisualLogic \cite{xu2025visulogic}}  
& {\scriptsize OKVQA \cite{okvqa}}  
&  
& {\scriptsize GQA \cite{hudson2019gqa}} 
& {\scriptsize MME \cite{fu2023mme}}  
& {\scriptsize MMStar \cite{chen2024we}} 
&  {\scriptsize MMVP\cite{tong2024eyes}} \\
\midrule
LLaVA-NeXT~\cite{liu2024llavanext} & 58.1 & 26.8 & \textbf{73.1}& 49.4& 25.7  & 34.0 & &  52.8 & 1296.1 & 39.7 & 58.7   \\
\textbf{w/ReasonCLIP} & \textbf{58.6} & \textbf{28.0} & 73.0 &\textbf{49.5}& 25.7 &\textbf{35.8} &  & \textbf{53.4}  & \textbf{1301.8} & \textbf{39.9} & \textbf{62.3}  \\
\bottomrule
\end{tabular}
}

\label{tab:mllm_results}
\end{table}

\noindent \textbf{Observation $\textbf{4.}$}  \quad \textit{ReasonCLIP as a visual tower unlocks stronger MLLM reasoning ability.} We replace CLIP with ReasonCLIP in LLaVA-NeXT~\cite{liu2024llavanext} under an identical training setup and evaluate the model on commonsense-oriented MLLM benchmarks. Tab.~\ref{tab:mllm_results} shows that, across multiple visual reasoning benchmarks, replacing CLIP with ReasonCLIP leads to consistent improvements, particularly on OKVQA, VisualLogic, GQA, and MMStar. This indicates that, within the same multimodal framework, introducing visual representations with stronger reasoning awareness can bring reproducible performance gains for downstream multimodal models. Meanwhile, we also conduct additional evaluations on several non-reasoning benchmarks (e.g., AI2D and SciQA), and the results demonstrate that the model's performance on these non-reasoning tasks remains stable.


\subsection{Ablation Study}

\begin{figure*}[!t]
\centering
\resizebox{0.95\linewidth}{!}{
\begin{minipage}[t]{0.58\linewidth}
    \captionsetup{type=figure}
    \vspace{0.8em}
    \begin{subfigure}{0.49\linewidth}
        \includegraphics[width=\linewidth]{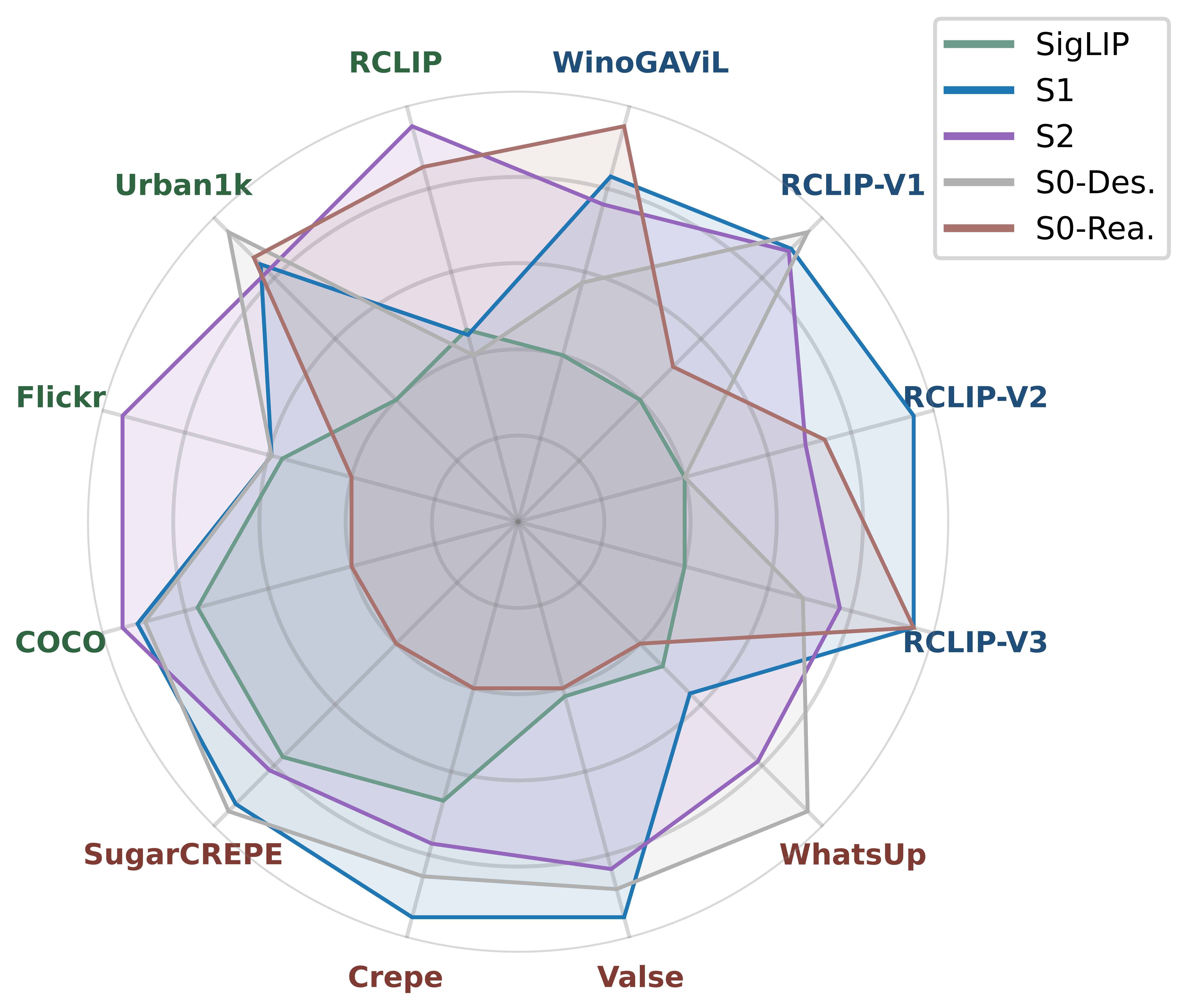}
    \end{subfigure}
    \hfill
    \begin{subfigure}{0.49\linewidth}
        \includegraphics[width=\linewidth]{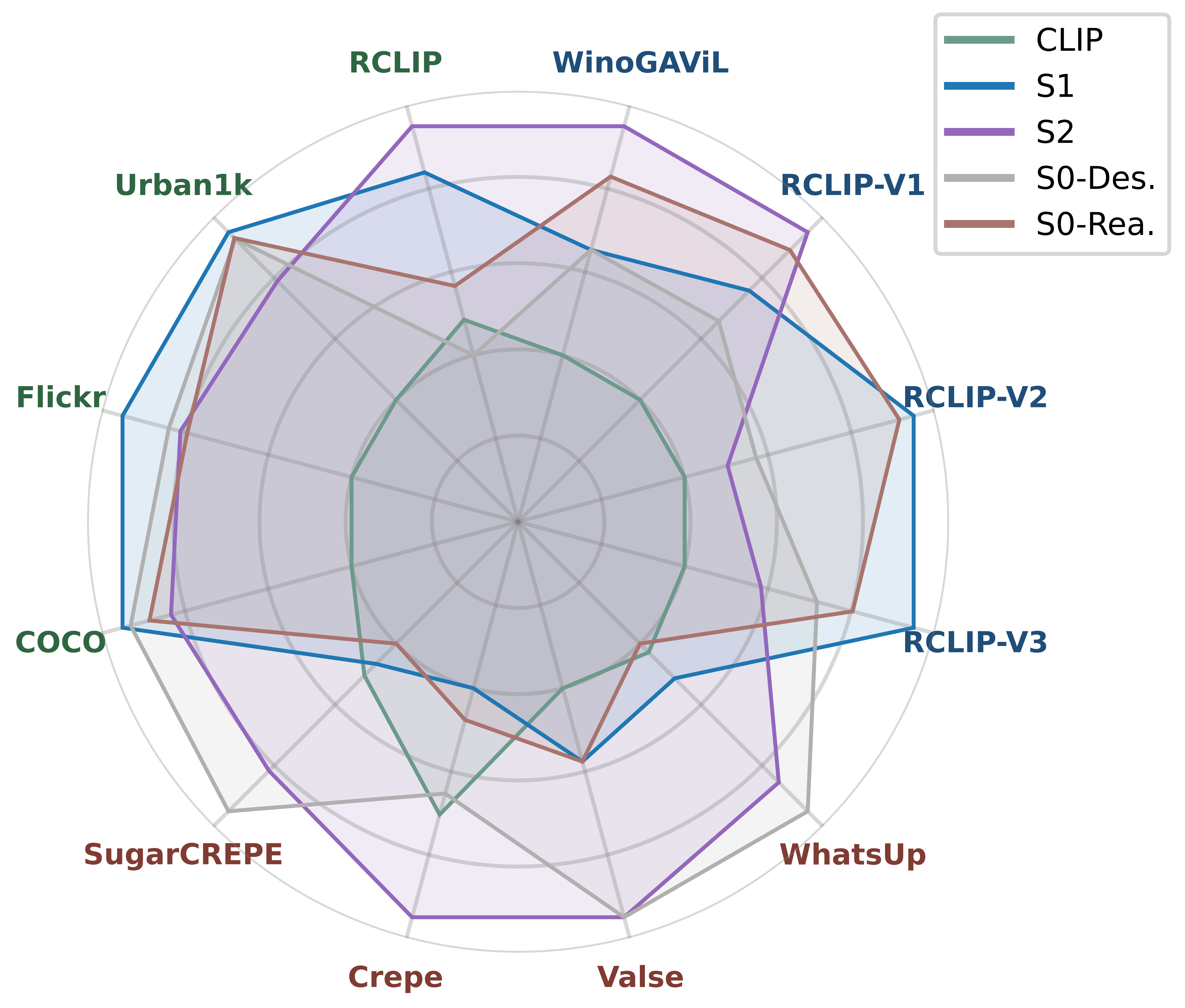}
    \end{subfigure}%
    \vspace{-5pt}
    \caption{Comparison of different training strategies.}
    \label{fig:radar}
\end{minipage}
\hfill
\begin{minipage}[t]{0.33\linewidth}
    \captionsetup{type=table}
    \caption{\textbf{Ablation study.}}
    \label{tab:ablation_study}
    \small
    \setlength{\tabcolsep}{3pt}
    \scalebox{0.65}{
        \begin{tabular}{lccccc}
            \toprule
            \multirow{2}{*}{Method}
                & \multicolumn{2}{c}{Retrieval}
                & \multicolumn{3}{c}{RCLIP-Bench} \\
            \cmidrule(lr){2-3} \cmidrule(lr){4-6}
                & COCO & Flickr & V1 & V2 & V3 \\
            \midrule
            \rowcolor{gray!7}
            \textbf{Stage 1} & 55.6 & \textbf{50.5} & \textbf{26.6} & \textbf{19.2} & \textbf{24.3} \\
            Des.-heavy       & \textbf{55.8} & 48.6 & 26.1 & 18.4 & 24.1 \\
            Rea.-heavy       & 52.5 & 46.0 & 23.5 & 15.3 & 19.0 \\
            Exp-Sched        & 53.2 & 47.7 & 25.8 & 17.6 & 23.3 \\
            \midrule
            \rowcolor{gray!7}
            \textbf{Stage 2} & 53.2 & \textbf{47.3} & \textbf{28.6} & \textbf{18.3} & \textbf{23.1} \\
            w/o cls          & \textbf{53.3} & 47.3 & \textbf{28.6} & 18.0 & 22.6 \\
            w/ L2            & 53.1 & 47.1 & 28.4 & 17.3 & 23.0 \\
            Single-Label     & 52.8 & 46.5 & 28.1 & 17.5 & 22.0 \\
            \bottomrule
        \end{tabular}%
    }
\end{minipage}%
}

\end{figure*}

\noindent\textbf{Training Strategy.} \quad
In Fig.~\ref{fig:radar}, we present radar plots for ReasonCLIP-L14-336 and ReasonSigLIP-So/14-384 under different training strategies, comparing their performance across three evaluation dimensions. 

For ReasonSigLIP, we observe that Stage 2 (S2) achieves the most balanced overall performance, particularly demonstrating stable advantages in retrieval tasks. However, on certain compositional benchmarks and RCLIP-Bench, Stage 1 (S1) performs better. This suggests that in scenarios requiring visually grounded reasoning, the progressive reasoning-aware alignment in S1 is more effective at preserving the synergy between visual semantic structure and reasoning signals, whereas explicit reasoning supervision in S2 may partially disrupt this balance.

For ReasonCLIP, the improvements from S1 are more pronounced, whereas those from S2 are relatively limited. This suggests that for CLIP backbones with limited capacity, reasoning enhancement must rely on strong visual grounding; excessive explicit supervision (e.g., S2) may weaken alignment. It also highlights the importance of large-scale pretraining and sufficient model capacity for effective reasoning improvement.

\noindent\textbf{Training Configurations.} \quad Table~\ref{tab:ablation_study} presents the ablation results under different training configurations, Des.-heavy / Rea.-heavy denote different branch weightings in Stage 1; Exp-Sched uses exponential scheduling; w/o cls removes category supervision; w/ L2 adds regularization; Single-Label replaces multi-label classification. The results show that our setting achieves the best overall performance, validating the effectiveness of our selected hyperparameters.

\section{Limitation and Outlook}

\noindent\textit{Limitations.}\quad
Due to computational constraints, our 58M-training is smaller than industrial CLIP pretraining. Our contribution is to validate a scalable reasoning-oriented paradigm rather than compete with ultra-large-scale settings. While evaluations are extensive, full coverage of CLIP applications is beyond scope; we release our models for broader community assessment.

\noindent\textit{Outlook.}\quad
In this paper, we revisit CLIP-style encoders and investigate whether visually grounded reasoning can be incorporated through pretraining alone. Across architectures and scales, the proposed ReasonCLIP-58M framework enhances visually grounded reasoning and integrates seamlessly into MLLMs. These results demonstrate that structured reasoning supervision can systematically extend the representational capacity of CLIP-style encoders, providing a practical pathway toward reasoning-aware vision-language foundation models.

\section*{Acknowledgments and Disclosure of Funding}
This research is funded by Khalifa University of Science and Technology through the Faculty Start-Ups under Project ID: KU-INT-FSU-2005-8474000775.
%
%
\bibliographystyle{splncs04}
\bibliography{ref}

\newpage

\appendix
\section{Dataset Cards}
\label{appendix: Dataset Cards}
\subsection{Reasoning Scope}
\label{appendix: Reasoning Level}
\begin{figure}[htbp]
    \vspace{-10pt}
    \centering
    \includegraphics[width=\linewidth]{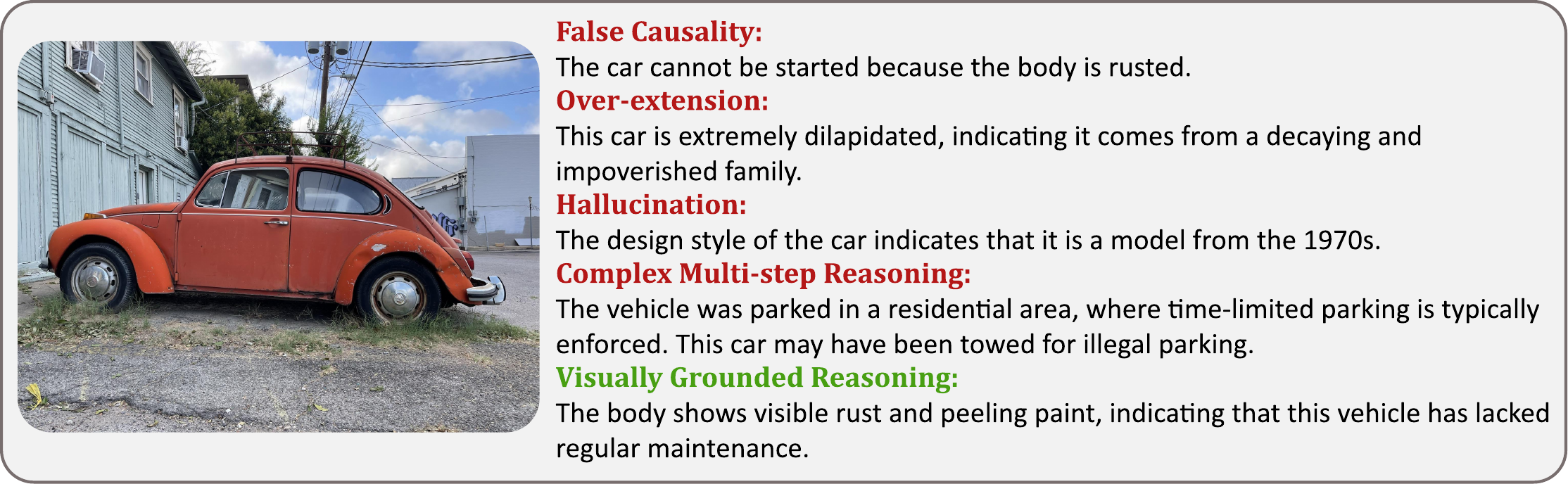}    
    \caption{
Representative examples for constructing excluded reasoning cases and visually grounded reasoning in ReasonCLIP.
} 
    \label{fig:app_exclude}
    \vspace{-10pt}
\end{figure}

As shown in Fig. \ref{fig:app_exclude}, to avoid substantially increasing the overall reasoning difficulty and to ensure data quality, we use prompts to constrain the reasoning level of the ReasonCLIP dataset to a limited range, while explicitly excluding samples that exhibit false causality, over-extension, hallucination, or complex multi-step reasoning beyond what is visually supported. 

\subsection{CC12M-Enhanced Dataset}
\label{appendix: CC12M-Enhanced}
\noindent\textbf{Data Statistic.}\quad We collect 10,388,539 images from CC12M Dataset, after annotation, each image is paired with 3 $T_{b}$ annotations, resulting in a total of \textbf{31,165,584} $I$-$T_{b}$ pairs. As shown in Fig. \ref{fig:app_stat} (a), The generated captions have an average length of 21.8 words with a variance of 15.8. In comparison, the raw captions have an average length of 17.3 words and a substantially higher variance of 163.3, indicating that our generated captions are more informative while maintaining significantly greater length consistency.

\noindent\textbf{Cost.}\quad The $T_{b}$ annotation, completed by Qwen2.5-VL-72B with Activation-aware Weight Quantization (AWQ), required approximately 1440 A100-64GB GPU hours.

\begin{figure}[htbp]
    \centering
    \includegraphics[width=\linewidth]{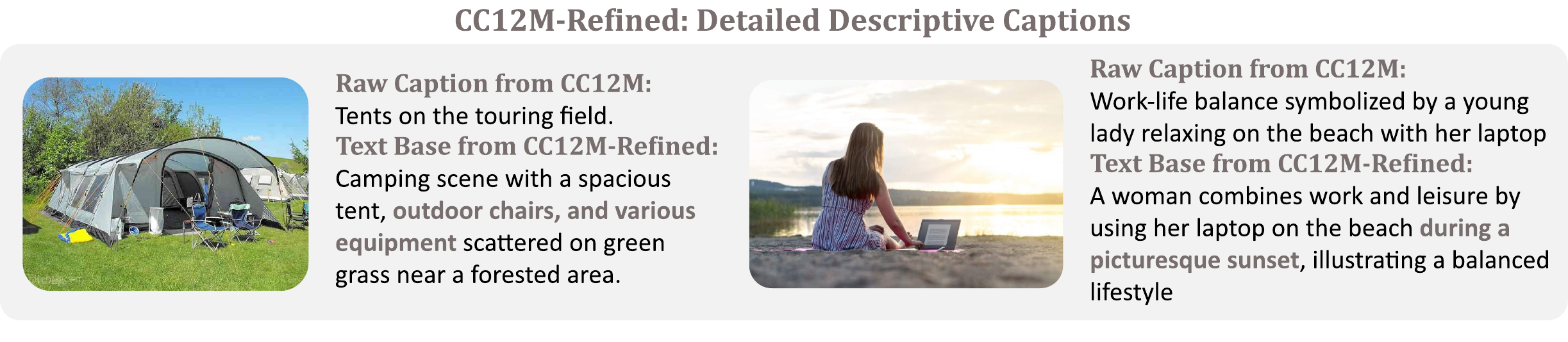}    
    \caption{
        Representative examples from CC12M-Refined Dataset.
    } 
    \label{fig:app_cc12m_refined}
\end{figure}

\noindent\textbf{Data Samples.}\quad Fig. \ref{fig:app_cc12m_refined} shows two data examples. 

\begin{figure}[htbp]
    \centering
    \includegraphics[width=\linewidth]{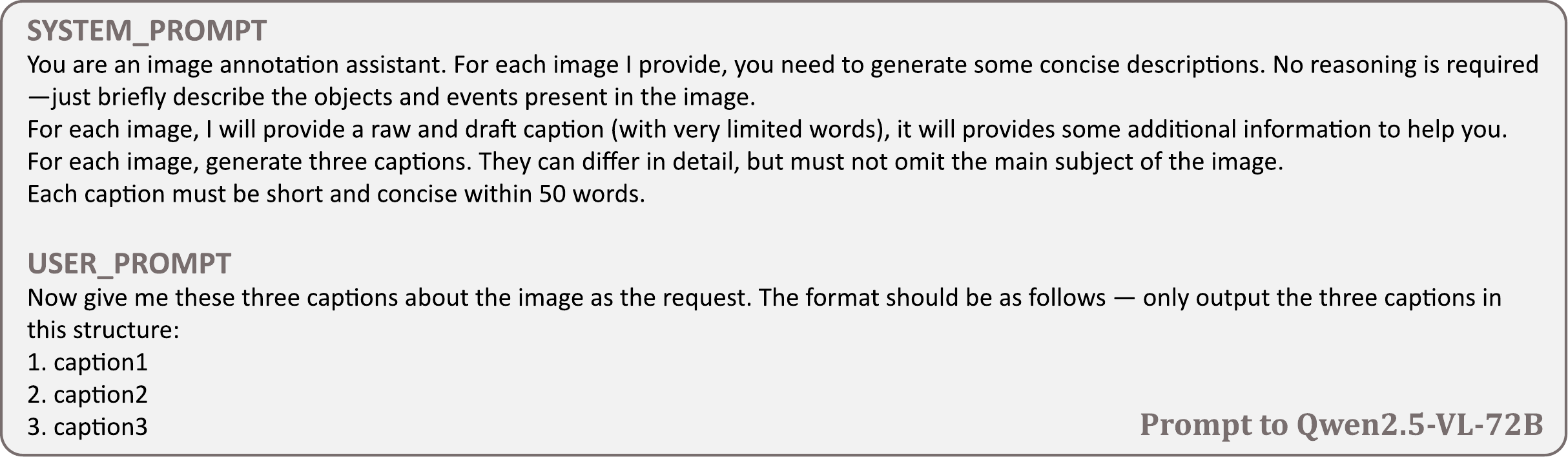}    
    \caption{
        Prompt for CC12M-Refined dataset generation.
    } 
    \label{fig:app_tb_prompt}
    \vspace{-10pt}
\end{figure}

\noindent\textbf{Prompt.}\quad Fig. \ref{fig:app_tb_prompt} shows the prompt for CC12M-Refined generation.

\subsection{ReasonLite-42M Dataset}
\label{appendix: ReasonLite-42M}
\begin{figure}[htbp]
    \centering
    \includegraphics[width=0.8\linewidth]{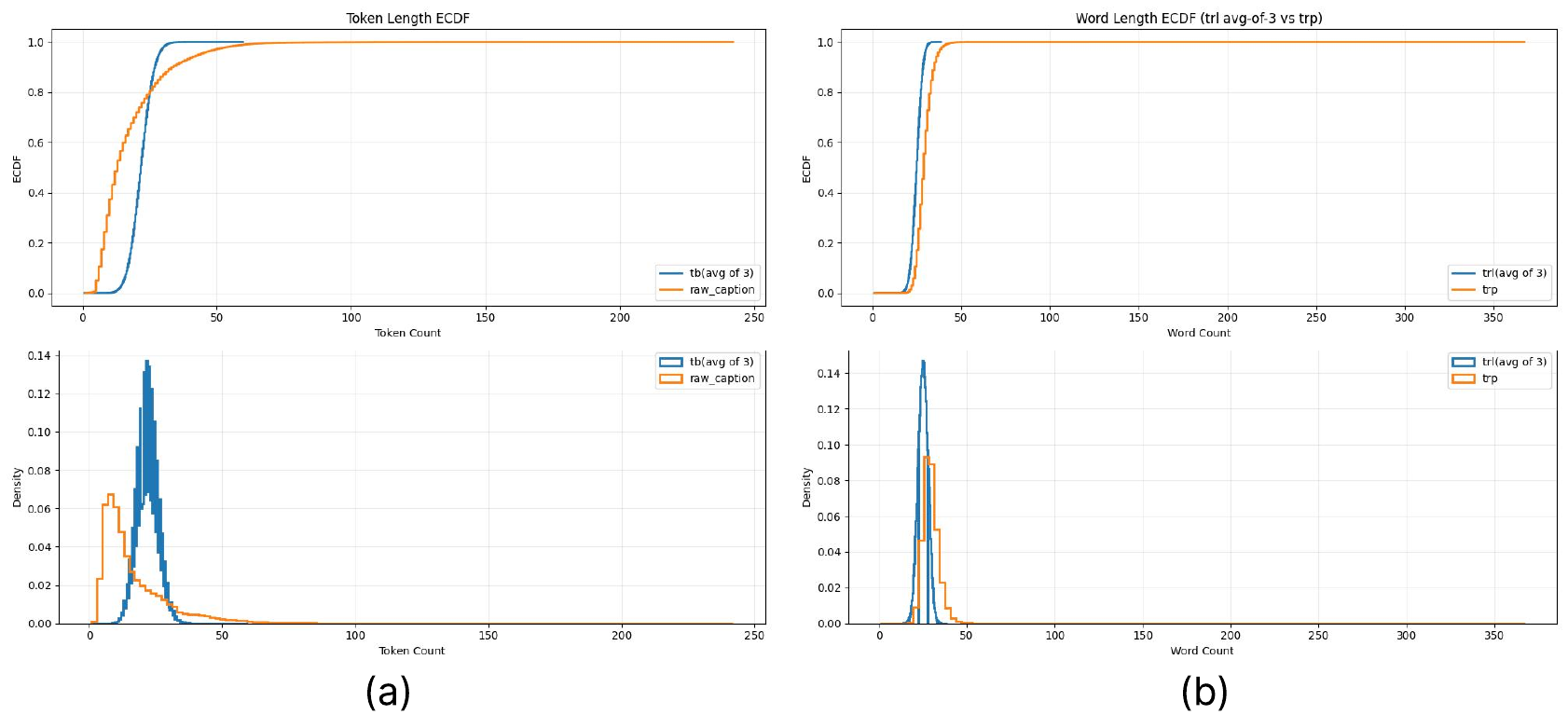}    
    \caption{
        (a) Statistics for raw captions and $T_{b}$, (b) Statistics for $T_{rl}$ and $T_{rp}$.
    } 
    \label{fig:app_stat}
    \vspace{-10pt}
\end{figure}

\noindent\textbf{Data Statistic.}\quad
We use 4,668,515 images from CC12M for ReasonLite Dataset, after annotation, each image is paired with 9 $T_{b}$-$T_{rl}$ annotations, resulting in a total of \textbf{42,016,635} $I$-$T_{b}$-$T_{rl}$ triplet pairs. As shown in Fig. \ref{fig:app_stat} (b), the generated captions have an average length of 25.0 words with a variance of 8.1.

\noindent\textbf{Cost.} The $T_{rl}$ annotation, completed by Qwen2.5-VL-72B, required approximately 1150 A100-64GB GPU hours.

\begin{figure}[htbp]
    \centering
    \includegraphics[width=\linewidth]{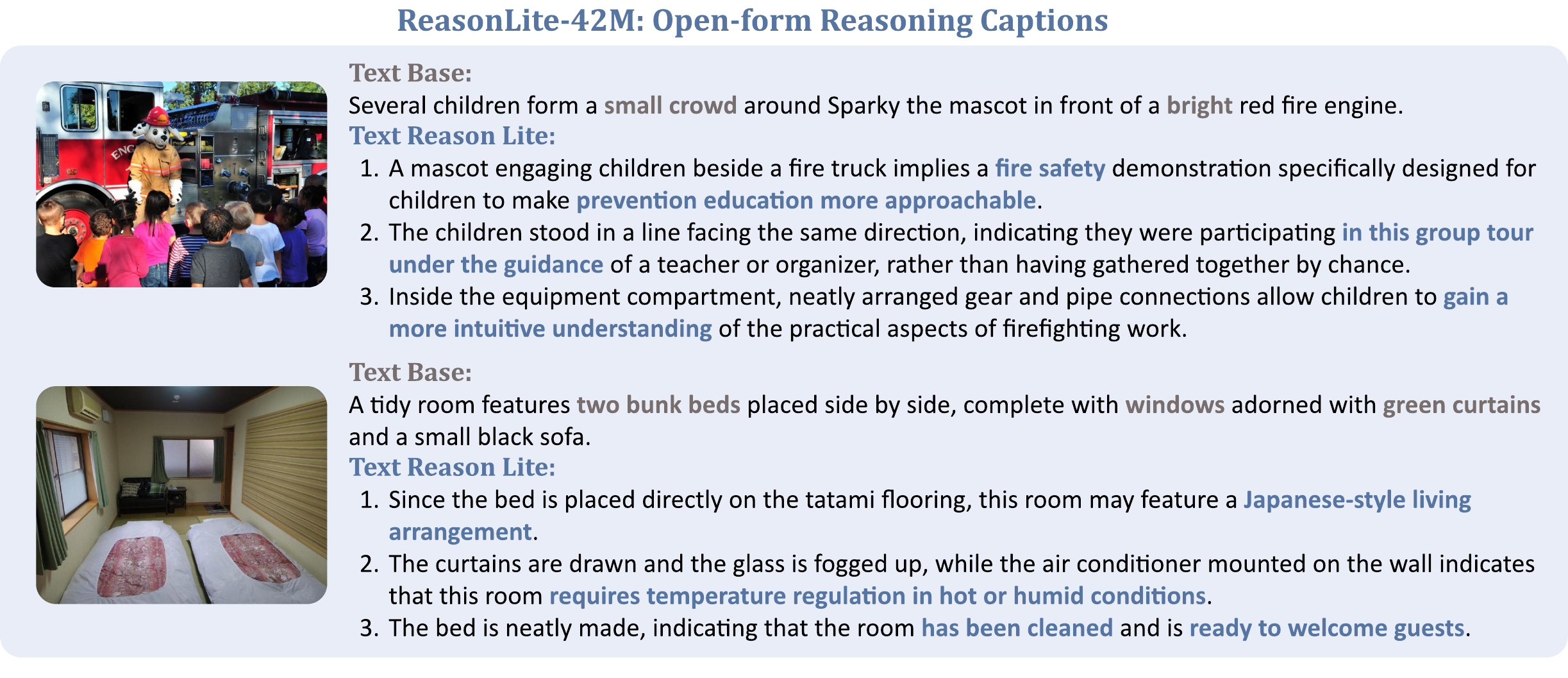}    
    \caption{
        Representative examples from ReasonLite-42M Dataset.
    } 
    \label{fig:app_reasonlite}
\end{figure}

\noindent\textbf{Data Samples.}\quad Fig. \ref{fig:app_reasonlite} shows two data examples from the ReasonLite Dataset.

\begin{figure}[htbp]
    \centering
    \includegraphics[width=\linewidth]{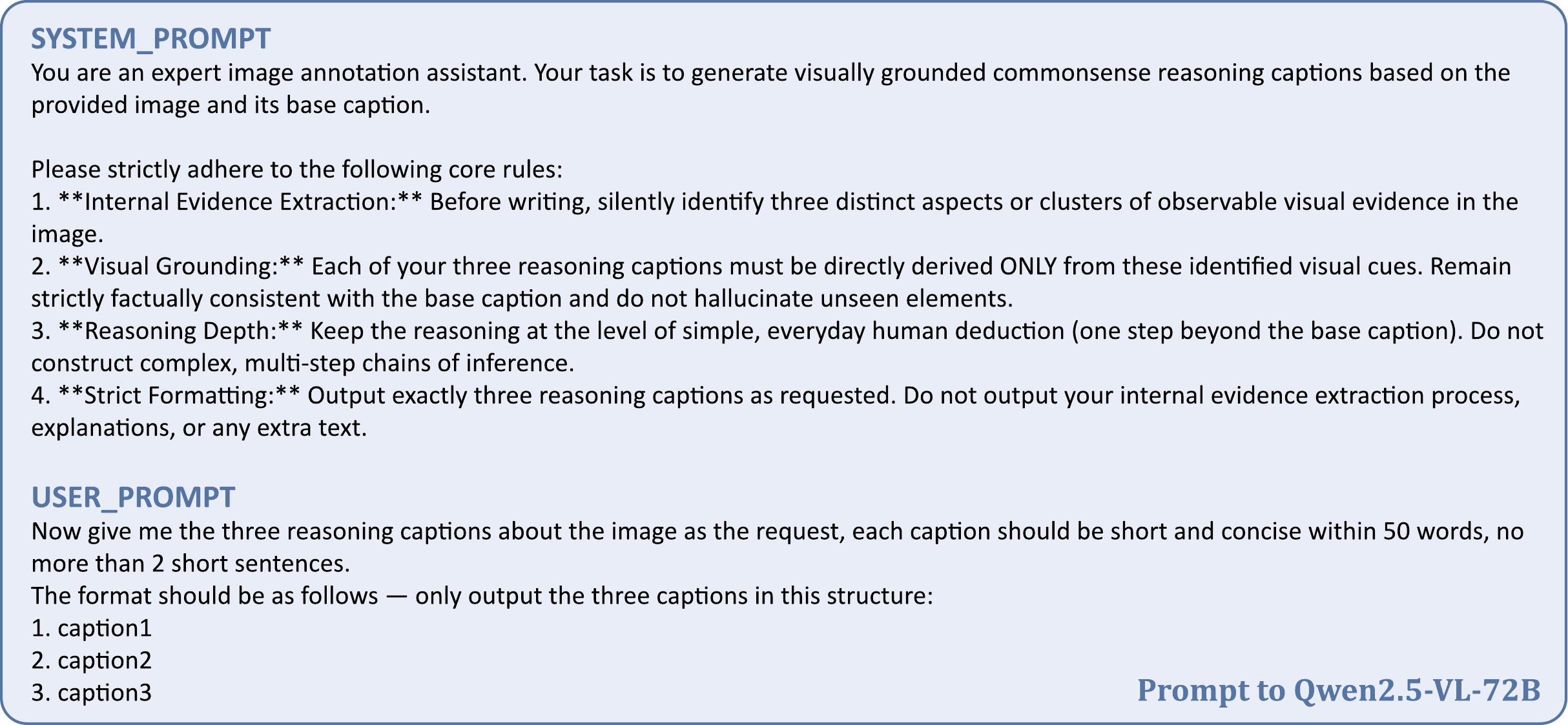}    
    \caption{
        Prompt for ReasonLite dataset generation.
    } 
    \label{fig:app_trl_prompt}
    \vspace{-10pt}
\end{figure}

\noindent\textbf{Prompt.}\quad
Fig. \ref{fig:app_trl_prompt} shows the prompt for ReasonLite Dataset generation.

\subsection{ReasonPro-16M Dataset}
\label{appendix: ReasonPro-16M}
\textbf{Data Statistic.}\quad We begin with 5,720,000 samples for classification. After the filtering stage, a total of 198,437 samples (3.47\%) are removed, leaving 5,521,563 valid samples. The removed data include cases that are unsuitable for classification (overly simple or overly complex content) as well as cases that do not satisfy the requirement of three reasoning categories. Then each image is paired with 3 $T_{rp}$ annotations, resulting in a total of \textbf{16,564,689} $\mathbf{img}$-$T_{rp}$ pairs. 

Among these examples, we obtain 4,766,178 samples of Spatial/Geometric Reasoning (S), 4,883,348 samples of Attribute/State Reasoning (A), 2,469,372 samples of Creature/Action Reasoning (C), 947,656 samples of Temporal/Phase Reasoning (T), and 3,498,135 samples of Physical Intuition Reasoning (P).

As shown in Fig. \ref{fig:app_stat} (b), the generated captions have an average length of 29.4 words with a variance of 18.2.

\noindent\textbf{Cost.}\quad The classification annotation costs approximately 280  GPU hours. The $T_{rp}$ annotation costs approximately 960 GPU hours. All annotations are completed by Qwen3-VL-32B.

\noindent\textbf{Reasoning Categories Definition.}\quad \label{Appendix:Reasoning Categories Definition} We provide specific definitions for the five reasoning categories in Table \ref{tab:appendix_reasoning_categories_detailed}. 
\begin{table*}[htbp]
\centering
\caption{Five categories of perception-grounded reasoning types in the proposed ReasonPro-16M Dataset.}
\setlength{\tabcolsep}{3pt}
\resizebox{\linewidth}{!}{
\begin{tabular}{lcl}
\toprule
\textbf{Category} & \textbf{ID} & \textbf{Definition} \\
\midrule
\makecell[l]{Spatial \\or\\ Geometric} & S &
\makecell[l]{
1.\ Understanding spatial relations between multiple entities, such as position,\\ direction, distance, occlusion, containment, or accessibility.\\
2.\ Focuses on how objects are arranged and interact in space.\\
3.\ Key idea: how placement or geometry affects visibility, reachability, or motion.
} \\
\midrule
\makecell[l]{Attribute \\or\\ State }& A &
\makecell[l]{
1.\ Understanding intrinsic properties or visible conditions of individual objects.\\
2.\ Includes appearance, surface texture, brightness, transparency, wetness,\\ deformation, openness, integrity, or on/off states.\\
3.\ Key idea: what physical or functional state each object is in and why, based on visible cues.
} \\
\midrule
\makecell[l]{Creature \\or\\ Action} & C &
\makecell[l]{
1.\ Understanding human or animal posture and actions to infer current or immediate behavior.\\
2.\ Involves body orientation, gesture, and interaction with objects or other agents.\\
3.\ Key idea: what the agent is doing or about to do.
} \\
\midrule
\makecell[l]{Temporal \\or\\ Phase} & T &
\makecell[l]{
1.\ Understanding the temporal stage of an event: just happened, ongoing, or about to happen.\\
2.\ Based on motion continuity, trajectory, or dynamic context.\\
3.\ Key idea: what phase or temporal transition the scene represents.
} \\
\midrule
\makecell[l]{Physical \\ Intuition} & P &
\makecell[l]{
1.\ Reasoning about relatively obvious physical relationships that can be inferred directly from visual cues.\\
2.\ Not relying on semantic common sense, but on visual physical intuition.\\
3.\ Key idea: intuitive judgments about stability, support, contact, forces, or other basic physical interactions.
} \\
\bottomrule
\end{tabular}
}
\vspace{-10pt}
\label{tab:appendix_reasoning_categories_detailed}
\end{table*}

\begin{figure}[htbp]
    \vspace{-10pt}
    \centering
    \includegraphics[width=0.8\linewidth]{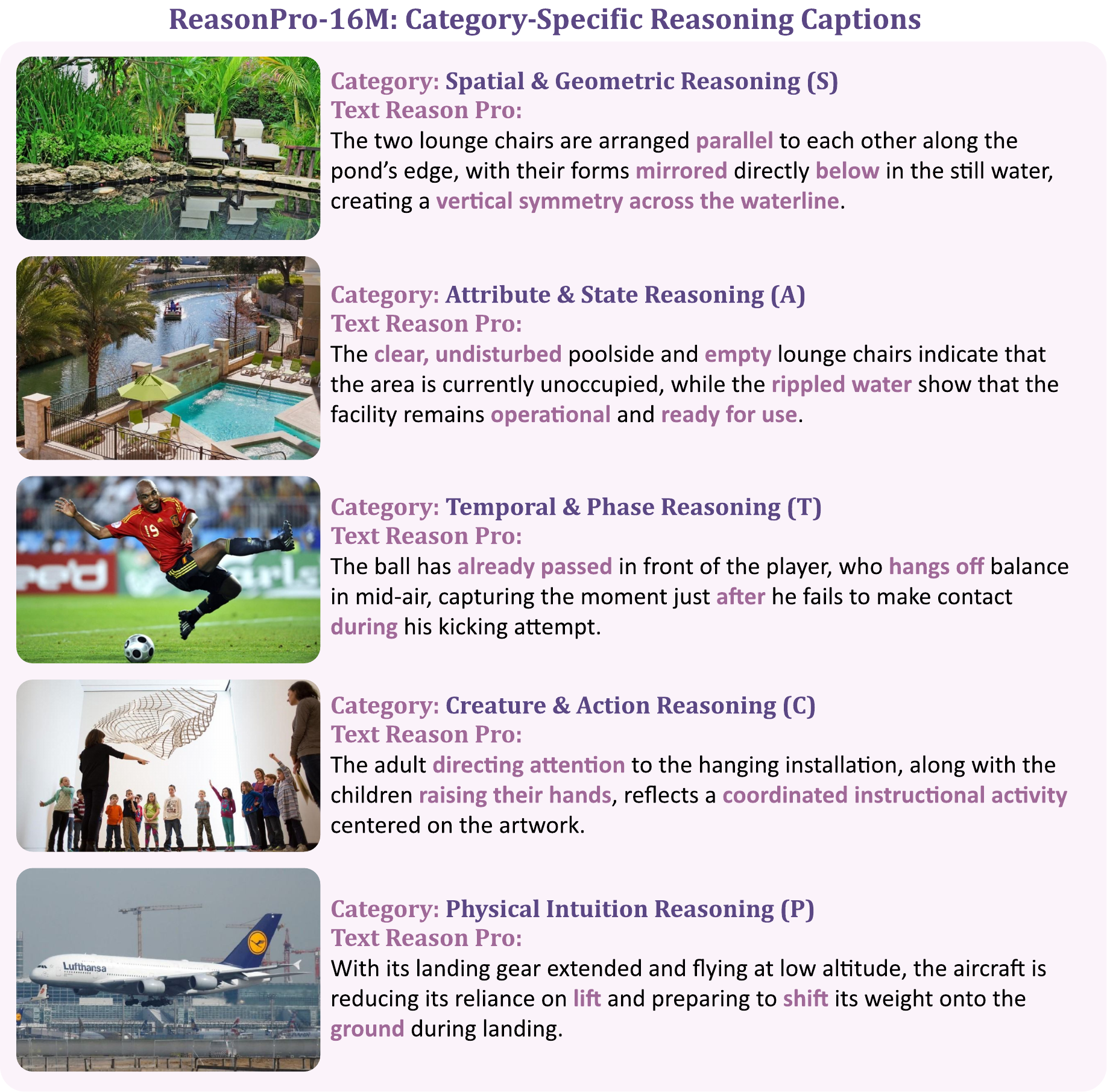}    
    \caption{
        Representative examples for ReasonPro-16M Dataset.
    } 
    \label{fig:app_reasonpro}
    \vspace{-10pt}
\end{figure}

\noindent\textbf{Data Samples.}\quad Correspondingly, we present data samples for each of the five reasoning categories from the ReasonPro dataset in Fig. \ref{fig:app_reasonpro}.

\begin{figure}[t]
    \centering
    \includegraphics[width=\linewidth]{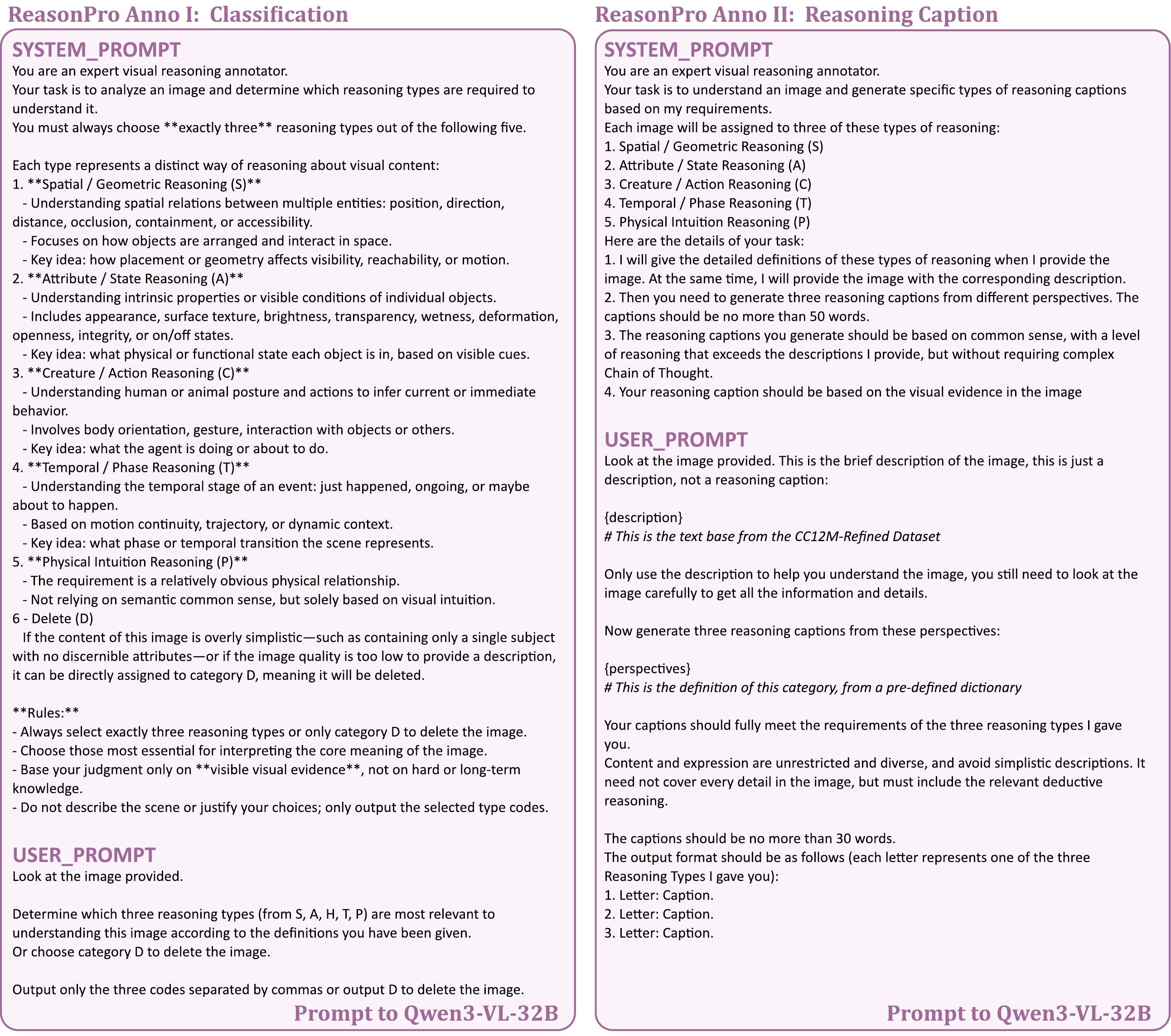}    
    \caption{
        Prompt for ReasonPro dataset generation.
    } 
    \label{fig:app_trp_prompt}
    \vspace{-10pt}
\end{figure}

\noindent\textbf{Prompt.}\quad
Fig. \ref{fig:app_trp_prompt} shows the prompt for ReasonPro Dataset generation.

\subsection{R-CLIP Bench}

\textbf{Data Statistic.} The R-CLIP Bench comprises three datasets: rclip\_5k\_v1, rclip\_5k\_v2, and rclip\_5k\_v3. Each dataset contains 5,000 unique images, and for each image, five distinct reasoning categories (tags) are defined. Each category includes one ground-truth (GT) descriptive sentence and four challenging negative distractors, resulting in a total of 125,000 annotations per dataset version.

The benchmark's design evolved across versions. The initial version, V1, focused on syntactic and structural variations with five tags (`SUBJ\_FORM', `NOUN\_SWAP', `REL\_PHRASE', `ATTR\_STATE', `SENT\_STRUCT'). The subsequent versions, V2 and V3, introduced a more fine-grained, semantic-oriented set of reasoning categories: Spatial (S), Attribute (A), Creature (C), Temporal (T), and Physical (P). This evolution is reflected in the sentence complexity; our analysis shows that sentences in V2 and V3 are longer and more descriptive than in V1. As detailed in Table~\ref{tab:rclip_bench_stats}, the mean token count for GT sentences increased from 29.07 in V1 to approximately 33.4 in V2 and V3.

Table~\ref{tab:rclip_bench_per_category} further breaks down the statistics for the semantic categories in V2 and V3, revealing that sentences related to the overall Scene (S) tend to be the longest, while those describing Creature/Actions (C) are the most concise. This detailed annotation allows for a nuanced evaluation of vision-language models' reasoning capabilities.

\begin{table}[h!]
\centering
\caption{Overall statistics for the R-CLIP Bench (GT sentences).}
\label{tab:rclip_bench_stats}
\begin{tabular}{l|cc|cc|c}
\toprule
\textbf{Dataset} & \multicolumn{2}{c|}{\textbf{Word Count}} & \multicolumn{2}{c|}{\textbf{Token Count}} & \textbf{Token/Word Ratio} \\
 & $\mu$ & $\sigma$ & $\mu$ & $\sigma$ & $\mu$ \\
\midrule
rclip\_5k V1 & 25.17 & 4.31 & 29.07 & 5.21 & 1.155 \\
rclip\_5k V2 & 29.70 & 6.26 & 33.41 & 7.23 & 1.125 \\
rclip\_5k V3 & 29.37 & 5.69 & 33.39 & 6.68 & 1.137 \\
\bottomrule
\end{tabular}
\end{table}

\begin{table}[h!]
\centering
\caption{Per-category statistics for R-CLIP Bench V2 and V3 (GT sentences).}
\label{tab:rclip_bench_per_category}
\begin{tabular}{l|cc|cc}
\toprule
\textbf{Category} & \multicolumn{2}{c|}{\textbf{Mean Token Count (V2)}} & \multicolumn{2}{c}{\textbf{Mean Token Count (V3)}} \\
 & $\mu$ & $\sigma$ & $\mu$ & $\sigma$ \\
\midrule
Scene (S) & 40.41 & 6.59 & 39.18 & 5.89 \\
Attribute (A) & 31.45 & 5.21 & 31.62 & 5.01 \\
Creature (C) & 29.18 & 5.58 & 29.33 & 5.23 \\
Temporal (T) & 32.93 & 6.01 & 33.35 & 5.87 \\
Physical (P) & 33.08 & 5.82 & 33.47 & 5.55 \\
\bottomrule
\end{tabular}
\end{table}

\begin{figure}
    \centering
    \includegraphics[width=1\linewidth]{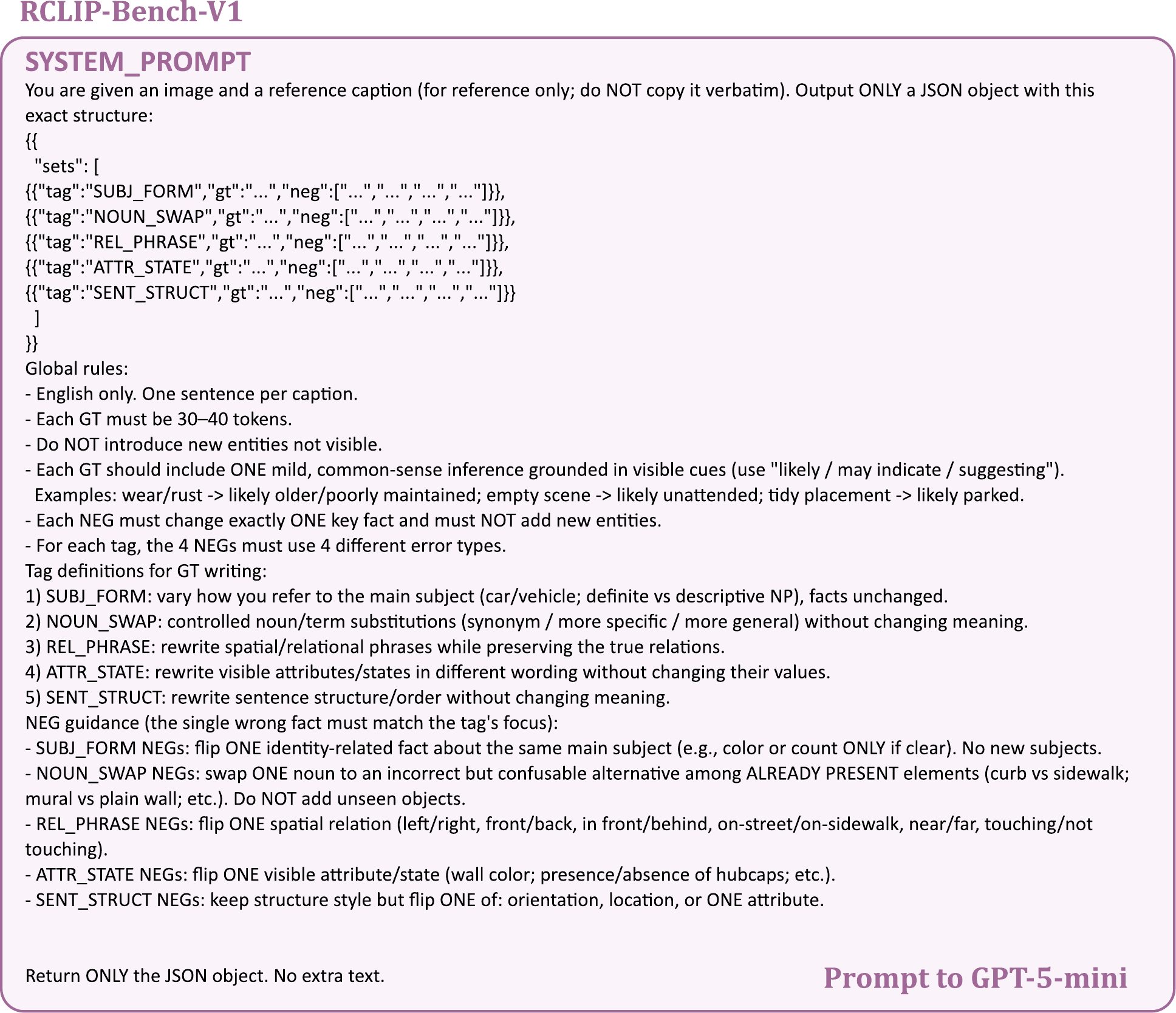}
    \caption{Prompt for RCLIP-Bench V1 generation.}
    \label{fig:rb_v1}
\end{figure}

\begin{figure}
    \centering
    \includegraphics[width=1\linewidth]{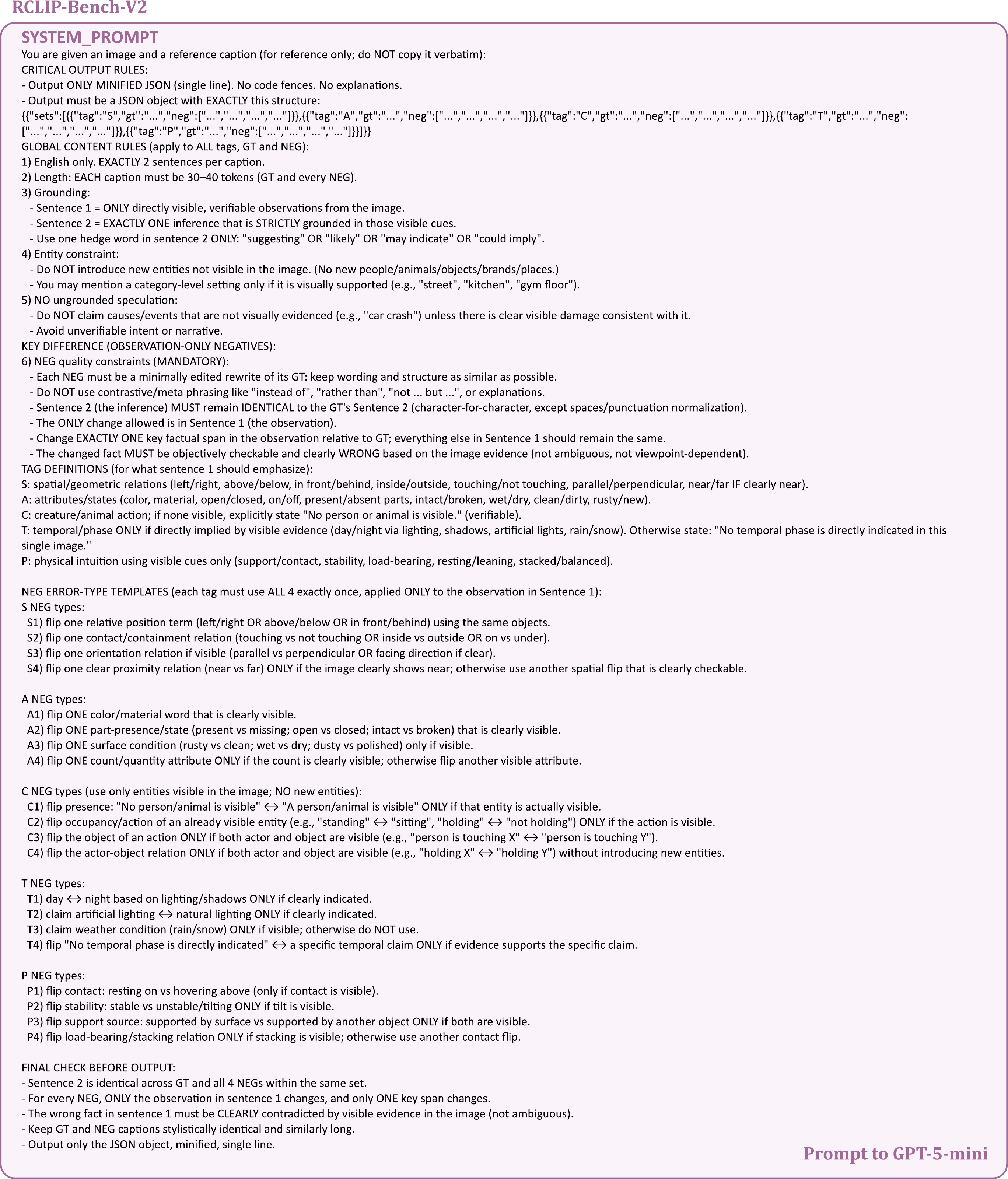}
    \caption{Prompt for RCLIP-Bench V2 generation.}
    \label{fig:rb_v2}
\end{figure}

\begin{figure}
    \centering
    \includegraphics[width=1\linewidth]{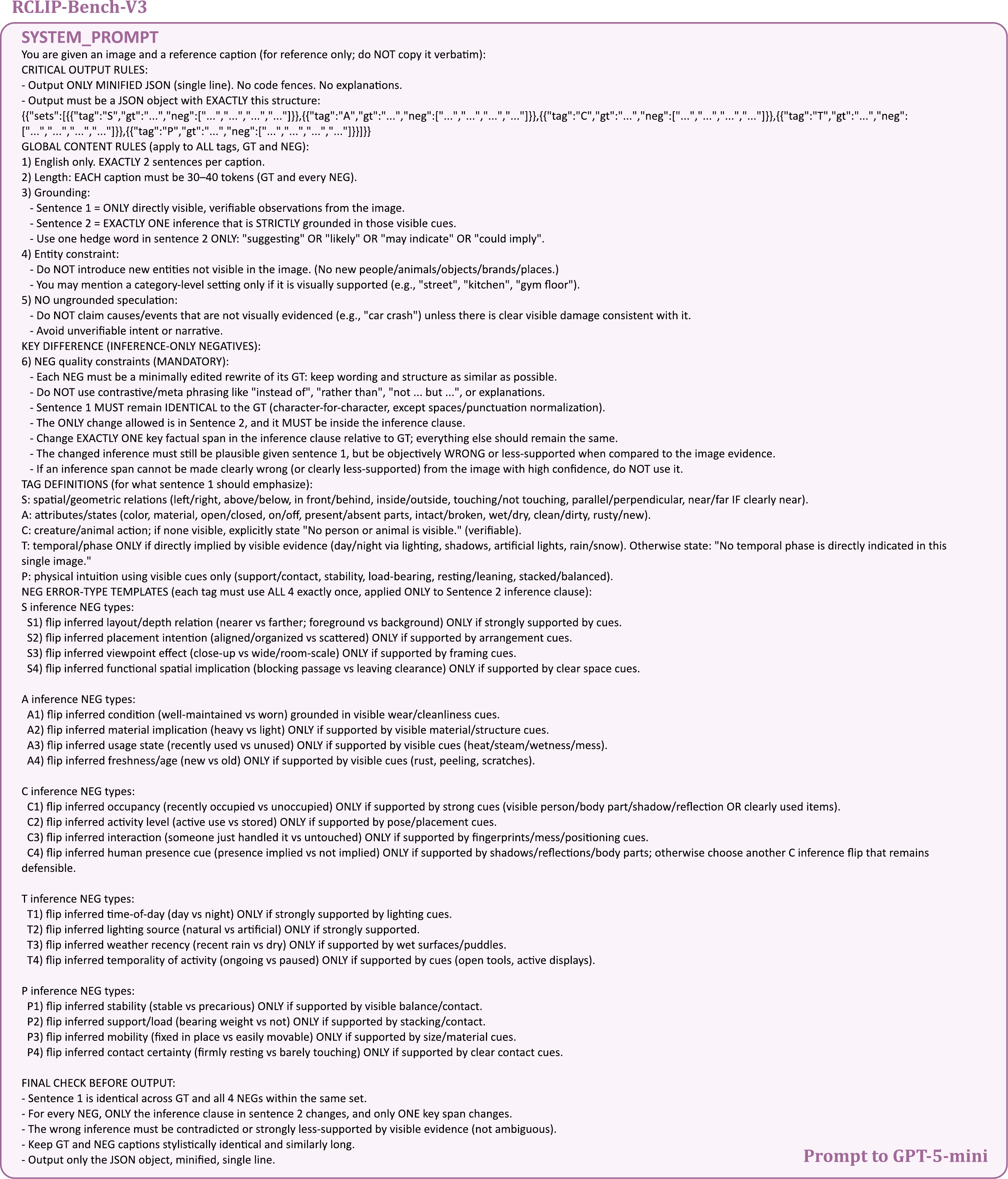}
    \caption{Prompt for RCLIP-Bench V3 generation.}
    \label{fig:rb_v3}
\end{figure}

\textbf{Prompt.} To ensure high-quality and controlled data generation for our benchmarks, we designed a series of detailed prompts for the language model. These prompts enforce strict rules on content, structure, length, and the types of errors introduced in negative examples. Fig. ~\ref{fig:rb_v1}, \ref{fig:rb_v2}, and \ref{fig:rb_v3} show the complete prompts used for generating the V1, V2, and V3 datasets, respectively.
\newpage
\section{Quality Control}
\label{appendix: Quality Control}
\subsection{Data Source}
We choose CC12M \cite{changpinyo2021conceptual} as our sole training data source because it provides a reasonable balance between dataset scale and image quality: although it is neither the largest nor the highest-quality dataset available, it is sufficiently good for large-scale pre-training while keeping computational costs manageable. 

In contrast, datasets such as OBELICS \cite{laurencon2023obelics} and WIT \cite{srinivasan2021wit} offer slightly higher image quality but contain long and overly verbose original captions that require re-annotation; feeding such long captions into the annotation models incurs substantial context-length costs, and truncating them prevents fully leveraging their advantages, making them less efficient overall than the more concise image–text pairs in CC12M. Similarly, although LLaVA-ReCap-CC12M \cite{liu2024llavanext} provides re-annotated captions for CC12M, their average length far exceeds the token limits of most vision encoders (e.g., 77 for CLIP and 64 for SigLIP), meaning additional compression or recaption is still needed. 

Meanwhile, datasets such as LAION-400M \cite{schuhmann2021laion} and DataComp \cite{datacomp} are too large to be practical for early-stage experiments, whereas classic datasets like COCO Captions \cite{lin2014microsoft} are too small to support systematic multilingual analysis. Taken together, we consider CC12M a sufficient and reasonable starting point under our current resources and objectives.
\subsection{Human Annotation}
To ensure the reliability of large-scale data generation and prevent systematic bias, we implemented a rigorous manual pilot validation process prior to formal generation. This process involved randomly sampling 500 initial generations for both the ReasonLite and ReasonPro stages. The evaluation was conducted through independent blind reviews by five graduate students with relevant backgrounds. During the review, annotators strictly filtered out unqualified samples containing visual hallucinations, false causality, multi-step over-extension, and factual inconsistencies with the base descriptive captions. 

In practice, we established a feedback loop: based on typical failure cases flagged by the annotators (e.g., confusing the ``Spatial/Geometric'' and ``Intuitive Physics'' categories in ReasonPro), we iteratively refined and constrained the prompts for the annotation model. As recorded in Tab. \ref{tab:app_human_anno}, a prompt version was locked in for large-scale generation only when the revised prompt achieved a pass rate strictly exceeding 99.5\% on a newly generated set of 500 samples, as comprehensively assessed by the five reviewers. Furthermore, after the large-scale generation task was completed, the team randomly inspected another 500 samples for a post-hoc check. The final overall pass rate for both stages stably remained above 99.0\%, demonstrating the effectiveness of this quality control strategy.

\begin{table*}[htbp]
  \centering
  \caption{Manual Pilot Validation and Prompt Iteration Record.}
  \label{tab:app_human_anno}
  \resizebox{\linewidth}{!}{
  \begin{tabular}{@{}llp{8cm}cl@{}}
    \toprule
    \textbf{Stage} & \textbf{Iteration} & \textbf{Optimization \& Adjustment} & \textbf{Pass Rate} & \textbf{Status} \\
    \midrule
    \multirow{3}{*}{\textbf{ReasonLite}} 
    & Round 1 & Initial test. The model occasionally performed multi-step reasoning without visual evidence. & 86.4\% & Rejected \\
    & Round 2 & Added negative constraints. Fixed most over-extension issues. & 95.2\% & Rejected \\
    & \textbf{Round 3} & Fine-tuned constraint: Forced reasoning statements to factually align with the base text $T_{b}$. & \textbf{99.6\%} & \textbf{Locked} \\
    \midrule
    \multirow{2}{*}{\textbf{ReasonPro}} 
    & Round 1 & Initial test. Found ambiguous boundaries between categories. & 89.8\% & Rejected \\
    & \textbf{Round 2} & Reinforced strict, mutually exclusive definitions for the five categories in the prompt. & \textbf{99.8\%} & \textbf{Locked} \\
    \midrule
    \textbf{Post-Anno} & \textbf{Final} & Random inspection after large-scale generation (mixed data from both stages). & \textbf{99.2\%} & \textbf{Passed} \\
    \bottomrule
  \end{tabular}}
  \vspace{-10pt}
\end{table*}
\subsection{Auto Annotation}
\noindent \textbf{Automated Rule-Based Filtering.} \quad
Following the large-scale automated generation, to further ensure data quality at a minimal computational cost and to eliminate low-quality or degenerate outputs potentially produced by annotation models, we implemented a rule-based filtering pipeline at the end of the process. Specifically, as shown in Tab. \ref{tab:app_auto_anno}, we established four core filtering dimensions. First, to maintain the linguistic consistency of the dataset, we removed all generated texts detected as non-English. Second, to prevent the model from falling into generative loops or producing meaningless redundant expressions, we discarded samples containing obvious fragment repetitions. Finally, we imposed strict upper and lower bounds on the token length of the texts, removing captions shorter than 10 tokens (which typically lack sufficient visual reasoning details) and those longer than 60 tokens (exceeding the model's processing capacity).
\begin{table*}[htbp]
  \centering
  \caption{Automated Rule-Based Filtering Criteria}
  \label{tab:app_auto_anno}
  \resizebox{\linewidth}{!}{
  \begin{tabular}{@{}llp{8cm}@{}}
    \toprule
    \textbf{Dimension} & \textbf{Trigger Condition} & \textbf{Purpose} \\
    \midrule
    \textbf{Language Consistency} 
    & \texttt{detect\_language(text) != 'en'} 
    & Filters out occasional non-English or mixed-language outputs generated by the LLM, maintaining strict English alignment. \\
    \hline
    \textbf{Minimum Length} 
    & \texttt{token\_count(text) < 10} 
    & Removes overly short, invalid responses. \\
    \hline
    \textbf{Maximum Length} 
    & \texttt{token\_count(text) > 60} 
    & Prevents over-extension or the generation of irrelevant, verbose hallucinations. \\
    \hline
    \textbf{Degeneration/Repetition} 
    & Contains repeated or looping sentences 
    & Intercepts common autoregressive degeneration phenomena in LLMs (e.g., "the cat is on the on the on the ..."). \\
    \bottomrule
  \end{tabular}
  }
\end{table*}

\subsection{Annotation Model Selection}
\noindent 1) \textit{Why choose Qwen2.5-VL-72B in ReasonLite?}

ReasonLite was created before the release of Qwen3-VL series, and at that time Qwen2.5-VL-72B was one of the strongest open-source vision-language models available.

\noindent 2) \textit{Why not use QvQ-72B-preview instead of Qwen2.5-VL-72B in ReasonLite?}

QvQ-72B-preview was not adopted because its reasoning capability was not sufficiently stable in our preliminary tests, and, despite explicit constraints on output length, it frequently produced excessively long captions.

\noindent 3) \textit{Why not using Qwen3-VL-235B-A22B in ReasonPro?}

Qwen3-VL-32B already meets our requirements for commonsense reasoning, and in our experiments scaling up to Qwen3-VL-235B-A22B significantly increased computational cost without yielding consistently better annotation quality. 

\noindent 4) \textit{Why not using the thinking version?}

For the same reason as in QA 3), the reasoning level required in our task does not demand the use of models operating in “thinking” mode. Such models incur substantially higher computational costs and often produce overly elaborate reasoning-style captions, which are not suitable for the CLIP-style models. 

\subsection{Copyright and Ethical Statements}
The CC12M dataset is released under a permissive license that allows free use for any purpose, although acknowledgment of Google LLC as the data source is appreciated. Our work uses CC12M following the same terms: \textbf{we access images only via provided URLs and do not redistribute any images.} As such, our usage does not involve copyright infringement or redistribution of third-party media.

Our study is purely methodological and is conducted entirely on existing publicly available image datasets. We do not collect new data or introduce additional personal, social, or ethical risks beyond those already associated with the original dataset.
\newpage
\section{Implementation Details}
\label{appendix: Training Details and More Results}

\subsection{Stage 0: Baseline Continual Pretraining}
\noindent\textbf{Training Parameters and Cost.}
In Stage 0 training, we employ identical hyperparameters for both the Description and Reasoning baselines, but utilize different training data. Specifically, the Description Baseline is trained using all base descriptive captions ($T_{b}$) from CC12M-Refined, whereas the Reasoning Baseline utilizes all reasoning captions ($T_{rl}$ and $T_{rp}$). Detailed training parameters are provided in Table \ref{tb:train_para_s0}.

For \textit{S0-Des.}, SigLIP2-So/14-384 training time on 32 GPUs is around 4 hours per epoch, while CLIP-L/14-224 training time is around 1.5 hours per epoch. \textit{S0-Rea.}'s consumption time is approximately 1.2 times that of \textit{S0-Des.}. 
\begin{table}[htbp]
\centering
\caption{Training Configuration for Stage\,0 Description (\textit{S0-Des.}) Continue Pretraining.}
\label{tb:train_para_s0}
\resizebox{0.9\linewidth}{!}{
\begin{tabular}{lcc}
\toprule
\textbf{Architecture} & SigLIP & CLIP \\
\midrule
Version & Huggingface Weight & Huggingface Weight \\
Scale  & siglip-so400m-patch14-384 & clip-vit-large-patch14 \\
Layers (Vision / Text) & 27 / 27 & 24 / 12 \\
Embedding Size (Vision / Text) & 1152 / 1152 & 1024 / 768 \\
Optimizer & AdamW & AdamW \\
AdamW $\beta_{1}$ & 0.9 & 0.9\\
AdamW $\beta_{2}$ & 0.999 & 0.999\\
AdamW $\epsilon$ & $1 \times 10^{-8}$ & $1 \times 10^{-8}$ \\
Lr (Vision / Text / Logit) & $1\times10^{-5}$ / $3\times10^{-5}$ / $5\times10^{-4}$ & $1\times10^{-5}$ / $3\times10^{-5}$ / $5\times10^{-4}$ \\
Weight Decay & $1\times10^{-4}$ & $1\times10^{-4}$  \\
Scheduler Type & Cosine & Cosine \\
Warm-up Ratio & 0.10 & 0.10 \\
Per-device Batch Size & 384 & 512 \\
Gradient Accumulation Steps & 2 & 2 \\
Effective Batch Size & 24{,}576 & 32{,}768 \\
Precision & bfloat16 & bfloat16 \\
Training GPUs & 32 (8 nodes $\times$ 4 GPUs) & 32 (8 nodes $\times$ 4 GPUs) \\
Total Samples & 31.2M (CC12M-Refined) & 31.2M (CC12M-Refined) \\
\bottomrule
\end{tabular}
}
\end{table}

\subsection{Stage 1: Reasoning-Aware Alignment}
\noindent\textbf{Training Parameters and Cost.}
In the first stage, we conducted training for 1 epochs on the ReasonLite-42M Dataset. Table \ref{tb:train_para_s1} reports the training parameters. For CLIP-L/14-224 and CLIP-L/14-336, the same training configuration is adopted, with the only difference being the device batch size, which was set to 512. For CLIP-B/32, we used a device batch size of 768. We use Flash Attention 2 \cite{dao2023flashattention2} for SigLIP2 and PyTorch SDPA for CLIP. For reference, in this stage, SigLIP2-So/14-384 training time on 32 GPUs is around 7.5 hours per epoch, while CLIP-L/14-224 training time is around 3.85 hours per epoch.

\begin{table}[htbp]
\centering
\caption{Training Configuration for Stage\,1.}
\label{tb:train_para_s1}
\resizebox{0.9\linewidth}{!}{
\begin{tabular}{lcc}
\toprule
\textbf{Architecture} & SigLIP & CLIP \\
\midrule
Version & Huggingface Weight & Huggingface Weight \\
Scale  & siglip-so400m-patch14-384 & clip-vit-large-patch14 \\
Layers (Vision / Text) & 27 / 27 & 24 / 12 \\
Embedding Size (Vision / Text) & 1152 / 1152 & 1024 / 768 \\
Optimizer & AdamW & AdamW \\
AdamW $\beta_{1}$ & 0.9 & 0.9\\
AdamW $\beta_{2}$ & 0.999 & 0.999\\
AdamW $\epsilon$ & $1\times 10^{-8}$ & $1\times 10^{-8}$ \\
Lr (Vision / Text / Logit) & 1$\times$10$^{-5}$ / 3$\times$10$^{-5}$/ 1$\times$10$^{-4}$ & 1$\times$10$^{-5}$ / 3$\times$10$^{-5}$/ 1$\times$10$^{-4}$ \\
Weight Decay & 0.05 & 0.05 \\
Scheduler Type & Cosine & Cosine \\
Warm-up Ratio & 0.10 & 0.10 \\
Per-device Batch Size & 384 & 512 \\
Gradient Accumulation Steps & 2 & 2 \\
Effective Batch Size & 24{,}576 & 32{,}768 \\
Precision & bfloat16 & bfloat16 \\
Training GPUs & 32 (8 nodes × 4 GPUs) & 32 (8 nodes × 4 GPUs) \\
Total Samples & 42.3M (ReasonLite) & 42.3M (ReasonLite) \\
L2 Reg. (w.r.t. Ori. Model) & Yes & Yes \\
coefficient $\beta$ & 1$\times$10$^{-5}$ & 1$\times$10$^{-5}$ \\
coefficient $\lambda$ ($T_b$ : $T_{rl}$) &
\begin{tabular}[c]{@{}l@{}}
0--20\%: (0.7, 0.3) \\
20--80\%: linearly $\downarrow$ to (0.5, 0.5) \\
80--100\%: fixed at (0.6, 0.4)
\end{tabular}
&
\begin{tabular}[c]{@{}l@{}}
0--20\%: (0.6, 0.4) \\
20--80\%: linearly $\downarrow$ to (0.3, 0.7) \\
80--100\%: fixed at (0.5, 0.5)
\end{tabular}\\
\bottomrule
\end{tabular}
}
\end{table}

\subsection{Stage 2: Explicit Reasoning Supervision}
\noindent\textbf{Training Parameters and Cost.}
In Training Stage 2, models are trained on ReasonPro-16M Dataset for 1 epoch. Table \ref{tb:train_para_s2} reports the training parameters. For CLIP-L/14-224 and CLIP-L/14-336, the same training configuration is adopted, with the only difference being the device batch size, which was set to 512. For CLIP-B/32, we used a device batch size of 768. For reference, in this stage, SigLIP2-so/14-384 training time on 32 GPUs is around 3.5 hours per epoch, while CLIP-L/14-224 training time is around 1 hour per epoch.

\noindent\textbf{Reasoning Classifier.} Table \ref{tab:reasoning-classifier-arch} shows the architecture of classifiers in Stage 2 training.

\begin{table}[htbp]
\centering
\caption{Training Configuration for Stage 2.}
\resizebox{0.9\linewidth}{!}{
\begin{tabular}{lcc}
  \toprule
  \textbf{Architecture} & SigLIP & CLIP \\
  \midrule
  Version & ReasonSigLIP-S1 & ReasonCLIP-S1 \\
  Scale  & siglip-so400m-patch14-384 & clip-vit-large-patch14 \\
  Lr (Vision / Text / Logit)  & 1$\times$10$^{-5}$ / 2$\times$10$^{-5}$/ 1$\times$10$^{-4}$ & 1$\times$10$^{-5}$ / 3$\times$10$^{-5}$/ 1$\times$10$^{-4}$ \\
  Weight Decay & 0.10 & 0.10 \\
  Warm-up Ratio & 0.10 & 0.10  \\
  Per-device Batch Size & 512 & 512  \\
  Gradient Accumulation Steps & 2 & 2 \\
  Effective Batch Size &  32{,}768 & 32{,}768 \\
  Lr (Classifier) &  1.5$\times$10$^{-3}$ & 1.5$\times$10$^{-3}$  \\
  Coefficient $\gamma$ & 0.05 & 0.10  \\
  L2 Reg. (w.r.t. S1 Model) & No & No \\
  \multirow{5}{*}{Total Samples}
    & 4.8M (Spatial/Geometric) & 4.8M (Spatial/Geometric)\\
    & 4.9M (Attribute/State) & 4.9M (Attribute/State)\\
    & 2.5M (Creature/Action) & 2.5M (Creature/Action) \\
    & 1.0M (Temporal/Phase) & 1.0M (Temporal/Phase)\\
    & 3.5M (Physical Intuition) & 3.5M (Physical Intuition)\\
  Training GPUs & 32 (8 nodes $\times$ 4 GPUs) & 32 (8 nodes $\times$ 4 GPUs)  \\
  \bottomrule
\end{tabular}
}
\label{tb:train_para_s2}
\end{table}

\begin{table}[t]
\centering
\caption{Reasoning classifier architecture.}
\label{tab:reasoning-classifier-arch}
\resizebox{0.65\linewidth}{!}{
\begin{tabular}{ll}
\toprule
\textbf{Component} & \textbf{Configuration} \\
\midrule
Text classifier & $\mathrm{Linear}(D, 5)$ \\
Image classifier & $\mathrm{Linear}(D, 5)$ \\
Input feature dimension $D$ & Backbone embedding dimension \\
Input features & L2-normalized text/image embeddings \\
Number of classes & 5 \\
Bias term & Enabled \\
Data type & \texttt{bfloat16} \\
\bottomrule
\end{tabular}}
\end{table}

\subsection{Hyperparameter Optimization}
In the ablation study (Table 7) of the main text, we explore the impact of different dynamic weight scheduling strategies on model performance during Stage 1. In this section, we provide the explicit parameters. 

During Stage 1, the model receives supervision signals from both the base descriptive text $T_b$ and the reasoning text $T_{rl}$, with their respective weights denoted as $\lambda(t)$ and $1-\lambda(t)$. To balance the preservation of the original semantic alignment with the injection of new reasoning capabilities, we designed a piecewise dynamic weight scheduling strategy and compared three specific configurations in our ablation study. 

Our default strategy (Ours) begins with a bias toward the descriptive text (0.6, 0.4), linearly transitions to (0.3, 0.7) during the 20\%--80\% phase of training, and maintains a balance (0.5, 0.5) for the final 20\%. In contrast, \textbf{Des.-heavy} maintains a higher descriptive weight throughout (decreasing from (0.8, 0.2) to (0.5, 0.5), and finally fixing at (0.6, 0.4)), which restricts the injection of reasoning capacity. Conversely, \textbf{Rea.-heavy} shifts too aggressively toward the reasoning text (decreasing from (0.4, 0.6) to (0.1, 0.9), and finally fixing at (0.2, 0.8)), thereby disrupting the model's foundational semantic alignment. 

Furthermore, we evaluated an exponential scheduling strategy (\textbf{Exp-Sched}), which applies a continuous exponential decay for the descriptive weight from an initial bias of (0.6, 0.4) down to a final balance of (0.5, 0.5) across the entire training process, bypassing the piecewise stages. However, this non-linear approach proved less stable and less effective than our piecewise linear transition.

\begin{table}[htbp]
  \centering
  \caption{Stage 1 Weight Scheduling Configurations: $(\lambda(t), 1-\lambda(t))$}
  \label{tab:weight_sched}
  \begin{tabular}{@{}lccc@{}}
    \toprule
    \textbf{Config} & \textbf{0\%--20\%} & \textbf{20\%--80\%} & \textbf{80\%--100\%} \\
    \midrule
    \textbf{Ours} & $(0.6, 0.4)$ & Linearly $\downarrow$ to $(0.3, 0.7)$ & Fixed at $(0.5, 0.5)$ \\
    \textbf{Des.-heavy} & $(0.8, 0.2)$ & Linearly $\downarrow$ to $(0.5, 0.5)$ & Fixed at $(0.6, 0.4)$ \\
    \textbf{Rea.-heavy} & $(0.4, 0.6)$ & Linearly $\downarrow$ to $(0.1, 0.9)$ & Fixed at $(0.2, 0.8)$ \\
    \textbf{Exp-Sched} & \multicolumn{3}{c}{Continuous exponential decay from $(0.6, 0.4)$ to $(0.5, 0.5)$} \\
    \bottomrule
  \end{tabular}
\end{table}

In the Stage 2 ablation study, we compared the parameter settings of the following three configuration variants: 
\textbf{w/o cls} removes the classification head modules from both the image and text encoders (i.e., discarding $\mathcal{L}_{img-cls}$ and $\mathcal{L}_{txt-cls}$), retaining only the base image-text alignment loss; 
\textbf{w/ L2} adds an L2 regularization term to the optimization objective to penalize model parameters from deviating too far from the original pre-trained weights; 
\textbf{Single-Label} replaces the multi-label classification (BCE Loss) on the image side with single-label classification (CE Loss), forcing the image and the current reasoning text to strictly share a single, unique category label.

\newpage
\section{More Results}
\subsection{Detailed Evaluation Settings}
\label{Appendix: Detailed Evaluation Settings}
\noindent\textbf{Benchmarks.}
For all benchmarks, we prioritize using the versions provided by CLIP-bench on Hugging Face whenever available. If unavailable, we utilize the official versions and conduct a fair re-evaluation across all models.

\noindent\textbf{Models.}
For evaluating existing models, we prioritize using official weight paths to ensure fairness. For example, for the SigLIP series, we prefer using the Hugging Face version of the weights over the OpenCLIP version.

\subsection{Detailed Results on Zero-shot Retrieval and Zero-shot Classification Benchmarks}
The results are shown in \ref{tab:app: retrieval and cls}. Although ReasonCLIP demonstrates consistent improvements across visually grounded reasoning and retrieval benchmarks, we observe a slight performance degradation in standard zero-shot classification tasks. Standard zero-shot classification is fundamentally object-centric, relying on short templates to recognize isolated entities. In contrast, our reasoning-oriented continual pretraining shifts the model's focus toward relation- and context-centric understanding, which requires encoding complex syntax, actions, and physical states, etc. Consequently, while the backbone's capacity for reasoning is significantly enhanced, its absolute sensitivity to fine-grained, single-object discrimination experiences a marginal compromise.
\label{appendix: Results: Retrieval}
\begin{table*}[htbp]
\centering
\caption{Full ablation results on zero-shot retrieval and zero-shot classification benchmarks of different training stages and direct continual pretraining.}
\label{tab:app: retrieval and cls}
\setlength{\tabcolsep}{2pt}
\resizebox{\linewidth}{!}{
\begin{tabular}{l
|ccc|ccc     
|ccc|ccc
|ccc|ccc
|ccc|ccc
|cccc}    
\toprule
& 
\multicolumn{6}{c|}{COCO-5K} &
\multicolumn{6}{c|}{Flickr-30K} &
\multicolumn{6}{c|}{Urban-1K} &
\multicolumn{6}{c|}{RCLIP-5K-V3} &
 \multicolumn{4}{c}{ImageNet} \\
Model &
\multicolumn{3}{c}{I $\rightarrow$ T} & \multicolumn{3}{c|}{T $\rightarrow$ I} &
\multicolumn{3}{c}{I $\rightarrow$ T} & \multicolumn{3}{c|}{T $\rightarrow$ I} &
\multicolumn{3}{c}{I $\rightarrow$ T} & \multicolumn{3}{c|}{T $\rightarrow$ I} &
\multicolumn{3}{c}{I $\rightarrow$ T} & \multicolumn{3}{c|}{T $\rightarrow$ I} &\\
&
R@1 & R@5 & R@10 & R@1 & R@5 & R@10 &
R@1 & R@5 & R@10 & R@1 & R@5 & R@10 &
R@1 & R@5 & R@10 & R@1 & R@5 & R@10 &
R@1 & R@5 & R@10 & R@1 & R@5 & R@10 &
val & v2 & Obj. & Ske.\\
\bottomrule
\rowcolor{gray!20}
\multicolumn{29}{l}{\textit{Scale - ViT-Base/32 @224 (86M)}} \\
CLIP 
& 50.0 & 75.0  & 83.4 & 30.4   & 56.0  & 66.9 
& 40.6 & 64.7  & 73.7  & 21.7  & 41.6  &51.1 
& 61.0 & 84.8 & 90.8 & 46.8 & 72.8 & 79.9 
& 51.0 & 75.3 & 84.1 & 26.7 & 47.2 & 55.5 
& \textbf{63.4}  &  \textbf{55.8} & \textbf{44.1} & \textbf{42.3}  \\

+ Stage 1 
& \textbf{56.2} & \underline{78.8} & \textbf{86.8} & \textbf{37.9} & \textbf{64.1} & \textbf{74.5} 
& \textbf{42.4} & \textbf{67.9} & \textbf{77.2} & \textbf{29.2} & \textbf{51.1} & \textbf{60.7} 
& \textbf{70.4} & \textbf{91.6} & \textbf{95.4} & \textbf{68.6} & \textbf{90.2} & \textbf{94.5} 
& 56.0 & 79.3 & 87.0 & 30.4 & 51.5 & 59.4 
& \underline{60.6}  &  52.9 & \underline{41.7} & \underline{41.6}  \\

+ Stage 2 
&  52.3 &  77.2 &  85.0 & 37.0  & 62.6  & 72.8 
&  \underline{41.9} &  \underline{66.4} & \underline{75.7}  &  27.5 & 49.0  & 58.4 
&  59.2 & 83.0 & 89.5 & 60.4 & 83.9 & 89.8 
& 55.3 & \underline{80.5} & \underline{88.0} & \underline{33.8} & \underline{56.0} & \underline{64.0} 
& 57.8 & 51.0 & 38.6 & 37.7 \\

Description CP 
& \underline{54.7} & \textbf{79.0} & \underline{86.2} & \underline{37.3} & \underline{63.1} & \underline{73.3} 
& 41.5 & \underline{66.4} & \underline{75.7} & \underline{28.2} & \underline{49.9} & \underline{59.5} 
& \underline{68.5} & \underline{90.5} & \underline{94.6} & \underline{66.9} & \underline{89.2} & \underline{93.2} 
& \underline{56.4} & 79.4 & 87.2 & 28.7 & 48.4 & 56.1 
& 59.5 & 51.7 & 40.8 & 40.4  \\

Reasoning CP 
& 53.4 & 78.0  & 85.8  & 36.5 &  62.3 & 72.4 
& 41.3 & 66.0  & 75.3  &26.0   & 46.8  & 56.3 
& 66.1 & 88.6 & 93.8 & 65.6 & 88.2 & 92.1 
&\textbf{58.9} & \textbf{82.5} & \textbf{89.9} & \textbf{34.1} & \textbf{56.5} & \textbf{64.3} 
& 59.9  & \underline{53.0}  & 40.0 & 40.8  \\
\bottomrule

\rowcolor{gray!20}
\multicolumn{29}{l}{\textit{Scale - ViT-Large/14 @224 (307M)}} \\
CLIP 
&  56.3 & 79.5  & 86.4  & 36.6  & 61.2  & 71.1 
&  48.8 &  73.0 & 81.1  & 28.2  &  49.6 & 59.0 
&  68.5 & 88.9 & 94.3 & 55.9 & 80.4 & 86.3 
& 54.0 & 77.9 & 86.1 & 31.7 & 52.3 & 59.5 
&  \textbf{75.6} &  \textbf{69.9} & \textbf{69.0} &  \textbf{59.6} \\

+ Stage 1 
& \textbf{64.5} & \textbf{85.4}  & \textbf{91.1}  & \underline{46.7}  & \textbf{71.6}  &\textbf{80.7} 
&\textbf{60.0}  &  \textbf{82.5} & \textbf{88.8}  & \textbf{40.9}  &\textbf{63.7} & \textbf{72.4} 
& \textbf{80.5} & \textbf{96.2} & \textbf{98.2} & \underline{79.7} & \underline{94.2} & \underline{96.8} 
&66.3 & 86.2 & 92.1 & 38.0 & 58.4 & 65.1 
& \underline{73.6}  & 67.3  & \underline{67.3} &  \underline{59.1} \\

+ Stage 2 
& 60.1  & 82.4  & 89.5  & 46.2  & 70.7  &79.7 
& 55.6 & 78.8  & 85.8  & 39.0  &61.8   &70.6 
& 74.3 & 92.9 & 95.6 & 75.5 & 91.5 & 95.5 
&  67.1 & \underline{89.1} & \underline{94.1} & \textbf{41.9} & \textbf{63.1} & 69.3 
&  70.4 & 64.5 & 62.3 & 53.4  \\

+ Stage 2 w/o cls 
& 60.3  & 82.6  & 89.7  & 46.3  & 70.8 & 79.9 
&  55.7 & 79.0  &  85.9 & 39.2  & 62.0  & 70.8 
& 75.6 & 92.6 & 95.5 & 75.4 & 92.0 & 95.0 
&\underline{67.3} & \underline{89.1} & \underline{94.1} & \textbf{41.9} & \textbf{63.1} & \textbf{69.5} 
& 70.5  &64.8   & 62.2 &  53.6 \\

Description CP 
&  \underline{64.0} &  \underline{84.7} & \underline{91.0}  & 46.3  & 71.3  & 80.0 
&  \underline{56.1} & \underline{79.3}  &  \underline{86.5} &  \underline{40.7} &  \underline{63.4} & \underline{72.0} 
&  \underline{80.1} & \underline{94.5} & 97.3 & 79.1 & 93.6 & 96.7 
&  66.0 & 86.8 & 92.2 & 36.2 & 55.5 & 62.3 
&73.3 & \underline{67.4} & 66.8 & 58.0  \\

Reasoning CP 
& 61.2 & 83.2  & 89.6 & \textbf{46.8}  & \underline{71.4}  & \underline{80.3} 
& 54.5 &78.1   & 85.5  & 40.0  & 62.9  &71.6 
&  79.6 & 94.4 & \underline{97.5} & \textbf{80.7} & \textbf{94.4} & \textbf{97.0} 
& \textbf{68.4} & \textbf{89.4} & \textbf{94.2} & \underline{41.8} & \underline{62.9} & \underline{69.4} 
&  72.6 & 66.6  &65.3  & 56.7  \\

\bottomrule
\rowcolor{gray!20}
\multicolumn{29}{l}{\textit{Scale - ViT-Large/14 @336 (307M)}} \\
CLIP 
& 58.0 & 80.9  & 87.9  &  37.0 & 61.6 & 71.6 
& 51.5  & 75.5  &  83.4 &  31.6 &  53.6 & 62.7 
& 73.0 & 90.3 & 95.2 & 57.0 & 79.5 & 86.3 
& 55.9 & 79.5 & 86.6 & 33.2 & 53.6 & 60.7 
& \textbf{76.6}  &  \textbf{71.0} & \textbf{72.0} & \textbf{61.0}  \\

+ Stage 1 
& \textbf{65.1} & \underline{85.7}  & \underline{91.3}  &  46.9 & \underline{71.7}  &\underline{80.6} 
& \textbf{60.9}  &  \textbf{83.1} & \textbf{89.5}  & \textbf{42.0}  &\textbf{64.8} & \textbf{73.2} 
&\textbf{83.0} & \textbf{96.3} & \textbf{98.2} & \underline{81.9} & \textbf{94.5} & \textbf{97.3} 
&  \underline{67.2} & \underline{86.7} & \underline{92.7} & \underline{38.5} & \underline{58.7} & \underline{65.3} 
& \underline{74.3}  &  \underline{68.3} & 69.1 & \underline{60.0}  \\

+ Stage 2 
&  61.3 & 83.0  & 89.6  & \underline{47.1}  & 71.3  & 80.4 
& 57.2 & 79.8  & 86.8  & 40.8  & 63.4  & 72.0 
&  76.5 & 91.9 & 95.2 & 78.3 & \underline{93.2} & 96.2 
& \textbf{68.1} & \textbf{89.7} & \textbf{94.9} & \textbf{42.9} & \textbf{63.4} & \textbf{69.6} 
& 70.8 & 65.2 & 64.0 & 53.7  \\

Description CP 
& \underline{64.6} & \textbf{86.0} & \textbf{91.7} & 46.7 & \underline{71.7} & 80.3 
&  \underline{57.3} & \underline{80.3} & \underline{87.5} & \underline{41.6} & \underline{64.3} & \underline{72.9} 
& 81.5 & 95.1 & \underline{97.6} & \textbf{82.0} & \textbf{94.5} & 97.1 
&56.4 & 79.4 & 87.2 & 28.7 & 48.4 & 56.1 
& 74.2 & 68.2 & \underline{69.5} & 59.3\\

Reasoning CP 
& 62.4 & 83.7  & 90.1  & \textbf{47.5}  &  \textbf{72.0} & \textbf{80.7} 
& 56.2 & 79.3  & 86.4  & 41.2  & 64.2   & \underline{72.9} 
& \underline{81.8} & \underline{95.8} & 97.3 & \textbf{82.0} & \textbf{94.5} & \underline{97.2} 
&58.9 & 82.5 & 89.9 & 34.1 & 56.5 & 64.3 
&  73.0 &  67.3 & 66.8 & 57.2  \\

\bottomrule
\rowcolor{gray!20}
\multicolumn{29}{l}{\textit{Scale - ViT-So400m/14 @384 (428M)}} \\
SigLIP 
& 72.6  & 90.1 & 94.4  & 54.3  & 76.8  & 84.2 
& \underline{69.9} & \underline{89.3} & \underline{93.9}  & 51.3 & 73.0  & 80.2 
& 74.5 & 91.9 & 94.9 & 73.4 & 88.3 & 92.4 
& 70.5 & 86.3 & \textbf{90.2} & 43.4 & 61.6 & 66.8 
& \textbf{83.1}  & \textbf{77.3}  & \textbf{77.0} & \textbf{74.7}  \\

+ Stage 1 
&  \underline{73.6} &  \textbf{91.2} & \textbf{94.8}  &  56.5 & 79.2  & 86.2 
&  69.3 & 88.8  & 93.8  & 52.3  &  73.8 & 80.9 
& 77.5 & 93.1 & 95.7 & \underline{82.3} & \textbf{94.5} & \textbf{97.0} 
&  68.3 & 84.4 & 87.7 & 45.5 & 63.7 & 69.3 
& 80.8 & \underline{74.8} & 74.6 & \underline{73.5} \\

+ Stage 2 
&  \textbf{73.7} & \underline{90.8}  & \textbf{94.8}  &  \underline{57.3} &  79.9  &  \underline{86.9} 
& \textbf{73.3}  &  \textbf{90.0} & \textbf{94.3}  & 53.9  &  75.3 & 82.3 
&  78.6 & 92.3 & \textbf{96.1} & 80.6 & 93.8 & 96.5 
& \underline{72.4} & \underline{86.4} & 89.5 & 49.5 & 67.8 & 73.0 
& 79.6 & 73.9 & 68.9 & 71.1  \\

+ Stage 2 w/o cls 
& 71.8 & 89.0 & 93.8 & \textbf{58.2} & \underline{80.7} & \textbf{87.6} 
& 69.1 & 88.6 & 93.4 & \underline{54.6} & \underline{76.2} & \underline{83.0} 
& 75.5 & 91.1 & 94.1 & 79.8 & 93.6 & 95.6 
& \textbf{72.7} & \textbf{87.0} & \underline{89.9} & \textbf{50.1} & \textbf{68.5} & \textbf{73.7} 
& 79.7 & 73.8 & 72.1 & 70.2  \\

Description CP 
& 73.1  &  \underline{90.8} & \underline{94.6}  & 56.6  & 79.4  & 86.4 
& 68.8  & 88.5 & 93.4  & 52.7  & 74.2  & 81.2 
& \textbf{79.8} & \underline{93.4} & \underline{96.0} & \textbf{82.8} & \underline{94.2} & \underline{96.9} 
& 69.7 & 85.1 & 88.2 & 43.3 & 61.3 & 66.7 
& \underline{81.1} & 74.6 & \underline{75.9} & 72.8 \\

Reasoning CP 
& 60.6  & 83.2  & 89.9  &  \textbf{58.2} & \textbf{80.8}  &\textbf{87.6} 
& 63.6  & 83.3  & 89.1  & \textbf{55.0}  & \textbf{76.4}  & \textbf{83.2} 
& \underline{79.6} & \textbf{94.1} & \textbf{96.1} & 80.8 & \underline{94.2} & 96.5 
&70.5 & 85.9 & 89.1 & \underline{49.9} & \underline{68.4} & \underline{73.3} 
&  77.6 & 71.1 & 69.4 & 70.8  \\

\hline
SigLIP2 
&  71.2 & 89.7 & \textbf{94.7} & 55.7 & 78.5 & 85.6 
& 69.4 & 88.6 & 93.5 & 52.5 & 73.9 & 81.0 
& 64.1 & 84.7 & 90.6 & 65.0 & 84.1 & 89.6 
&  64.9 & 78.2 & 80.7 & 38.5 & 54.8 & 60.0 
&\textbf{83.2} & \textbf{77.7} & \underline{80.2} & \textbf{75.7}\\

+ Stage 1 
&  \underline{72.8} & \underline{89.8} & 94.0 & 58.2 & 80.4 & 87.1 
& \underline{69.6} & \underline{89.0} & \underline{93.6} & 53.6 & 74.8 & 81.7 
&\underline{75.0} & \underline{92.8} & 95.7 & \underline{75.5} & 91.0 & 94.3 
&  64.5 & 77.7 & 81.0 & 40.5 & \underline{56.7} & 61.8 
& \underline{82.3} & \underline{76.6} & 79.1 & \underline{74.3} \\

+ Stage 2 
& \textbf{73.3} & \textbf{89.9} & \underline{94.3} & \underline{59.0} & \underline{81.1} & \underline{87.7} 
& \textbf{73.8} & \textbf{91.2} & \textbf{94.9} & \underline{54.8} & \underline{76.2} & \underline{83.1} 
&73.9 & 91.7 & \underline{95.8} & \underline{75.5} & \underline{92.3} & \underline{95.6} 
&\textbf{68.4} & \textbf{80.5} & \textbf{82.7} & \underline{43.8} & \textbf{60.6} & \underline{65.2} 
&  81.1 & 75.4 & 77.3 & 72.6\\

Description CP 
&71.6 & 89.1 & 93.9 & 58.1 & 80.6 & 87.3 
& 69.3 & 88.6 & 93.4 & 53.7 & 75.0 & 81.8 
& \textbf{76.3} & \textbf{92.9} & 95.6 & 75.2 & 91.1 & 93.8 
& 66.5 & 79.1 & 81.6 & 39.7 & 55.2 & 60.1 
& 82.2 & 76.1 & \textbf{80.8} & 74.2  \\

Reasoning CP 
& 66.2 & 85.8 & 91.4 & \textbf{60.0} & \textbf{81.7} & \textbf{88.0} 
& 67.4 & 85.8 & 91.0 & \textbf{55.8} & \textbf{77.0} & \textbf{83.7} 
& 74.1 & 92.5 & \textbf{96.0} & \textbf{77.0} & \textbf{92.6} & \textbf{95.7} 
& \underline{67.2} & \underline{79.5} & \underline{81.8} & \textbf{44.4} & \textbf{60.6} & \textbf{65.3} 
&73.0 & 66.6 & 71.6 & 66.6\\

\bottomrule

\rowcolor{gray!20}
\multicolumn{29}{l}{\textit{Scale - ViT-Giant Opt/16 @384 (1B)}} \\
SigLIP2 
& \underline{72.4} & \textbf{90.6} & \textbf{94.7} & 56.5 & 78.6 & 85.5 
& 69.0 & 89.0 & 93.7 & 53.0 & 74.4 & 81.3 
& 58.6 & 78.3 & 85.3 & 58.2 & 78.2 & 83.6 
& 65.2 & \underline{79.7} & \textbf{83.1} & 40.0 & 57.0 & 62.5 
& \textbf{84.4} & \textbf{79.2} & \textbf{82.6} & \textbf{77.0} \\
+ Stage 1 
& 72.0 & \underline{89.5} & 94.0 & 59.3 & 81.3 & 88.0 
& 71.4 & 90.1 & \underline{94.5} & 55.5 & 76.6 & 83.3 
& 72.2 & \textbf{92.9} & \textbf{96.2} & \textbf{77.9} & \underline{92.3} & \textbf{96.0} 
&  65.1 & 78.4 & 81.1 & 43.6 & 61.1 & 66.6 
& \underline{82.9} & \underline{77.5} & \underline{80.8} & \underline{75.8} \\

+ Stage 2 
& \textbf{73.7} & 89.3 & 93.9 & \underline{60.1} & \underline{81.7} & \underline{88.5} 
& \textbf{75.3} & \textbf{91.6} & \textbf{95.3} & \underline{56.7} & \underline{77.5} & \underline{84.0} 
& 71.0 & 90.4 & 94.0 & 75.5 & 90.0 & 94.7 
&  \textbf{67.4} & \underline{79.7} & 81.9 & \underline{47.2} & \underline{64.8} & \underline{70.2} 
& 80.0 & 74.4 & 76.2 & 71.2\\

Description CP 
& 72.1 & 89.4 & \underline{94.3} & 59.4 & 81.1 & 87.8 
& 71.5 & \underline{90.2} & \underline{94.5} & 55.5 & 76.5 & 83.2 
& \textbf{76.3} & \textbf{92.9} & 95.6 & 75.2 & 91.1 & 93.8 
& 65.6 & 78.5 & 81.0 & 41.9 & 59.0 & 64.5 
& 82.8 & 77.3 & \underline{80.8} & 75.4 \\

Reasoning CP 
& 70.5 & 88.5 & 93.2 & \textbf{61.5} & \textbf{82.8} & \textbf{89.1} 
& \underline{71.7} & 89.4 & 93.9 & \textbf{57.8} & \textbf{78.6} & \textbf{84.9} 
& \underline{74.1} & \underline{92.5} & \underline{96.0} & \underline{77.0} & \textbf{92.6} & \underline{95.7} 
&  \underline{67.3} & \textbf{80.1} & \underline{82.9} & \textbf{47.9} & \textbf{65.5} & \textbf{70.7} 
& 78.3 & 72.3 & 76.3 & 71.1  \\
\bottomrule
\end{tabular}}
\end{table*}

\subsection{Detailed Results on Visually Grounded Reasoning Benchmarks}
The detailed results are shown in \ref{tab:app: commonsense_reasoning}.
\label{appendix: Results: Visual Ground Reasoning}

\begin{table}[htbp]
\centering
\caption{Visually Grounded Reasoning performance of ReasonCLIP.}
\label{tab:app: commonsense_reasoning}
\setlength{\tabcolsep}{3pt}
\resizebox{\linewidth}{!}{
\begin{tabular}{l|ccccc|cccccc|cccccc|cccccc}
\toprule
\multirow{3}{*}{Model}  

& \multicolumn{5}{c|}{WinoGAViL \cite{bitton2022winogavil}} 
& \multicolumn{18}{c}{RCLIP-Bench \textbf{(Ours)}} \\
& \multicolumn{5}{c|}{Candidates} 
& \multicolumn{6}{c|}{V1: Visual Grounding} 
& \multicolumn{6}{c|}{V2: Evidence Awareness} 
& \multicolumn{6}{c}{V3: Visual Reasoning} \\
 &  C5 & C6 & C10 & C12 & Avg. 
& Att. & Nou. & Rel. & Sen. & Sub. & Avg.
& S. & A. & H. & T. & P. & Avg.
& S. & A. & H. & T. & P. & Avg. \\
\bottomrule
\rowcolor{gray!20}
\multicolumn{24}{l}{\textit{Scale - ViT Base/32 (86M)}} \\
CLIP
&54.9 & 52.0 & 44.7 & 38.2 & 50.6
& 22.4 & 28.9 & \textbf{16.2} & 17.0 & 27.2 & 22.3
& 12.9 & 15.4 & 11.3 & \underline{26.5} & 15.2 & 16.2
& \underline{19.6} & 23.7 & 21.6 & 22.5 & \underline{25.8} & 22.6
\\
+ Stage 1
&56.5 & \underline{55.2} & \underline{49.2} & \underline{42.3} & \underline{53.4}
& 22.5 & \textbf{31.4} & 14.2 & 18.5 & \underline{31.4} & \underline{23.6}
& \textbf{18.2} & \textbf{19.3} & \textbf{25.0} & \textbf{31.5} & \textbf{21.3} & \textbf{23.1}
& 18.5 & \underline{31.6} & 21.5 & \underline{23.5} & 23.6 & \underline{23.8}
\\
+ Stage 2
&\textbf{59.8} & \textbf{55.9} & \textbf{52.7} & \textbf{46.4} & \textbf{55.8}
& \underline{23.9} & 29.9 & 15.1 & 18.1 & 30.4 & 23.5
& 12.1 & 14.7 & \underline{16.3} & 22.7 & 14.4 & 16.0
& 17.2 & 26.5 & \underline{24.3} & 20.2 & 22.2 & 22.1
\\
Description CP
&\underline{56.6} & 54.3 & 49.0 & 42.0 & 53.0
& 21.9 & 30.0 & 15.0 & \underline{18.6} & \underline{31.4} & 23.4
& 13.0 & 15.5 & 13.7 & 25.1 & \underline{20.2} & \underline{17.5}
& 19.0 & \textbf{34.2} & \textbf{25.3} & 17.1 & \textbf{26.9} & \textbf{24.5}
\\
Reasoning CP
&56.1 & 54.0 & 48.5 & 40.8 & 52.6
& \textbf{26.9} & \underline{31.2} & \textbf{16.2} & \textbf{19.8} & \textbf{32.1} & \textbf{25.2}
& \underline{14.7} & \underline{17.8} & 11.1 & 25.1 & 16.2 & 17.0
& \textbf{21.2} & 25.3 & 18.5 & \textbf{23.6} & 22.9 & 22.3
\\
\bottomrule

\rowcolor{gray!20}
\multicolumn{24}{l}{\textit{Scale - ViT Large/14 @224 (307M)}} \\
CLIP
&54.3 & 52.3 & 42.8 & 39.3 & 50.4
& 20.4 & 31.1 & 13.9 & 17.5 & 29.8 & 22.5
& 8.2 & 13.0 & 12.5 & 32.1 & \textbf{20.8} & 17.3
& \textbf{23.4} & 25.8 & \textbf{24.1} & 21.2 & \textbf{29.3} & \textbf{24.8}
\\
+ Stage 1
&58.1 & 57.2 & 48.7 & 48.4 & 55.4
& 25.8 & \underline{36.7} & 15.0 & 19.3 & 36.0 & 26.6
& 11.7 & \textbf{19.2} & \textbf{15.7} & 34.3 & 15.2 & \textbf{19.2}
& 19.6 & \textbf{28.3} & 18.0 & \textbf{26.7} & \underline{28.8} & \underline{24.3}
\\
+ Stage 2
&\textbf{62.8} & \underline{62.2} & \textbf{57.2} & \underline{56.0} & \textbf{61.1}
& \textbf{30.1} & 36.5 & \textbf{17.6} & 20.6 & \underline{38.0} & \textbf{28.6}
& 11.4 & 17.4 & 13.9 & \underline{34.4} & 14.4 & 18.3
& 20.5 & \underline{26.0} & \underline{23.6} & 21.6 & 23.5 & 23.1
\\
+ Stage 2 w/o cls
&\textbf{62.8} & \textbf{62.5} & \underline{55.9} & \textbf{56.2} & \textbf{61.1}
& \textbf{30.1} & 36.1 & \textbf{17.6} & \underline{20.9} & \textbf{38.5} & \textbf{28.6}
& 11.4 & 17.2 & 13.8 & 33.9 & 13.6 & 18.0
& 19.6 & 25.8 & 23.5 & 20.7 & 23.3 & 22.6
\\
Description CP
&59.1 & 57.6 & 49.5 & 47.6 & 56.0
& 24.6 & 34.9 & 15.4 & 20.1 & 36.1 & 26.2
& \underline{13.0} & 18.1 & 10.8 & \textbf{34.9} & \underline{15.9} & 18.5
& \underline{22.5} & 25.4 & 20.2 & 18.5 & 22.7 & 21.9
\\
Reasoning CP
&60.6 & 58.7 & 49.9 & 48.7 & 57.2
& 29.5 & \textbf{37.9} & 16.3 & \textbf{21.2} & 37.8 & 28.5
& \textbf{13.7} & \underline{19.1} & \underline{14.8} & 33.4 & 14.8 & \textbf{19.2}
& 22.4 & 24.3 & 18.2 & \underline{22.9} & 22.9 & 22.1
\\
\bottomrule

\rowcolor{gray!20}
\multicolumn{24}{l}{\textit{Scale - ViT Large/14 @336 (307M)}} \\
CLIP
&53.7 & 52.4 & 41.9 & 40.8 & 50.2
& 21.3 & 31.7 & 13.7 & 17.6 & 30.3 & 22.9
& 8.2 & 13.3 & 13.5 & 32.1 & \textbf{21.3} & 17.7
& \textbf{24.2} & 26.0 & \textbf{25.4} & 21.2 & \underline{28.3} & \underline{25.0}
\\
+ Stage 1
&58.1 & 56.2 & 49.0 & 45.9 & 54.8
& 26.4 & \underline{37.5} & 15.4 & 19.7 & 36.8 & 27.2
& 11.7 & \underline{19.8} & \textbf{15.6} & \underline{34.3} & 14.9 & \textbf{19.3}
& 20.3 & \textbf{29.7} & 19.5 & \textbf{28.8} & \textbf{28.6} & \textbf{25.4}
\\
+ Stage 2
&\textbf{62.9} & \textbf{61.2} & \textbf{55.9} & \textbf{53.3} & \textbf{60.2}
& \textbf{31.1} & \textbf{38.0} & \textbf{18.5} & \textbf{21.3} & \textbf{38.5} & \textbf{29.5}
& 12.2 & 17.4 & 14.0 & 31.4 & 15.2 & 18.0
& 20.9 & \underline{26.8} & \underline{23.8} & \underline{22.3} & 24.4 & 23.6
\\
Description CP
&57.7 & 56.3 & 48.3 & 47.3 & 54.8
& 23.6 & 35.9 & 14.8 & 20.3 & 35.5 & 26.0
& \underline{12.3} & 18.9 & 9.5 & \textbf{34.6} & \underline{16.0} & 18.2
& 21.9 & 25.5 & 21.4 & 18.8 & 23.0 & 22.1
\\
Reasoning CP
&\underline{61.7} & \underline{58.9} & \underline{51.2} & \underline{50.1} & \underline{58.0}
& \underline{30.7} & 37.4 & \underline{16.8} & \underline{21.0} & \underline{38.0} & \underline{28.8}
& \textbf{14.1} & \textbf{20.2} & \underline{14.2} & 32.9 & 14.5 & \underline{19.2}
& \underline{22.2} & 26.7 & 19.9 & 21.5 & 25.3 & 23.1
\\
\bottomrule

\rowcolor{gray!20}
 \multicolumn{24}{l}{\textit{Scale - ViT So400m/14 @384(428M)}} \\
SigLIP
&63.9 & \underline{62.0} & \underline{58.6} & \underline{56.8} & \underline{61.7}
& 24.0 & 33.3 & 14.3 & 21.2 & 29.3 & 24.4
& 13.0 & 17.3 & 28.0 & \textbf{30.3} & 22.2 & 22.2
& 20.6 & 25.6 & 17.7 & \textbf{30.5} & 14.5 & 21.8
\\
+ Stage 1
&59.1 & 58.2 & 50.8 & 51.0 & 56.8
& 32.1 & \textbf{38.5} & \underline{17.2} & 25.1 & 41.1 & \underline{30.8}
& \underline{16.1} & \textbf{29.1} & \textbf{28.8} & 24.9 & \textbf{30.1} & \textbf{25.8}
& \textbf{26.2} & \textbf{34.3} & 21.0 & \underline{23.9} & \textbf{24.5} & \textbf{26.0}
\\
+ Stage 2
&61.8 & 58.7 & 54.2 & 52.1 & 58.5
& \underline{32.6} & 37.3 & 16.7 & \underline{25.3} & \underline{41.6} & 30.7
& 14.4 & 25.1 & 25.8 & \underline{27.4} & 27.5 & 24.1
& 20.7 & 27.5 & \textbf{26.0} & 20.7 & 19.3 & 22.8
\\
+ Stage 2 w/o cls
&\textbf{64.6} & 61.9 & 57.5 & 56.6 & \textbf{61.8}
& 31.3 & 36.7 & 16.5 & 24.5 & 40.0 & 29.8
& 14.6 & 20.3 & 27.3 & 24.5 & 26.3 & 22.6
& 17.9 & 21.0 & \underline{24.8} & 19.3 & 19.3 & 20.5
\\
Description CP
&56.0 & 54.3 & 46.2 & 45.7 & 52.9
& \textbf{33.1} & \underline{37.8} & \textbf{17.7} & \textbf{26.6} & \textbf{42.6} & \textbf{31.5}
& 14.7 & 23.7 & 19.7 & 23.7 & \underline{29.0} & 22.2
& \underline{23.1} & \underline{31.6} & 21.5 & 19.8 & \underline{21.8} & \underline{23.6}
\\
Reasoning CP
&\underline{64.3} & \textbf{64.7} & \textbf{62.0} & \textbf{59.9} & 60.9
& 25.7 & 34.5 & 14.9 & 22.2 & 31.8 & 25.8
& \textbf{16.4} & \underline{26.9} & \underline{28.4} & 24.4 & 26.0 & \underline{24.4}
& 20.0 & 25.6 & 23.9 & 21.9 & 21.4 & 22.6
\\
\hline
SigLIP2
&63.2 & 61.1 & 57.2 & 56.4 & 60.8
& 15.7 & 19.9 & 14.3 & 12.1 & 17.6 & 15.9
& \textbf{17.6} & 19.1 & \underline{30.7} & \textbf{35.7} & 18.3 & 24.3
& \textbf{20.5} & 23.6 & \textbf{23.1} & \textbf{25.0} & 16.2 & 21.7
\\
+ Stage 1
&66.1 & \underline{64.3} & 60.0 & \underline{60.3} & \underline{64.0}
& \underline{18.1} & 20.9 & \textbf{15.0} & \underline{12.5} & \underline{17.8} & \underline{16.9}
& \underline{17.2} & \textbf{27.5} & 30.5 & 27.9 & \textbf{28.0} & \textbf{26.2}
& \underline{20.4} & \textbf{34.3} & 18.3 & \underline{23.1} & \textbf{26.1} & \textbf{24.4}
\\
+ Stage 2
&\underline{66.7} & 63.1 & \underline{60.7} & 58.2 & 63.5
& \textbf{18.4} & \textbf{22.5} & \underline{14.9} & \textbf{13.0} & \textbf{17.9} & \textbf{17.3}
& 16.9 & 24.3 & 28.9 & 28.0 & 24.4 & 24.5
& 18.4 & 25.3 & 21.5 & 20.0 & 19.2 & 20.9
\\
Description CP
&64.4 & 62.8 & 59.0 & 56.2 & 62.1
& \underline{18.1} & 21.3 & 14.3 & 12.1 & 17.2 & 16.6
& 16.6 & 23.2 & 28.0 & \underline{29.7} & \underline{27.7} & 25.1
& 20.1 & \underline{28.6} & \underline{22.1} & 22.1 & \underline{23.1} & \underline{23.2}
\\
Reasoning CP
&\textbf{68.1} & \textbf{67.8} & \textbf{66.6} & \textbf{64.3} & \textbf{64.9}
& 15.5 & \underline{21.4} & 14.8 & 12.3 & 17.6 & 16.3
& 17.0 & \underline{27.4} & \textbf{32.3} & 29.4 & 22.3 & \underline{25.7}
& 19.7 & 24.9 & 19.7 & 22.3 & 21.3 & 21.6
\\
\bottomrule

\rowcolor{gray!20}
 \multicolumn{24}{l}{\textit{Scale - ViT Giant Opt/16 @384 (1B)}} \\
SigLIP2
&65.1 & 63.8 & \underline{62.9} & 60.1 & \underline{63.7}
& 15.0 & 20.4 & \textbf{15.2} & 11.5 & 15.6 & 15.6
& 16.1 & 19.3 & \textbf{32.1} & \textbf{28.7} & 21.2 & 23.5
& \textbf{20.2} & 23.5 & \underline{20.5} & \textbf{29.8} & 16.9 & 22.2
\\
+ Stage 1
&66.1 & \underline{64.3} & 60.0 & \underline{60.3} & 61.6
& \underline{17.8} & \textbf{22.3} & 14.7 & 13.3 & \underline{18.7} & \underline{17.4}
& 16.1 & \textbf{28.4} & 27.8 & 24.8 & \underline{30.0} & 25.4
& \underline{20.1} & \textbf{30.4} & 15.9 & \underline{25.6} & \textbf{25.8} & \textbf{23.6}
\\
+ Stage 2
&\underline{66.7} & 63.1 & 60.7 & 58.2 & 62.7
& 17.1 & 20.6 & \underline{14.8} & \textbf{13.4} & 17.8 & 16.8
& 16.2 & 24.9 & 24.6 & 27.6 & 24.9 & 23.6
& 19.4 & 26.4 & 19.6 & 21.2 & 21.0 & 21.5
\\
Description CP
&62.1 & 62.5 & 60.8 & 59.9 & 61.8
& \textbf{18.2} & \underline{22.1} & 14.2 & 13.2 & \textbf{20.1} & \textbf{17.6}
& \textbf{16.9} & 26.1 & 27.7 & 27.0 & \textbf{31.2} & \textbf{25.8}
& 19.7 & 27.1 & \textbf{20.8} & 23.2 & \underline{25.5} & \underline{23.3}
\\
Reasoning CP
&\textbf{68.1} & \textbf{67.8} & \textbf{66.6} & \textbf{64.3} & \textbf{67.4}
& 17.3 & 20.8 & 14.7 & \textbf{13.4} & 18.2 & 16.9
& \underline{16.5} & \underline{27.6} & \underline{31.6} & \textbf{28.7} & 24.2 & \underline{25.7}
& 19.2 & \underline{28.4} & 18.7 & 23.3 & 22.1 & 22.3
\\
\bottomrule
\end{tabular}
}
\end{table}

\subsection{Detailed Results on Compositional Reasoning Benchmarks}
The detailed results are shown in \ref{tab:app: compositional_reasoning}.
\label{appendix: Results: Compositional}
\begin{table}[htbp]
\centering
\caption{Abalation Results on compositional reasoning benchmarks of different Stages.}
\label{tab:app: compositional_reasoning}
\resizebox{0.7\linewidth}{!}{
\begin{tabular}{lcccccccc}
\toprule
 \multirow{2}{*}{Model} & \multirow{2}{*}{WhatsUp} & \multirow{2}{*}{VALSE} & \multirow{2}{*}{CREPE} & \multirow{2}{*}{SugarCrepe} & \multicolumn{2}{c}{SugarCrepe++} & \multirow{2}{*}{Average} \\
\cmidrule(lr){6-7}
& & & &  &  ITT & TOT & \\
\bottomrule
\rowcolor{gray!20}
\multicolumn{8}{l}{\textit{Version - ViT-B/32 @224}} \\
CLIP     & 41.0 & 67.4 & 23.9 & 73.2 & 60.0 & 46.7 & 52.0 \\
+ Stage 1    & 42.4 & 68.7 & 18.1 & 75.0 & \underline{63.4} & \underline{63.2} & 55.1 \\
+ Stage 2    & \textbf{51.3} & 72.5 & \underline{24.0} & 75.6 & 61.5 & 61.5 & \underline{57.7} \\
+ READ \cite{kwon2025enhancing} & 45.2 & \textbf{79.2} & \textbf{44.9} & \textbf{88.5} & \textbf{69.2} & \textbf{67.0} & \textbf{65.7} \\
Description CP   & \underline{46.6} & \underline{75.1} & 20.4 & \underline{77.7} & 63.0 & 61.5 & 57.4 \\
Reasoning CP     & 43.7 & 67.8 & 19.6 & 74.1 & 62.5 & 57.6 & 54.2 \\
\bottomrule

\rowcolor{gray!20}
\multicolumn{8}{l}{\textit{Version - ViT-L/14 @224}} \\
CLIP     & 40.7 & 68.8 & \underline{20.5} & 73.4 & 60.6 & 44.3 & 51.4 \\
+ Stage 1    & 41.3 & 71.1 & 16.5 & 75.2 & \textbf{65.0} & \underline{60.2} & 54.9 \\
+ Stage 2    & \underline{46.0} & \underline{75.9} & \textbf{24.8} & \textbf{78.2} & 63.1 & 59.5 & \textbf{58.0} \\
+ Stage 2 w/o cls    & 41.2 & 71.8 & 18.9 & 74.8 & 62.8 & 59.8 & 54.9 \\
Description CP   & \textbf{47.2} & \textbf{76.4} & 20.2 & \underline{77.9} & 63.2 & \textbf{60.4} & \underline{57.6} \\
Reasoning CP     & 42.1 & 70.9 & 16.2 & 73.1 & \underline{63.3} & 58.0 & 53.9 \\
\bottomrule

 \rowcolor{gray!20}
 \multicolumn{8}{l}{\textit{Version - ViT-L/14 @336}} \\
CLIP     & 41.8 & 68.8 & \underline{20.9} & 74.8 & 60.6 & 44.1 & 51.8 \\
+ Stage 1    & 42.7 & 71.2 & 16.1 & 74.4 & \textbf{65.2} & \underline{60.0} & \underline{54.9} \\
+ Stage 2     & \underline{46.3} & \textbf{76.3} & \textbf{24.8} & \underline{78.1} & 63.8 & 59.4 & \textbf{58.1} \\
Description CP   & \textbf{47.3} & \textbf{76.3} & 20.1 & \textbf{79.5} & \underline{65.0} & \textbf{60.4} & \textbf{58.1} \\
Reasoning CP     & 41.5 & \underline{71.2} & 17.3 & 73.7 & 63.8 & 57.1 & 54.1 \\
\bottomrule

\rowcolor{gray!20}
 \multicolumn{8}{l}{\textit{Version - ViT-so400m/14 @384}} \\
SigLIP & 47.6 & 72.0 & 18.0 & 83.0 & 67.9 & 51.2 & 56.6 \\
+ Stage 1    & 48.2 & \underline{77.5} & \underline{23.7} & 86.9 & \textbf{74.0} & \textbf{75.6} & \underline{64.3} \\
+ Stage 2    & \underline{49.7} & 76.3 & 20.1 & 84.1 & \underline{73.1} & 72.9 & 62.7 \\
+ Stage 2 w/o cls    & 49.3 & \textbf{78.2} & \textbf{26.7} & \textbf{88.2} & \underline{73.1} & 72.9 & \textbf{64.7} \\
Description CP   & \textbf{50.8} & 76.8 & 21.7 & \underline{87.5} & 72.2 & \underline{73.2} & 63.7 \\
Reasoning CP   & 47.1 & 71.8 & 12.5 & 73.6 & 72.4 & 62.5 & 56.7 \\
\hline
SigLIP2 & 44.4 & 70.8 & 17.0 & \textbf{82.1} & \textbf{67.5} & 49.2 & 55.2 \\
+ Stage 1    & \underline{44.9} & \underline{76.2} & \underline{21.3} & 75.6 & 65.6 & \textbf{73.0} & \textbf{59.4} \\
+ Stage 2    & 44.6 & \textbf{76.8} & \textbf{21.7} & 76.0 & 65.8 & 70.7 & \underline{59.3} \\
Description CP   & \textbf{45.3} & 74.9 & 18.6 & \underline{78.7} & \underline{66.8} & \underline{71.6} & \underline{59.3} \\
Reasoning CP  & 44.2 & 70.9 & 15.6 & 67.0 & 62.6 & 68.2 & 54.8 \\
\bottomrule
\rowcolor{gray!20}
 \multicolumn{8}{l}{\textit{Version - ViT-Giant Opt/16 @384}} \\
SigLIP2 & \textbf{46.9} & 71.4 & 16.5 & \textbf{83.0} & 67.4 & 48.8 & 55.7 \\
+ Stage 1    & 45.6 & \underline{76.2} & \textbf{21.2} & 78.2 & 68.4 & \textbf{72.7} & \textbf{60.4} \\
+ Stage 2    & 45.3 & \textbf{77.9} & \underline{20.5} & 76.8 & 67.0 & 66.9 & 59.1 \\
Description CP   & \underline{46.8} & 74.9 & 18.9 & \underline{79.2} & \underline{68.8} & \underline{71.2} & \underline{59.9} \\
Reasoning CP  & 45.3 & 74.6 & 18.0 & 74.5 & \textbf{69.5} & 68.1 & 58.3 \\
\bottomrule
\end{tabular}
}
\end{table}

\newpage
\section{Literature Review}
\label{appendix: Literature Review}

\subsection{History of CLIP-style Models}
While the main text primarily focuses on descriptive CLIP models and reasoning-enhanced models, here in Table \ref{tab:ex_clip_foundation}   we provide a more comprehensive overview of CLIP's development and optimizations across various directions, highlighting only representative works.
\begin{table*}[htbp]
\centering
\caption{CLIP models.}
\resizebox{\linewidth}{!}{
\begin{tabular}{llll}
\toprule
Model & Reference & \makecell[l]{Core Contribution \\ \& Objective} & Type \\
\midrule

CLIP & \cite{radford2021learning} & \makecell[l]{Image–text contrastive pretraining. \\ Learning a shared image–text embedding space.} & Model (Vanilla) \\
\midrule

ALIGN & \cite{jia2021scaling} & \makecell[l]{Larger data scale can compensate for noise. \\ Scaling up with noisy web data.} & Data (Scaling)\\
\midrule

OpenCLIP &\cite{ilharco_gabriel_2021_5143773} & \makecell[l]{Open-source reproduction with LAION dataset. \\ Reliable reimplementation of CLIP.} & Model (Reproduction)\\
\midrule

RegionCLIP & \cite{zhong2022regionclip} & \makecell[l]{Region-level CLIP pretraining. \\ Aligning visual regions with text.} & Spatial (Detection)\\
\midrule

GLIP & \cite{li2022grounded} & \makecell[l]{Language-driven detection head. \\ Unify detection with language grounding.} & Spatial (Detection)\\
\midrule

Winoground &\cite{thrush2022winoground}  & \makecell[l]{CLIP doesn’t understand compositional language. \\ Expose weakness in compositional semantics.} & Reasoning (Benchmark)\\
\midrule

DenseCLIP & \cite{rao2022denseclip} & \makecell[l]{Use CLIP text embeddings as dense classifiers. \\ Extend CLIP to pixel or patch-level tasks.} & Spatial (Segmentation)\\
\midrule

CyCLIP & \cite{goel2022cyclip} & \makecell[l]{Cycle-consistency loss. \\ Improve consistency between image-text spaces.} & Reasoning (Consistency)\\
\midrule

FLIP & \cite{li2023scaling} & \makecell[l]{Mask-based efficient CLIP training. \\ Enable faster, scalable CLIP pre-training.} & Model (Efficiency) \\
\midrule

SigLIP & \cite{zhai2023sigmoid} & \makecell[l]{Sigmoid-based contrastive loss. \\ Improve performance and efficiency.} & Model (Loss)\\
\midrule

NegCLIP & \cite{yuksekgonul2023when} & \makecell[l]{Composition-aware hard negatives. \\ Diagnose compositional failures.} & Reasoning (Benchmark) \\
\midrule

MetaCLIP & \cite{xu2023demystifying}& Reveal CLIP’s hidden data curation mechanism. & Data (Curation)\\
\midrule

DFN & \cite{fang2023data} &\makecell[l]{Learned data filtering for CLIP pre-training.\\
Select high-quality data via a trained scoring network.} & Data (Filtering)\\
\midrule

SCLIP & \cite{wang2024sclip} &\makecell[l]{Reform self-attention for dense CLIP inference.\\
Introduce correlative self-attention.} & Spatial (Segmentation)\\
\midrule

TripletCLIP & \cite{patel2024tripletclip} &\makecell[l]{Triplet supervision with generated negatives.  \\ Improve compositional reasoning.} & Reasoning (Logic) \\
\midrule

Long-CLIP & \cite{zhang2024long} &\makecell[l]{Extend CLIP to long-context text inputs.\\
Via positional embedding adaptation.} & Model (Long-context)\\
\midrule

ResCLIP & \cite{yang2025resclip}& \makecell[l]{Residual cross-layer attention for dense CLIP inference.\\
Leverage intermediate cross-correlation attention.}  & Spatial (Segmentation)\\
\midrule

NegBench & \cite{alhamoud2025vision} & \makecell[l]{NegBench Data. \\ Evaluate negation.} & Reasoning (Benchmark) \\
\midrule

MetaCLIP2 & \cite{chuang2025meta} & \makecell[l]{Scale CLIP to worldwide multilingual data.\\
Metadata-driven multilingual curation.} & Data (Multilingual) \\
\midrule

FSC-CLIP & \cite{oh2024preserving} & \makecell[l]{Fine-grained hard negative supervision for compositionality.\\
Preserve zero-shot multi-modal performance.}& Reasoning (Composition)\\
\midrule

SuperCLIP & \cite{zhao2025superclip} &\makecell[l]{Add classification supervision to CLIP pretraining.\\
Recover fine-grained token-level alignment.} & Model (Loss)\\
\midrule

READ-CLIP & \cite{kwon2025enhancing} &\makecell[l]{Reconstruction and alignment objectives for text encoder.\\
Enhance compositional reasoning beyond contrastive learning} & Reasoning (Composition)\\
\midrule

\end{tabular}
}

\label{tab:ex_clip_foundation}
\end{table*}

\subsection{Visual Backbone in MLLMs}
As a supplement to the main text, this section briefly discusses the evolution of visual backbones for MLLMs and justifies our focus on CLIP-style encoders. Early visual encoders for MLLMs were predominantly based on \textbf{Self-Supervised Learning (SSL)}, such as DINO~\cite{zhang2022dino}
and its successor DINOv2~\cite{oquab2023dinov2}, as well as AIM~\cite{el2024scalable}, which learn rich visual representations purely from unlabeled image data. These were later joined by \textbf{Contrastive Language-Image Pre-training (CLIP-style)} models, including CLIP~\cite{radford2021learning}, SigLIP~\cite{zhai2023sigmoid}, and the EVA-CLIP series~\cite{sun2023eva,sun2024eva}, which align visual features with natural language descriptions via large-scale contrastive training. As MLLMs matured, large-scale empirical studies began systematically comparing these two families. Recent works~\cite{tong2024cambrian,mckinzie2024mm1} consistently find that CLIP-style, language-supervised encoders significantly outperform SSL-based counterparts when integrated into MLLMs, owing to the inherent alignment between their representations and the semantic space of language models. This evidence motivates our work to focus on the CLIP family as the backbone of choice. Our proposed \textbf{ReasonCLIP} further enhances CLIP-style encoders with reasoning-oriented supervision through non-intrusive continual pre-training, bridging the gap between descriptive alignment and compositional reasoning.

\end{document}